\documentclass[3p,times]{elsarticle}

\usepackage{times}
\usepackage[utf8]{inputenc} 
\usepackage[T1]{fontenc}    
\usepackage{hyperref}       
\usepackage{url}            
\usepackage{booktabs}       
\usepackage{amsfonts}       
\usepackage{nicefrac}       
\usepackage{microtype}      
\usepackage{xcolor}         
\usepackage{enumitem}
\usepackage{caption, rotating, subcaption}
\usepackage{graphicx}
\usepackage{amsmath}
\usepackage{amssymb}
\usepackage{mathtools}
\usepackage{amsthm}
\usepackage{float}
\usepackage{multirow}
\usepackage{Files/picins}
\usepackage{adjustbox}
\usepackage{soul}

\usepackage{lipsum} 
\usepackage{capt-of} 

\def\X {\mathbf{X}}
\def\x {\mathbf{x}}
\def\Y {\mathbf{Y}}

\def \Rspace { \mathbb{R}}

\theoremstyle{remark}
\newtheorem*{remark}{Remark}

\begin{document}

\begin{frontmatter}

\title{From Canopy to Ground via ForestGen3D: Learning Cross-Domain Generation of 3D Forest Structure from Aerial-to-Terrestrial LiDAR}


\author[a]{Juan Castorena}
\author[b]{E. Louise Loudermilk}
\author[c]{Scott Pokswinski}
\author[a]{Rodman Linn}
\address[a]{Los Alamos National Laboratories, Los Alamos, NM, 48124 USA}
\address[b]{Southern Research Station, Disturbance and Prescribed Fire Laboratory, 320 E Green St., Athens, GA 30606.}
\address[c]{New Mexico Consortium, 4200 W Jemez Rd 200, Los Alamos, NM 87544}

\begin{abstract}
The 3D structure of living and non-living components in ecosystems plays a critical role in determining ecological processes and feedbacks from both natural and human-driven disturbances. Anticipating the effects of wildfire, drought, disease, or atmospheric deposition depends on accurate characterization of 3D vegetation structure, yet widespread measurement remains prohibitively expensive and often infeasible.
We present ForestGen3D, a cross-domain generative framework that preserves ALS observed 3D forest structure while inferring missing sub-canopy detail. ForestGen3D is based on conditional denoising diffusion probabilistic models trained on co-registered ALS and terrestrial LiDAR (TLS) data, exploiting the empirical containment of TLS structure within the ALS derived spatial envelope. The model generates realistic TLS-like point clouds that remain spatially consistent with ALS geometry, enabling landscape-scalable reconstruction of full vertical forest structure. To ensure ecological plausibility, we introduce a geometric containment prior based on the convex hull of ALS observations and provide theoretical and empirical guarantees that generated structures remain spatially consistent. 
We evaluate ForestGen3D at tree, plot, and landscape scales using real-world data from mixed conifer ecosystems, and show through qualitative and quantitative geometric and distributional analyses that it produces high-fidelity reconstructions closely matching TLS reference data in terms of 3D structural similarity and downstream biophysical metrics, including tree height, DBH, crown diameter, and crown volume. We further demonstrate that the resulting expected point containment (EPC) serves as a practical proxy for generation quality in settings where TLS ground truth is unavailable. Our results demonstrate that ForestGen3D enhances the utility of ALS only environments by inferring ecologically plausible sub-canopy structure while faithfully preserving the landscape heterogeneity encoded in ALS observations, thereby providing a richer 3D representation for ecological analysis, structural fuel characterization and related remote sensing applications.
\end{abstract}


\end{frontmatter}

\section{Introduction}
\label{sec:intro}

Wildfires, drought, pathogens, and other natural disturbances are increasingly transforming forested landscapes, driving the need for improved ecological monitoring and fuel structure characterization. These disturbances are shaped not only by vegetation type but also by the three-dimensional (3D) arrangement of biomass across vertical and horizontal layers. The 3D structure of both living and non-living components of vegetation plays a critical role in regulating ecological processes, including energy and moisture fluxes, carbon storage, and fire behavior \cite{vierling2008lidar, kane2010comparisons, loudermilk2017role, Marcozzi_2025}.
\begin{figure*}[ht]
    \begin{minipage}{1.0\linewidth}
	\centering
	\includegraphics[width=0.9\linewidth]{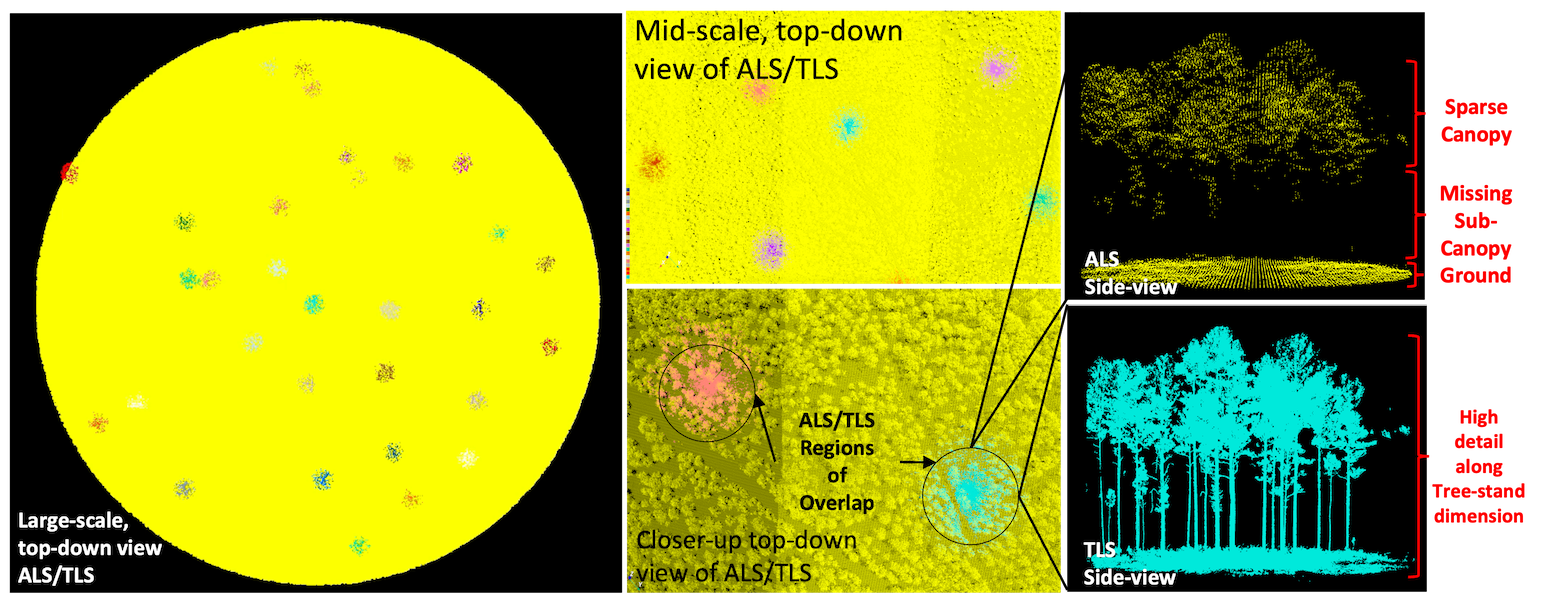}
	\captionof{figure}{Characteristics of ALS (in yellow) and TLS data at a surveying forest site, covering ~2.5 km in diameter with 31 TLS scans randomly distributed (a color-coded blob per scan). Overlapping ALS in top-right shows a side-view point cloud with sparse/missing information below the canopy. Co-registered TLS in the bottom-right, on the other hand shows high detail along the tree-stand direction but is limited in the lat-long direction due to physical range limitations of TLS (as shown in the top-down images).}
	\label{fig:broad_als_local_tls}
    \end{minipage}%
\end{figure*}
However, large-scale collection of detailed 3D vegetation structure remains a major bottleneck. Most landscape-scale ecological assessments still rely on field plots that are statistically upscaled using summary metrics \cite{bechtold2005enhanced}. To spatially represent forests, synthetic layouts are often generated by populating plots with idealized geometric models (e.g., cylinders, ovals), randomly placed according to field statistics \cite{kohler1998effects, dufour2012capsis, DEWERGIFOSSE2022150422, marechaux2021tackling}. While informative, these methods often overlook spatial relationships between vegetation components and fail to capture the full 3D spatial heterogeneity, both of which are crucial for modeling dynamic ecological processes. 
A related class of methods common in visualization, and gaming applications uses procedural forest generation, where tree forms and vegetation distributions are instantiated from parametric rules or stylized templates. Tools such as SpeedTree \cite{speedtree} and Unity’s Terrain Engine \cite{unityterrain} exemplify this approach, offering efficient generation of stylized vegetation for rendering purposes. While these models can quickly populate large virtual landscapes, they typically prioritize visual coherence and rendering performance over fidelity with real-world forest conditions, and are therefore less suitable for scientific analyses that depend on detailed and ecologically grounded 3D representations of forest fuel structure.

Over the past two decades, 3D remote sensing technologies particularly terrestrial laser scanning (TLS) \cite{pokswinski2021simplified} and aerial laser scanning (ALS) \cite{hyyppa2008review} have become increasingly valuable for vegetation mapping. TLS offers high-resolution 3D point clouds of stems, understory, and ground structure but is constrained to plot-scale deployment due to cost and labor demands. ALS, by contrast, provides broad landscape coverage and effectively captures canopy structure but suffers from occlusion and fails to resolve sub-canopy features in dense forests, even in full-waveform systems \cite{nelson1984determining, mallet2009full, castorena2015sampling}. These limitations are illustrated in Fig.\ref{fig:broad_als_local_tls}: while ALS provides landscape-scale coverage of the upper canopy, TLS offers detail below the canopy but lacks coverage in the lateral/longitudinal directions.
Surprisingly, most methods in forest mapping and monitoring do not exploit the complementary nature of these two sensing modalities. Instead, they typically rely on either ALS or TLS independently, developing analysis pipelines tailored to a single data source. This leads to structural blind spots for instance, missing understory in ALS-based workflows or extrapolated spatial sparsity in TLS-based ones that fail to respect landscape heterogeneity and that persist in operational products and ecological models \cite{calders2020terrestrial, hyyppa2012advances}.

\begin{figure*}[ht]
    \begin{subfigure}[b]{1.0\textwidth} 
	   \centering
	   \includegraphics[width=0.85\linewidth]{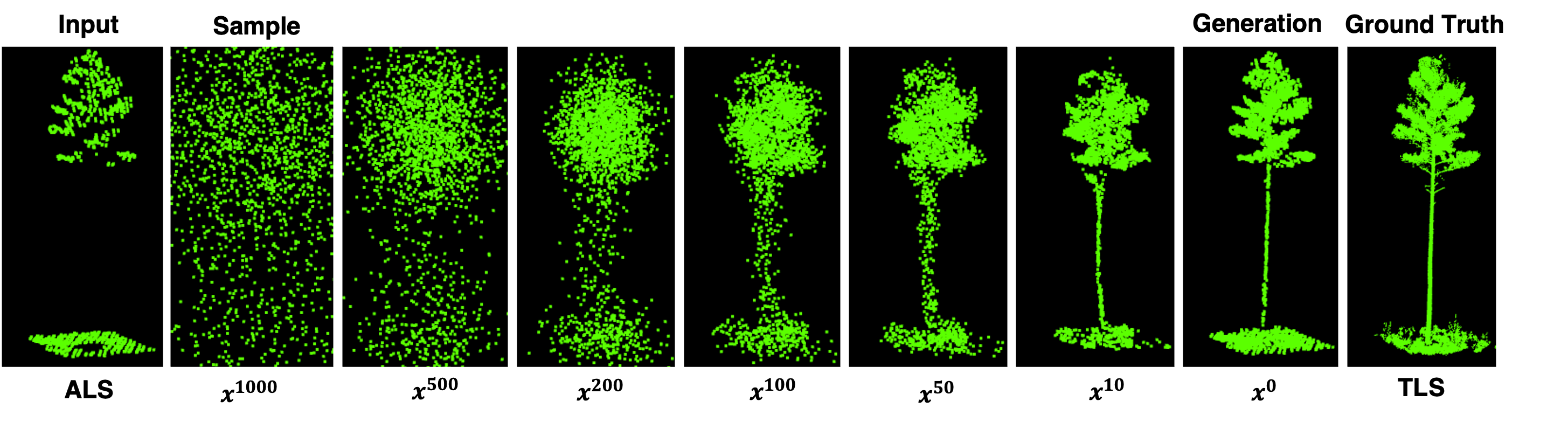}
            \captionsetup{labelformat=empty}
    \end{subfigure}
     \caption{ForestGen3D based on denoising diffusion probabilistic model (DDPM) formulation. Model iteratively denoises a sample guided by ALS measurements and generates a sample of forest tree 3D structure from the learned training distribution. Note that the 3D structural information provided by ALS is effective for constraining 3D structural generation in the vertical tree stand direction.} 
	\label{fig:denoising}
\end{figure*}

In this work, we propose to bridge these sensing gaps by learning generative models of 3D fuel structure that leverage the strengths of both data sources.
To achieve this, we develop ForestGen3D, a cross-domain generative modeling framework based on denoising diffusion probabilistic models (DDPMs). DDPMs are a class of deep generative models that learn the underlying data distribution by gradually adding noise to training examples and then learning to reverse this process through iterative denoising \cite{ho2020denoising,song2020score}. Once trained, the model can synthesize new samples by starting from pure Gaussian noise and progressively refining structure in a manner consistent with the learned distribution. In our case, the DDPM is conditioned on ALS inputs so that the denoising process respects the geometric information encoded in airborne LiDAR while generating plausible sub-canopy structure. These models are trained here on a new dataset of co-registered ALS/TLS tree 3D point clouds referred to as the \textit{CoLiDAR Forest3D (ALS/TLS)} and collected across mixed conifer vegetation ecosystems. Once trained, the model can generate novel 3D vegetation samples by starting from random noise and iteratively refining structure through a learned denoising process, guided by ALS input as shown in Fig. \ref{fig:denoising}. 
To the best of our knowledge, this is the first approach to leverage conditional diffusion models for cross-domain generation of 3D forest structure, bridging the gap between sparse, landscape-scale ALS and detailed, plot-scale, TLS representations. This enables scalable ecological modeling of forests in full vertical profile beyond existing upsampling or procedural simulation methods. 

Existing 3D generative and completion models, such as PCN \cite{yuan2018pcn}, 3D-GAN \cite{wu2016learning}, latent-GAN \cite{achlioptas2018learning}, and PointFlow \cite{yang2019pointflow}, have so far been developed and tested primarily on man-made objects from synthetic datasets such as ShapeNet. While effective in those controlled domains, their suitability for modeling real-world forest vegetation remains largely unknown. Forest structure presents fundamentally different challenges: complex topologies with intertwined crowns, occlusion-driven sparsity due to canopy cover, and vertical heterogeneity across understory, stems, and crowns. These characteristics may degrade the performance of models designed for compact, uniformly sampled shapes. Moreover, existing methods are typically trained without reference to real-world sensing modalities such as LiDAR, and therefore do not account for the anisotropic density, partial observability, and sensor-specific artifacts present in ecological data. In contrast, the objective of our approach is to generate ecologically realistic 3D vegetation structure directly from ALS/TLS data collected in mixed conifer forests, capturing both the natural variability of vegetation and the sensor dependent effects inherent in terrestrial and airborne LiDAR measurements.
In the sections that follow, we present our dataset and training approach, describe the DDPM-based model, and evaluate ForestGen3D across tree, plot, and landscape scale experiments. Quantitative and qualitative results demonstrate that the proposed approach approximates TLS-level structural detail using ALS data alone.

\section{ForestGen3D}
\label{sec:approach}

The core problem we address is generating synthetic forests by improving the fidelity of individual tree structures beyond conventional coarse-shape models. To achieve this, we explicitly preserve ecosystem fidelity by conditioning the generation process on ALS inputs, which provide canopy level structure tied to real forest layouts. Rather than modeling entire plots or landscapes directly which can compromise fine-grained detail and impose significant computational demands, we focus on learning 3D structure at the tree scale, where the physical representation of vegetation is most accurate. This localized modeling approach allows for more efficient training and data preparation, reducing the need to capture broad ecosystem variation in a single model. Moreover, it enables scalable deployment by applying the generative model independently to each detected tree at the plot or landscape scale, under the assumption that only regions beneath vegetation are susceptible to occlusion in ALS data.
To support this effort, we constructed a 3D forest dataset of co-registered ALS/TLS tree samples, which we used to train cross-domain denoising diffusion probabilistic models (DDPMs). These models learn the statistical distribution of tree 3D structure in a target domain (e.g., TLS) conditioned on observations in a reference domain (e.g., ALS). DDPMs start from random Gaussian noise and iteratively refine structure through a learned denoising process, enabling flexible and efficient generation of high-resolution 3D tree forms. Fundamentally, the framework learns to generate samples from the TLS distribution conditioned on a measured sample from the ALS distribution. Figure~\ref{fig:denoising} illustrates this process: an ALS input guides the denoising steps, which progressively reconstruct the full 3D structure of a tree, including stem and understory geometry missing from the input.

\subsection{Training dataset} \label{Ssec:train_dataset}

We construct a new 3D forest dataset \textit{CoLiDAR-Forest3D (ALS/TLS) dataset}, containing 2,900 co-registered TLS/ALS tree scans for training, including both individual trees and clusters of closely packed trees. The dataset was constructed using the co-registration approach from \cite{castorena2023automated} to align TLS scans with overlapping ALS data and the deep-learning based tree detection method from \cite{windrim2019forest} to extract individual trees.
The raw data were collected from a mixed-conifer forest at Fort Stewart in the United States \cite{snider2025fort}, primarily featuring species such as longleaf pine, slash pine, loblolly pine, and turkey oak. The understory consists of graminoids, forbs, and shrubs, including wiregrass, gallberry, and saw palmetto.
\begin{figure*}[t]
    \centering
    \includegraphics[width=1.0\linewidth]{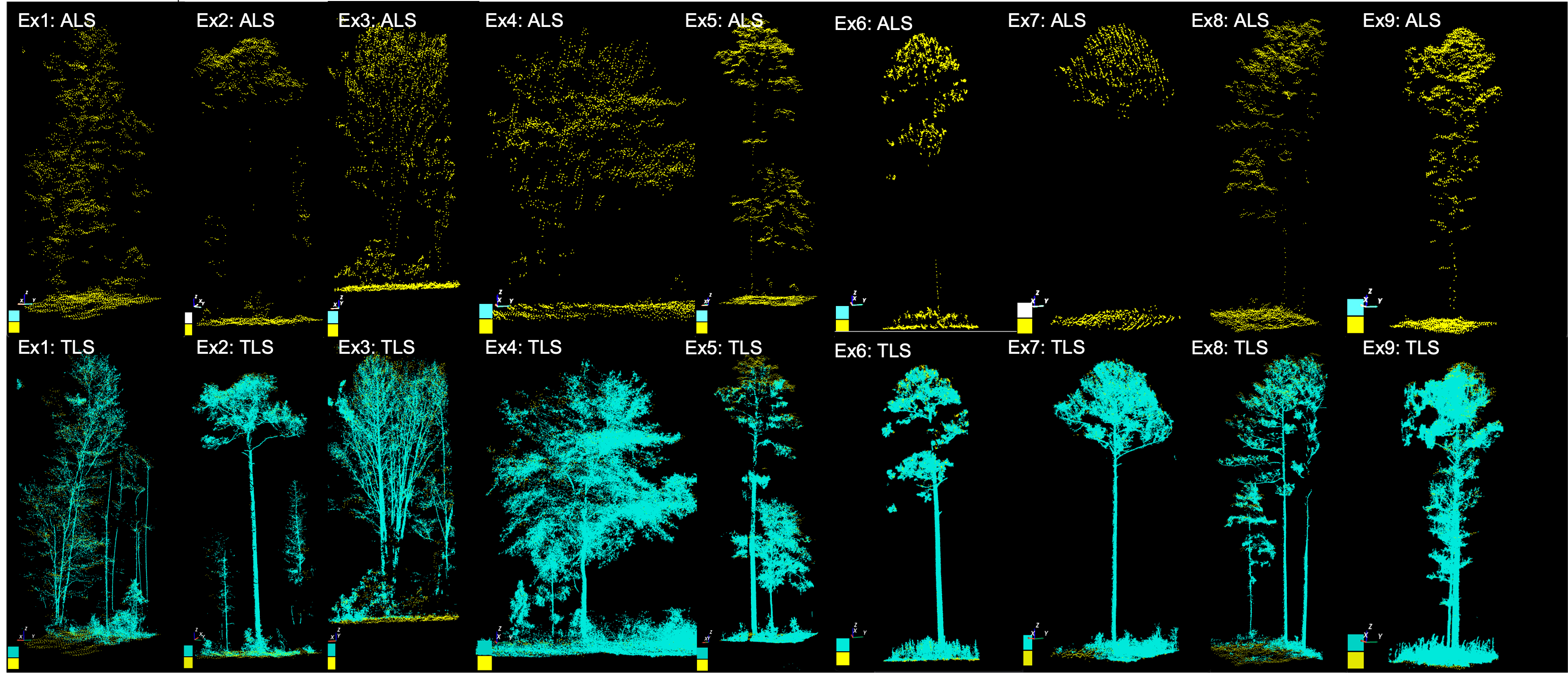}
    \caption{9 examples of the 3D training dataset consisting of 2900 examples of tree-segmented and co-registered aerial lidar scanning (ALS)/ terrestrial lidar scanning (TLS) trees. ALS trees in yellow in  the top row, corresponding co-registered TLS trees in aqua color in the bottom row.}
    \label{fig:dataset}
\end{figure*}
%
TLS data was acquired using a Leica BLK360 LiDAR scanner with an 830 nm wavelength laser and a field of view (FOV) of 300° (vertical) and 360° (horizontal). 
Scans were collected using a single-scan acquisition strategy, in which the instrument is deployed once at a plot centered position rather than through multiple co-registered scans. Although single viewpoint TLS does not fully eliminate occlusion, especially in dense forest understories, this strategy has been shown to be both effective and efficient for characterizing tree and plot level structural attributes across a wide range of forest ecosystems. Prior work demonstrates that randomized single-scan deployments provide reliable summary estimates of forest characterizations, while enabling practical data collection at larger spatial and ecological extents \cite{pokswinski2021simplified, hawley2018novel, loudermilk2009ground}. 
Consequently, our dataset prioritizes broader ecological coverage and diversity of vegetation conditions rather than full multi-view TLS completeness, aligning with the objective of training a generative model that generalizes across landscape heterogeneity.
Scans were taken from static tripod positions, with 240 scans randomly distributed across a 2.5 km diameter area. Each TLS scan contained approximately 8 million points, with around 4.6 million from the ground. Since most returns were concentrated within a 25 meter range, points beyond this threshold were removed to mitigate the non-uniform point density. A representative sampling pattern of TLS scans across a region is shown in Fig. \ref{fig:broad_als_local_tls}, where each TLS scan is represented by a color-coded blob.
%
ALS data was collected using a Galaxy T2000 LiDAR sensor mounted on a fixed-wing aircraft. The sensor operates at a 1064 nm wavelength and captures up to eight returns per pulse, with a 28° FOV, flown at an altitude of approximately 1,675 meters. The system achieves a point density of approximately 15 points per square meter and maintains an accuracy of 0.03 to 0.25 meters RMSE within a 150–6,500 meter range. The surveyed area was covered using multiple overlapping flight lines, including seven cross strips to reduce point sparsity and enhance coverage. Unlike TLS, ALS points are not affected by satellite signal occlusion and were globally aligned using a Trimble CenterPoint RTX RTK system for centimeter-level accuracy. A regional scale ALS point cloud is shown in Fig. \ref{fig:broad_als_local_tls} in yellow, with orthographic (top-down) views on the left and middle and a side-view on the top right.

Each TLS scan was initially positioned using GPS-based estimates in global coordinates. These positions were used to extract a corresponding 50 meter radius ALS pointcloud subset around each TLS scan center. The automatic and targetless co-registration approach from \cite{castorena2023automated} was applied to refine the position provided by GPS while also aligning the TLS scan to the ALS point-cloud relative to rotations. 
Each co-registered pair was manually verified by an expert to ensure qualitative alignment between spatial features.
Tree detection was performed using the deep learning (DL) based method from \cite{windrim2019forest}, which detects trees or tightly clustered groups of trees/vegetation by fitting 2D bounding boxes from a top-down bird’s-eye view and then expanding each bounding box to 3D informed by the lowest-to-highest point elevation difference. This approach ensures that the resulting 3D bounding boxes capture the full vertical extent of vegetation, from the top of the canopy to the lowest ground point. These bounding boxes were used to extract tree-level point clouds by truncating the points in the ALS and TLS scans inside each 3D box.
To mitigate the non-uniform point density in TLS, we truncated each point cloud to a 25m radius and downsampled each TLS scan to a 0.10 cm voxel size. Each tree scan was then centered and normalized within the $[0,1]^3$ range.
The dataset was manually curated to remove incomplete trees (e.g., those significantly cut by bounding boxes), and incorrect detections. The final dataset consists of 2,900 TLS/ALS tree or tree-cluster pairs. Fig. \ref{fig:dataset} illustrates nine ALS/TLS example pairs, with ALS trees in yellow (top row) and corresponding TLS scans in aqua (bottom row). 
%
For validation, we construct a set that consists of 300 trees sampled from plots that are geographically distinct from those used for training and testing. These trees span the same range of mixed-conifer vegetation types but originate from separate ALS/TLS co-registered plots, ensuring that the model is evaluated on structurally similar but spatially independent examples. The validation set is used exclusively for hyperparameter tuning, monitoring reconstruction error during training, and selecting the best checkpoint under an early stopping criterion. Note that the test set is held out entirely until final evaluation, and no samples in the validation set appear in the training or test splits.

\pichskip{10pt}
\parpic[r][b]{%
    \begin{minipage}{65mm}
        \includegraphics[width=\linewidth]{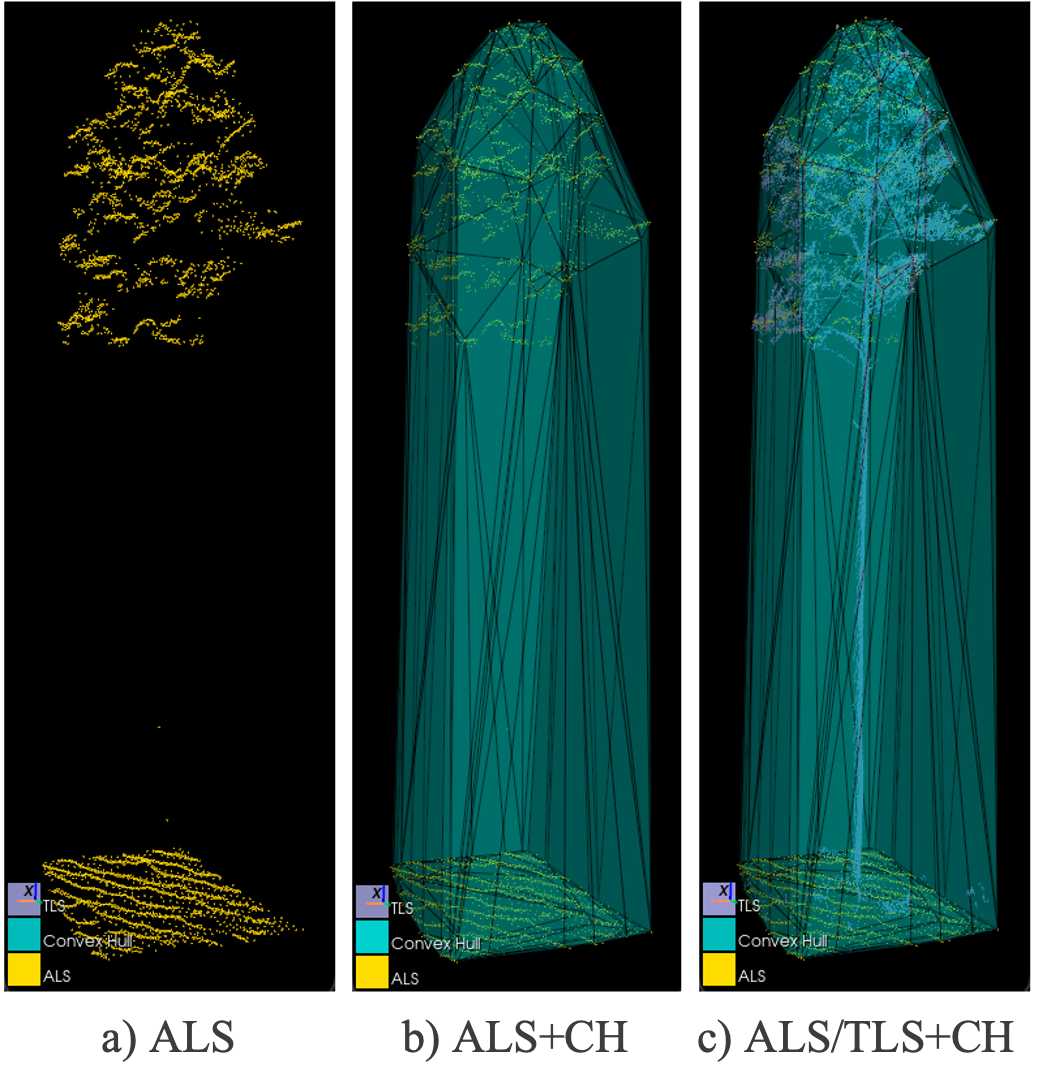}%
        \captionsetup{hypcap=false}
        \captionof{figure}{
        Empirical geometric relationship between ALS and TLS point clouds. ALS (in yellow) captures upper-canopy geometry, forming a sparse convex envelope (in green), while TLS points (in purple) representing stems, branches, and understory are largely contained within this envelope. This empirical property motivates the containment assumption exploited in this work when learning cross-domain mappings between ALS and TLS data.
        }
        \label{fig:containment}
    \end{minipage}
}

Due to the physical operation principles of LiDAR, TLS and ALS yield structurally complementary but distributionally distinct 3D point cloud characteristics. ALS captures forest structure from an aerial, top-down perspective and provides relatively complete sampling of the upper canopy features, while TLS operates from the ground and more effectively resolves stems, understory and internal structure. This leads to a broader spatial support for ALS compared to TLS.
Let $\X \in \Rspace^{N \times 3}$ and $\Y \in \Rspace^{M \times 3}$ denote the TLS and ALS point clouds for an individual tree (or small group of trees), and let $\mathcal{B}_{\text{ALS}} = \text{Conv}(\Y)$ be the convex hull (i.e., convex envelope) of the ALS data.
\textit{Assumption:} We adopt a localized containment assumption at the individual tree scale, where TLS points are expected to lie within the convex envelope of their corresponding co-registered ALS points with high probability (see Figure \ref{fig:containment}). This is expressed as:
\begin{equation} \label{epc}
    \mathbb{E}_{(\X, \Y) \sim \mathcal{D}} \left [ \mathbb{E}_{\x \sim \X} [1( \x \in \mathcal{B}_{\text{ALS}})] \right ] \geq 1-\epsilon
\end{equation}
 for a small constant $\epsilon \ll 1$, with $(\X, \Y) \sim \mathcal{D}$ drawn from the dataset of co-registered tree level TLS/ALS pairs.  Here, $\x \sim \X$ are 3D points uniformly sampled from the TLS point cloud, and $1(\cdot)$ is the indicator function. In our empirical analysis of the training \textit{CoLIDAR-Forest3D (ALS/TLS)} dataset, approximately 3.9\% of TLS points fall outside the ALS convex hull, i.e., $\epsilon=0.04$, validating the assumption at the tree scale. We emphasize that this prior is not assumed to hold globally at the plot or landscape scale and that a variety of factors may violate containment.
However, the small deviations from containment that may be locally present are expected and well justified, as they can result from sensor misalignment, resolution differences, vegetation visible from the ground but occluded from the aerial view, and inherent measurement noise in both ALS and TLS systems. Despite these localized discrepancies, the overall containment expectation introduces a geometric prior that is implicitly learned by our conditional generative model. This prior not only helps constrain synthesis within plausible ALS-guided structure, but also provides a useful means of measuring the spatial fidelity of the generated points relative to the ALS envelope.

\subsection{3D Cross-domain Denoising Diffusion Generative Models} \label{Ssec:generation}

Denoising diffusion probabilistic models are generative deep neural networks that operate within a variational inference framework \cite{ho2020denoising}. Their primary objective is to learn an approximation of the underlying training data distribution while enabling flexible and efficient sample generation from this learned distribution. In our context, we adopt DDPMs in a \textit{cross-domain} setting, where the goal is to generate TLS-like 3D vegetation structure conditioned on co-registered ALS measurements. This enables the synthesis of dense sub-canopy and near-ground details in regions where ALS data is sparse or occluded.
To this end, we train the model to approximate the conditional distribution \( p(\X \mid \Y) \), where ALS point clouds represent sparse, canopy dominated measurements, and TLS point clouds provide detailed ground-to-canopy interior structural information. During inference, the model begins from isotropic Gaussian noise and iteratively refines the sample using structural cues provided by ALS input, producing 3D point clouds that resemble high-resolution TLS samples respecting ALS fidelity.

\begin{figure}[ht]
    \centering
    \includegraphics[width=0.80\linewidth]{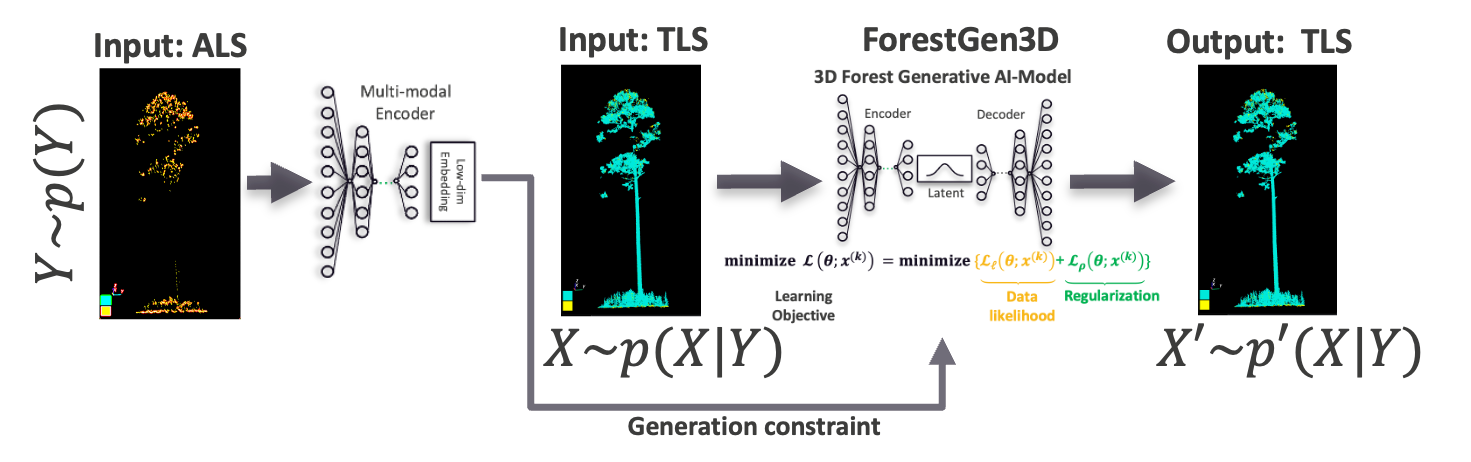}
    \caption{Schematic of ALS-to-TLS cross-domain ForestGen3D. The forward process encodes a TLS sample into Gaussian noise, while the reverse process learns to denoise a sample conditioned on ALS input following a denoising diffusion probabilistic model (DDPM) formulation, reconstructing high-resolution 3D vegetation structure.}
    \label{fig:schematic}
\end{figure}

The DDPM architecture consists of two components: a forward diffusion process that progressively corrupts the input point cloud \( \X^0 \in \mathbb{R}^{N \times 3} \) over \( T \) steps by multiplying it with a decaying factor while adding Gaussian noise, and a learned reverse process that denoises the corrupted sample back to a clean reconstruction. This yields a training trajectory:
$\X^0 \rightarrow \X^1 \rightarrow \cdots \rightarrow \X^T \sim \mathcal{N}(0, \mathbf{I})$, as the forward (diffusion) process,  $ \X^T \rightarrow \X^{T-1} \rightarrow \cdots \rightarrow \hat{\X}^0$ as the reverse (denoising) process.
The model conditions each denoising step on ALS observations \( \Y \), enabling it to learn structure consistent with the canopy geometry while generating plausible TLS content beneath.
The generative backbone used in ForestGen3D corresponds to the Conditional Generation Network (CGNet) architecture in \cite{lyu2021conditional}, which enables us to condition denoising based on ALS point cloud measurements. CGNet consists of three principal components: a PointNet++ based denoising network \cite{qi2017pointnet}, a ConditionNet that encodes ALS structure, and a FeatureMapper that fuses both representations. The denoising network follows a hierarchical, U-Net architecture that extracts features from noisy point clouds using multi-scale grouping and radius based neighborhood search. Four hierarchical levels are used, employing radii of 0.1, 0.2, 0.4, and 0.8~m with 32 neighbors sampled at each scale. Feature dimensions progress through $\{32, 64, 128, 256, 512\}$, and residual multi layer perceptron (MLP) blocks with skip connections are used to improve stability across diffusion steps. The diffusion timestep is embedded using a 128 dimensional positional encoding and concatenated to point level features.
ALS-based conditioning is provided through the ConditionNet, which mirrors the hierarchical structure of the denoising network and extracts both local and global structural features from the ALS point cloud. Its feature dimensions follow $\{32, 32, 64, 64, 128\}$, enabling the network to capture the vertical distribution patterns, and scan sparsity. The resulting ALS features are aligned with the denoising hierarchy and supplied to the FeatureMapper, which integrates noisy point cloud features and ALS information using FiLM-style modulation \cite{perez2018film}. This module applies shared MLP feature transforms, radius-based grouping, and an attention mechanism with grouped feature transformations and batch normalized activations. Together, these components ensure that the reverse diffusion process is guided both by ALS derived structure and by the learned statistical distribution of TLS geometry.
A high-level schematic of the model architecture is shown in Fig.~\ref{fig:schematic}.

The model is trained on point clouds \( \X,\Y  \sim p_{data}(\X, \Y) \in \mathbb{R}^{N \times 3} \times \mathbb{R}^{M \times 3} \), sampled from aligned TLS/ALS pairs. The forward diffusion process follows a first-order Markov chain by adding Gaussian noise over \( T \) steps (e.g., $T=1000$. The reverse process learns to approximate \( p(\X^{t-1} \mid \X^t, \Y) \), thereby denoising the latent variable back to a plausible TLS point cloud. Figure~\ref{fig:denoising} illustrates the generative process in a leave-one-out setting, where a withheld TLS scan is reconstructed using only its associated ALS input. The figure shows the denoising trajectory starting from an initial sample \( \X^T \sim \mathcal{N}(0, \mathbf{I}) \) and progressively refining toward a tree structure that closely approximates the held-out TLS reference. Notably, the model is able to synthesize sub-canopy features that are missing from the ALS input, highlighting the DDPM's capacity to infer occluded structural details.

\begin{remark}[Containment Consistency under ELBO] \label{remark:containment}
Because the conditional DDPM is trained via ELBO minimization \cite{ho2020denoising} to approximate the true conditional distribution \( p_{\text{data}}(\X \mid \Y) \), and the training data satisfies an approximate containment condition at the tree scale, we expect the learned generative model \( p_\theta(\X \mid \Y) \) to inherit this spatial bias. 

Formally, we can bound the expected point containment (EPC) for tree-level generation as:
\begin{equation} \label{elbo_containment}
    \mathbb{E}_{\Y} \left[ \mathbb{E}_{\X \sim p_\theta(\cdot \mid \Y)} \left[ \mathbb{E}_{\x \sim \X} \left[ 1(\x \in \mathcal{B}_{\text{ALS}}(\Y)) \right] \right] \right] \geq 1 - \epsilon - \delta
\end{equation}
where \( \epsilon \) reflects the empirical containment gap (e.g., 3.9\% in our CoLiDAR Forest3D dataset), and \( \delta \) accounts for model approximation error due to ELBO. Since the convex hull is computed per tree (or tight tree cluster), the spatial bias is preserved during training and inference, and we observe in practice that containment violations remain rare and localized. A full proof of Eq.~\eqref{elbo_containment} is provided in the Appendix.
\end{remark}

\section{Experimentation}
\label{sec:experiments}

The experiments in this section are designed to address two key questions: (1) Can conditional DDPM models effectively learn the detailed 3D structural distribution of forest vegetation from TLS data with ALS guidance? and (2) Is the approach scalable to plot-scale and landscape-scale deployment, enabling accurate synthesis of tree-level structure and improvement of downstream ecological biometrics commonly used in forest management?.
The first set of experiments, presented in Section~\ref{Ssec:training_and_validation}, evaluates the generative capacity of DDPMs at the tree scale. We assess whether the model's inductive biases and architecture are well-suited for representing the hierarchical and self-similar structure of vegetation. This includes analysis of the model's approximation quality via reconstruction error, as well as ablation studies that explore the sensitivity of generation quality to key DDPM hyperparameters such as diffusion steps.
To quantitatively assess reconstruction error, we compute Chamfer Distance (CD) and Earth Mover’s Distance (EMD) \cite{fan2017point} between generated and reference TLS point clouds, using held-out tree samples. In addition, we examine the relationship between CD, EMD, and our expected point containment (EPC) metric . The goal of this analysis is to evaluate whether containment can serve as a practical proxy for CD and/or EMD in scenarios where ground-truth TLS data is unavailable.
The second set of experiments, detailed in Sections \ref{Ssec:treescale}-\ref{Ssec:regionalscale}, demonstrates the model’s generalization and scalability from tree to landscape scale scenarios in mixed conifer ecosystems. At the tree and plot scales, we evaluate the consistency of ALS-guided generation in matching spatial extents and density patterns observed in ground truth TLS scans. At the regional scale, we deploy the model across 200 meter radius ALS-only regions and assess the realism of synthesized vegetation. For both plot and landscape cases, we compute downstream ecological biometrics including tree height, diameter at breast height (DBH), crown diameter (CrD), and crown volume (CrV) based on ALS, TLS, ALS+TLS, and ALS+ForestGen3D samples. These comparisons highlight the model's ability to augment sparse ALS with realistic sub-canopy structure, leading to improved and more complete biometric estimates. Furthermore, we confirm that the containment prior is preserved in inference time landscape scale deployments, as evidenced by low out of envelope deviation statistics across diverse mixed-conifer landscapes.

\subsection{Implementation Details}

ForestGen3D is implemented in Python/PyTorch with CUDA acceleration. All experiments were conducted on an NVIDIA Tesla V100S-PCIE-32GB GPU, with a total training time of approximately 20 hours for 100 epochs. The diffusion process follows a standard DDPM formulation, using a linear noise schedule with $\beta_0 = 1\times10^{-4}$, $\beta_T = 2\times10^{-2}$, and $T=1000$ diffusion steps, although we also test other $T$ settings. The model is trained using a mean squared error (MSE) loss, while the Earth Mover’s Distance (EMD) and Chamfer distance metric is optionally monitored during validation to ensure that the learned denoising process maintains geometric stability. Absolute coordinates, local neighborhood features, global shape descriptors, and ALS conditioning are included throughout the denoising process. Grouped feature attention, batch normalization, and residual connections are enabled across all modules to stabilize training. Radius neighborhood search is used throughout both encoding and decoding, while a $k$-nearest neighbors (kNN) strategy with $k=8$ is employed for feature propagation during decoding. At inference, this configuration produces a complete 3D tree structure in approximately 0.4~seconds. To support reproducibility and reuse, the ForestGen3D model implementation, trained weights, and associated configuration files are expected to be released publicly upon publication of the manuscript.

\subsection{Training and Validation} \label{Ssec:training_and_validation}
\begin{figure}[ht]
    \begin{subfigure}{0.2645\linewidth}
	\includegraphics[height=11em,keepaspectratio]{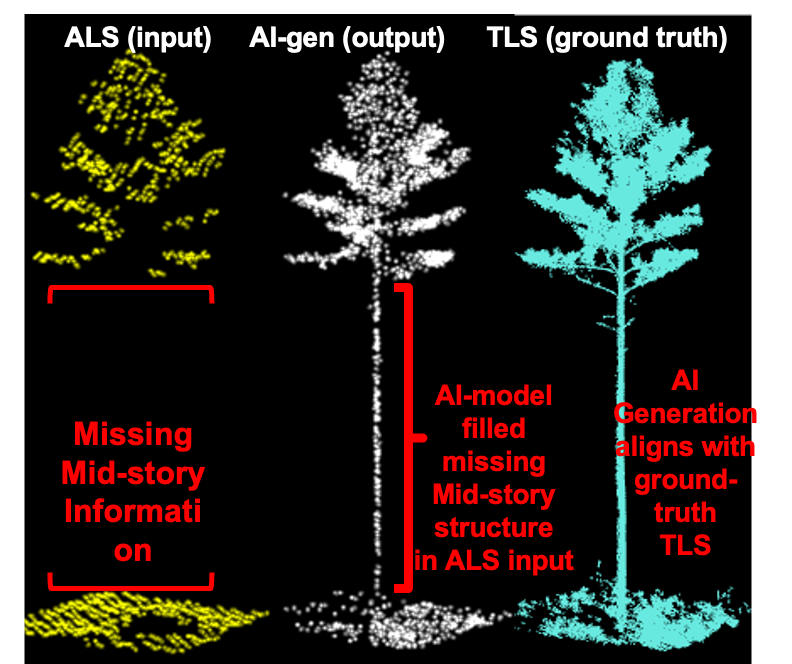}
	\caption{Example 1.}
	\label{fig:ex1_als2tls_train}
    \end{subfigure}%
    \begin{subfigure}{0.3\linewidth}
	\includegraphics[height=11.175em,keepaspectratio]{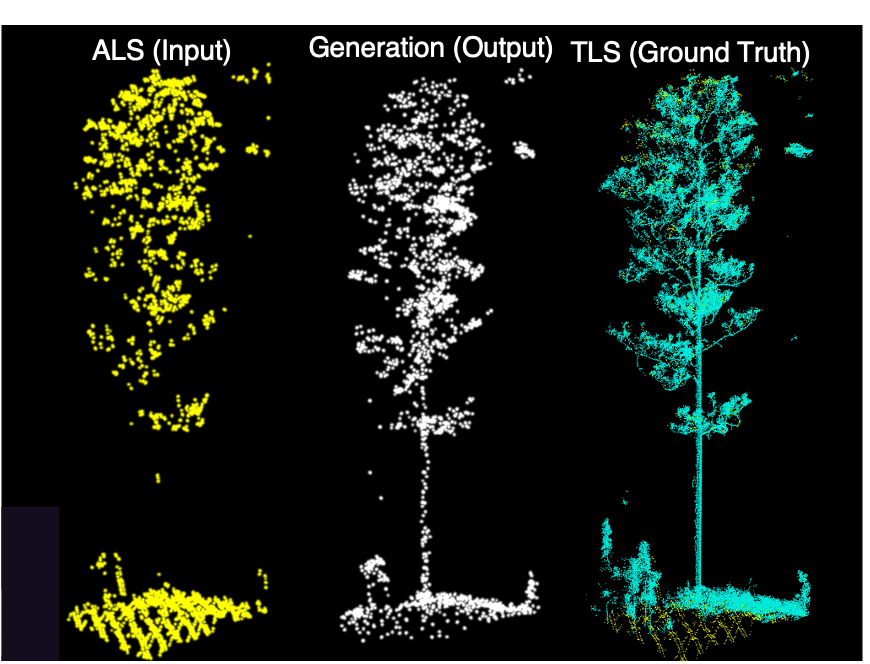}
	\caption{Example 2.}
	\label{fig:ex2_als2tls_train}
    \end{subfigure}
    %
    %
    \begin{subfigure}{0.40\linewidth}
	\includegraphics[height=11.08em,keepaspectratio]{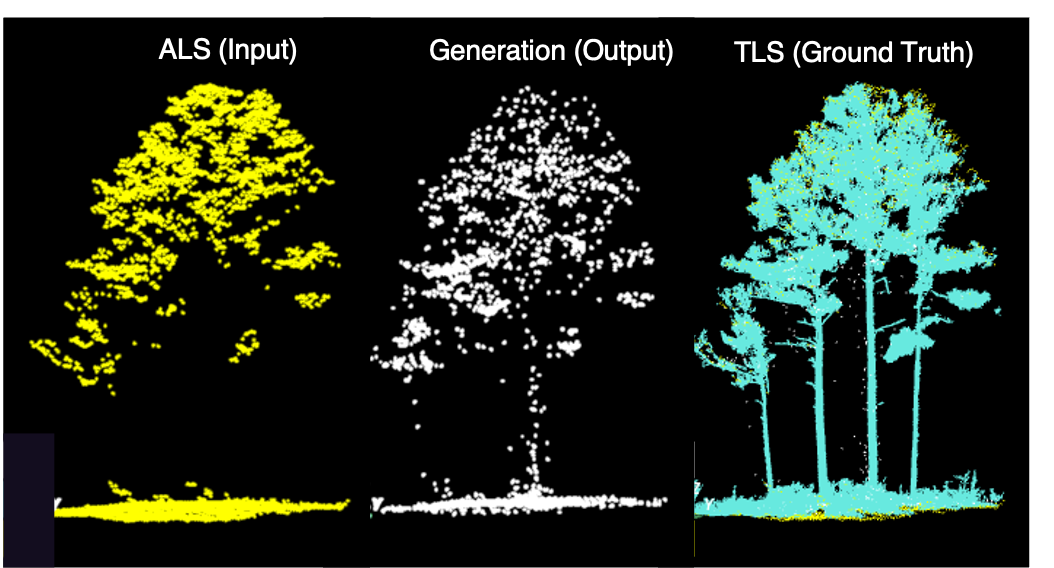}
	\caption{Example 3.}
	\label{fig:ex4_als2tls_train}
    \end{subfigure}
    \caption{Three examples of ForestGen3D generation at the  tree-scale conditioned by ALS inputs in the training set. 
    This qualitative examples showcase the architecture's capacity to encode and represent the 3D structural details of trees.}
    \label{fig:ALStoTLS_training}
\end{figure}

To evaluate the inductive biases and approximation fidelity of our DDPM-based generative framework, we conduct a series of ablation experiments. During training, each batch consists of 16 randomly selected ALS/TLS example pairs. Each point cloud is uniformly subsampled to a fixed size of $N = 2048$. This random subsampling enhances dataset diversity by introducing spatial variability across training epochs, enabling the model to encounter a broader range of structural configurations over time. Although the \( N = 2048 \) resolution may appear limiting, it applies only at the level of individual trees. When aggregated over larger spatial extents such as plots or landscapes, this per tree resolution naturally scales to yield detailed reconstructions across entire forested regions. Moreover, higher-resolution representations can be achieved through ensemble sampling, in which multiple stochastic generations are produced for the same conditional ALS input and merged coherently to create denser 3D tree structures. As an illustration of the capabilities of the generative model to approximate 3D tree structures given the existing settings, we include Figure \ref{fig:ALStoTLS_training} showcasing three examples in the training set with ALS input (first column), AI-generation (second column) and ground truth TLS (third column) in each example. Here, we see that our model is able to generate samples at good qualitative approximations of the 3D structural features of TLS distributions conditional on the ALS inputs across the entire vertical tree-stand direction. Figures \ref{fig:ex1_als2tls_train} and \ref{fig:ex2_als2tls_train} depict instances where the ALS input contains a single tree, which the model accurately approximates, demonstrating its ability to infer complex internal structures from sparse input. Figure \ref{fig:ex4_als2tls_train} further illustrates its performance in closely packed tree settings. Here, ground truth TLS shows four trees with tightly packed canopies, while the model generates fewer distinct trunks than the ground truth. Despite this, the output remains visually realistic and spatially consistent with what a human observer might infer from the incomplete ALS scan.

\begin{figure}[ht]
    \centering
            \begin{subfigure}{0.37\linewidth}
	       \centering
	       \includegraphics[width=1.0\linewidth]{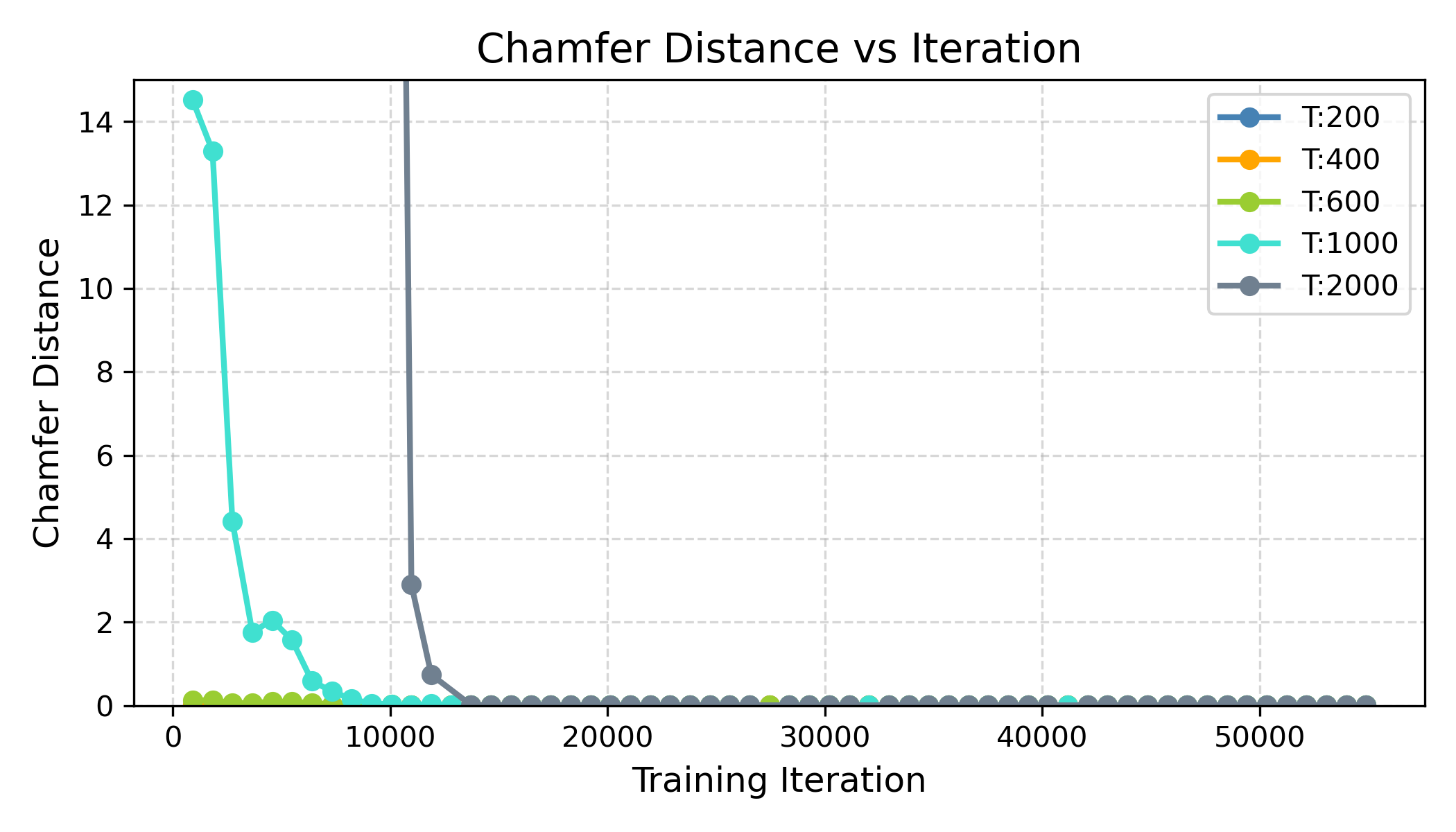}
	       \caption{Chamfer Distance (CD)}
	       \label{fig:chamfer}
            \end{subfigure} 
            \begin{subfigure}{0.37\linewidth}
	       \includegraphics[width=1.0\linewidth]{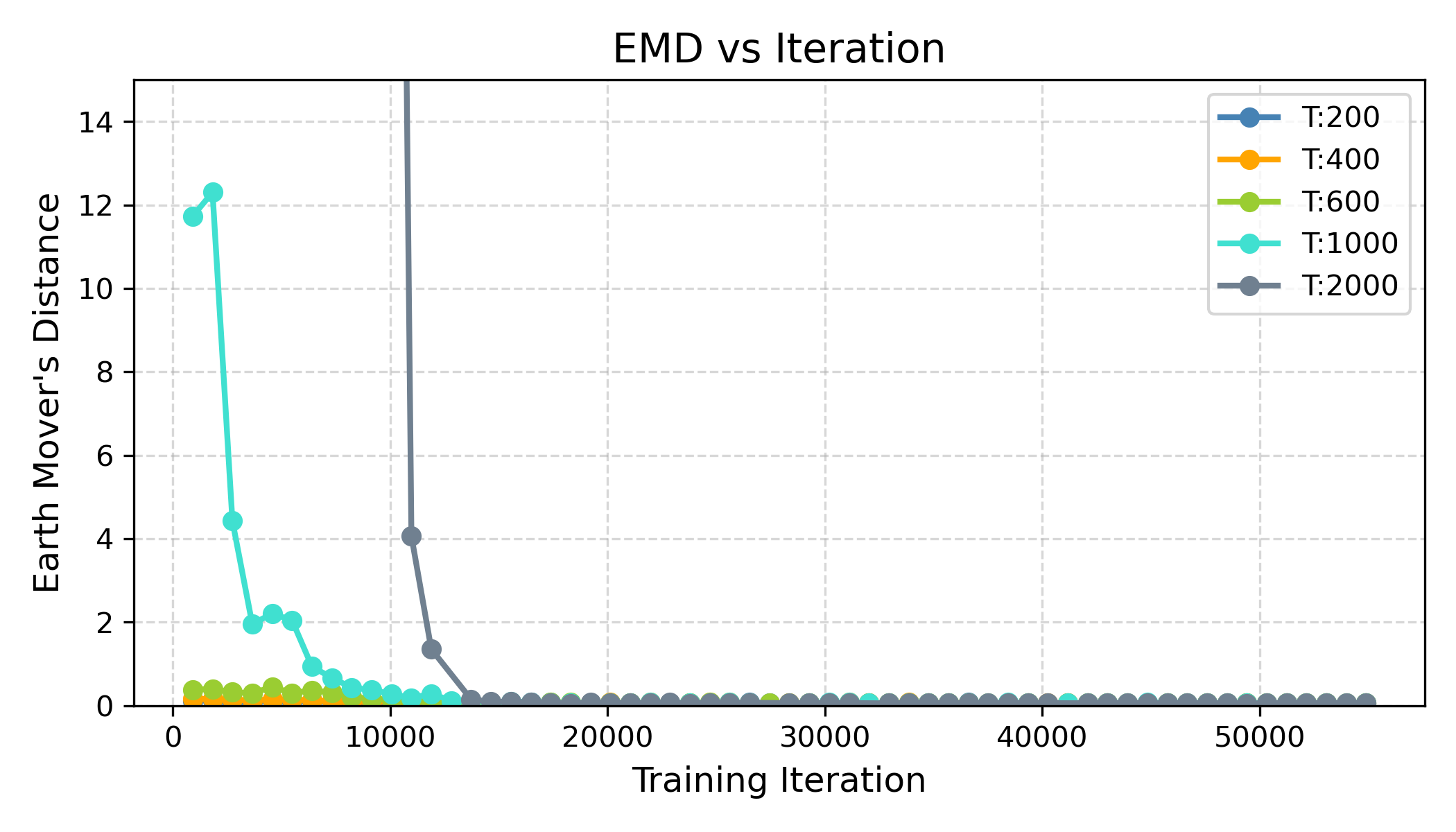}
	       \caption{Earth Movers Distance (EMD)}
	       \label{fig:emd}
            \end{subfigure}
    \caption{ Chamfer Distance (CD) and Earth's movers distance (EMD) errors in the validation set as a function of training iterations. Both metrics consistently decrease with training iterations across all diffusion schedules.}
    \label{fig:hyperparameter}
\end{figure}

Ablation studies to analyze model reconstruction performance where conducted by varying the number of diffusion steps $T \in \{$200, 400, 600, 1000, 2000$\}$, diffusion parameters \( \beta_0 = 10^{-4} \), \( \beta_T = 0.02 \), and batch size 16 constant. These experiments aim to analyze the error of the model at training time and also the spatial containment consistency in eq.\eqref{elbo_containment}. We track reconstruction error across training iterations using both Chamfer Distance (CD) and Earth Mover’s Distance (EMD), computed between generated and ground truth TLS point clouds on a held-out validation set of 300 tree examples. As shown in Figs.~\ref{fig:chamfer}-\ref{fig:emd}, both CD and EMD consistently decrease with training iterations across all diffusion schedules, indicating convergence and improving fidelity. Notably, models with higher diffusion steps (e.g., $T=1000$) exhibit slower convergence but ultimately achieve slightly lower reconstruction error, while smaller $T$ values converge quickly with slightly higher error floors.

\begin{figure}[ht]
    \centering            
            \begin{subfigure}{0.33\linewidth}
	       \includegraphics[width=1.0\linewidth]{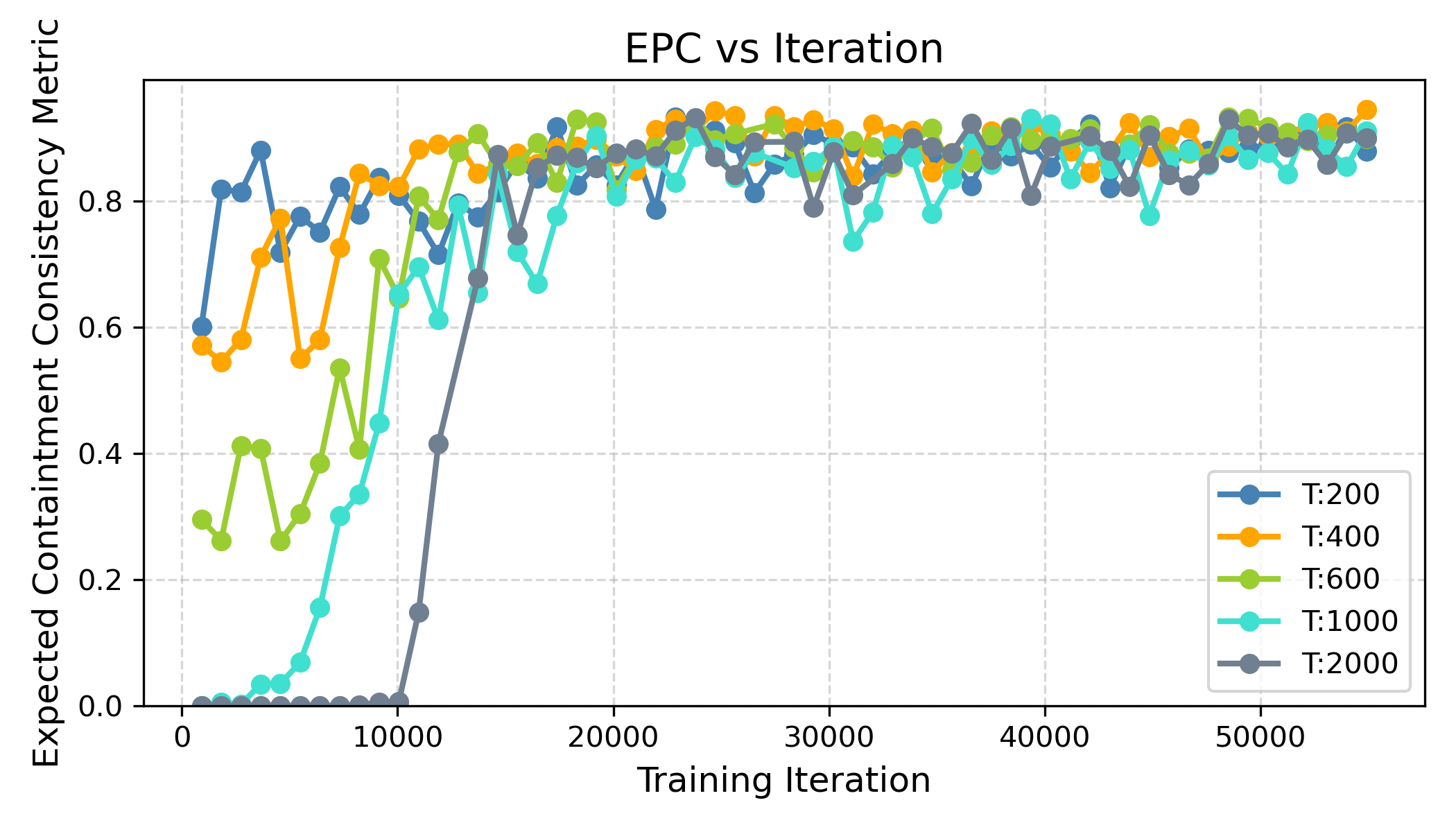}
	       \caption{Expected Point Containment (EPC)}
	       \label{fig:epc}
            \end{subfigure} 
            \begin{subfigure}{0.33\linewidth}
	       \centering
	       \includegraphics[width=1.0\linewidth]{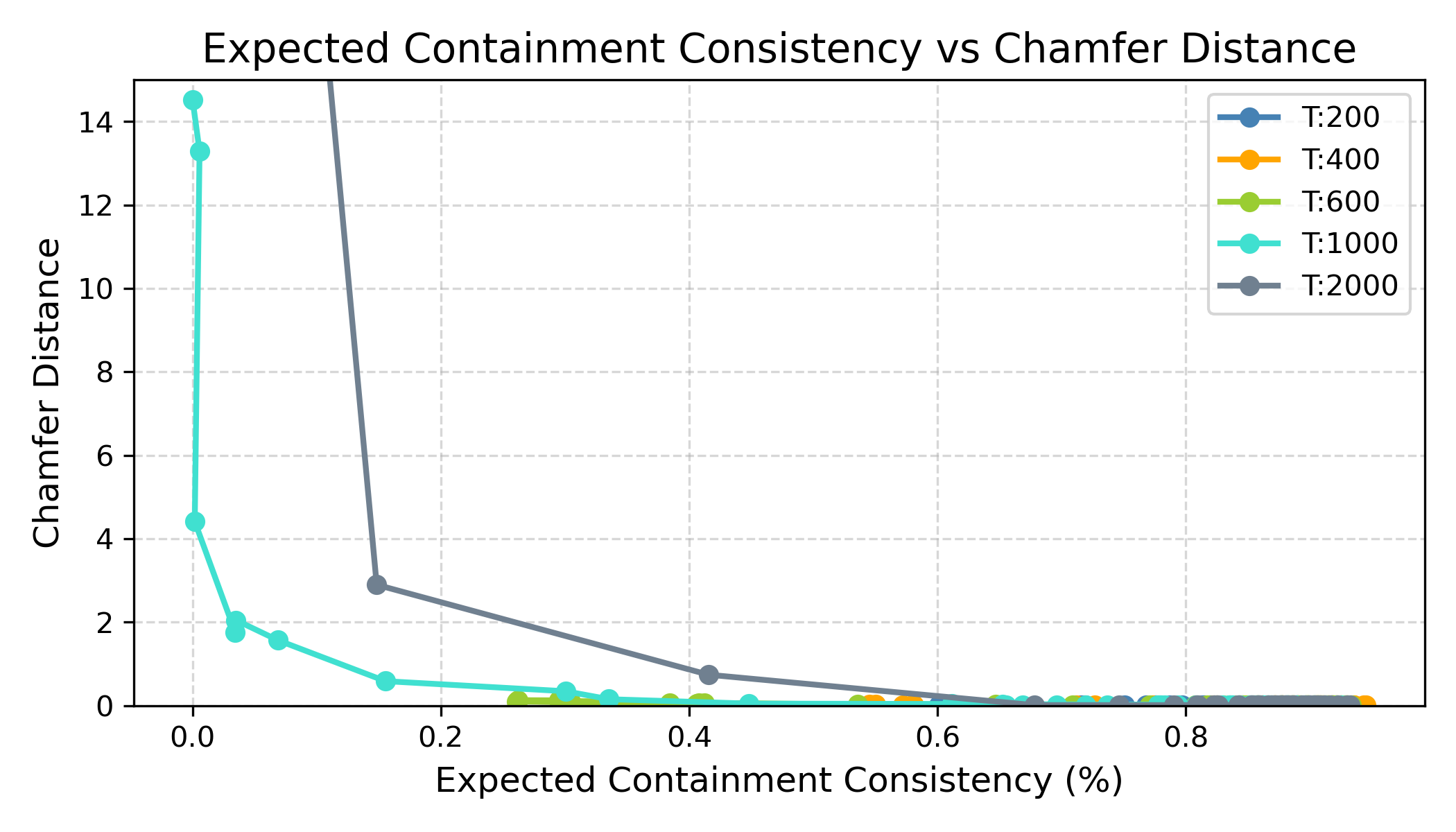}
	       \caption{Chamfer Distance (CD) versus EPC}
	       \label{fig:chamfer_epc}
            \end{subfigure} 
            \begin{subfigure}{0.33\linewidth}
	       \includegraphics[width=1.0\linewidth]{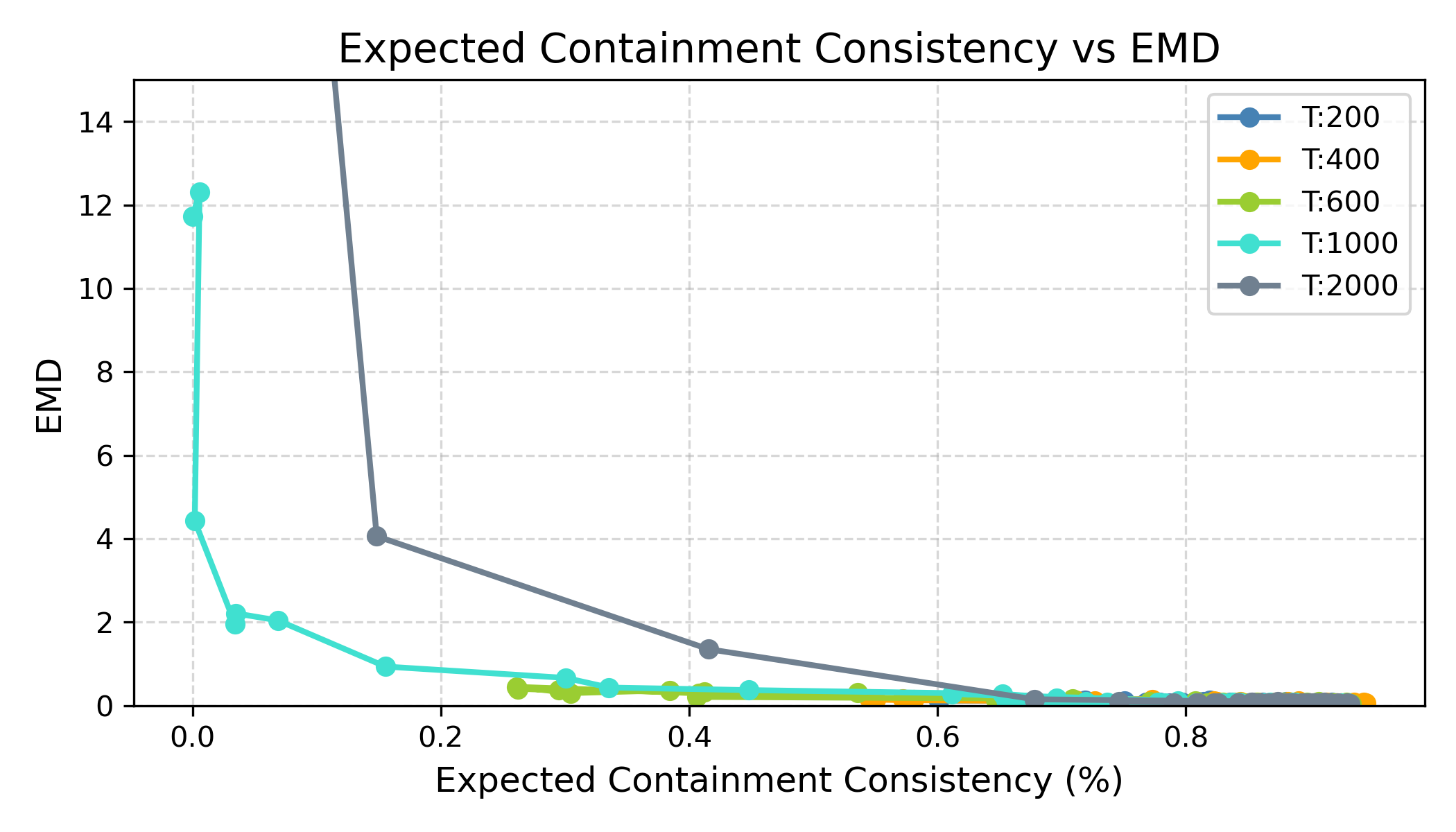}
	       \caption{Earth Movers Distance (EMD) versus EPC}
	       \label{fig:emd_epc}
            \end{subfigure}
    \caption{Relationship between containment behavior measured by expected point containment (EPC) and reconstruction error metrics in validation set. There is a strong inverse correlation between EPC and error metrics acrross all denoising time-step schedules $T$: as EPC increases, both Chamfer Distance (CD) and Earth's movers distance (EMD) rapidly decrease.}
    \label{fig:epc_plots}
\end{figure}

In parallel, we assess the containment behavior of generated point clouds using the Expected Point Containment metric, which quantifies the proportion of generated points lying within the ALS convex envelope. Fig.\ref{fig:epc} shows that EPC over training iterations mirror reconstruction improvement, suggesting that containment behavior improves with model learning. To investigate the link between spatial consistency and reconstruction fidelity, we plot CD and EMD directly against EPC in Figs.~\ref{fig:chamfer_epc} and \ref{fig:emd_epc}, respectively. Across all diffusion schedules, we observe a strong inverse correlation: as EPC increases, both CD and EMD rapidly decrease, ultimately plateauing at low error values for high containment ratios.
This relationship reveals an important insight: when generated trees remain well contained within the convex hull of ALS inputs, as measured by EPC, this behavior serves as a meaningful indicator of generative quality. In particular, high EPC values consistently appear alongside low reconstruction errors (e.g., CD, EMD), suggesting that EPC can serve as a practical proxy for generation quality in scenarios where ground-truth is unavailable. 
Moreover, the observed alignment between containment behavior and reconstruction fidelity empirically supports the theoretical guarantee presented in Remark~\ref{remark:containment}, demonstrating that the learned conditional distribution \( p_{\theta}(\X|\Y) \) inherits the spatial containment property embedded in the training data through ELBO minimization. These findings not only validate our containment assumption, but also highlight its practical utility for model evaluation in operational forest structure generation pipelines lacking TLS ground truth.

\subsection{Testing dataset} \label{Ssec:test_dataset}
For testing, we used 50 TLS scans collected from geographic regions that do not overlap with those of the training and validation areas, though sharing similar ecosystem characteristics and collected with the same sensors used for the training data. We evaluated the model’s performance across three spatial scales. At the tree scale, we followed the same co-registration and tree detection procedures used during training, resulting in 1,457 individual test trees. This setup enabled direct, one-to-one comparisons between generated samples and corresponding TLS ground truth reference data at high structural fidelity. At the plot scale, we constructed test plots by aggregating ALS data over spatial extents corresponding to each TLS scan, truncating ALS coverage to match the TLS field of view (e.g., 25 meters in radius ). This allowed for evaluation of the model’s ability to reconstruct localized forest structure. Finally, at the regional scale, we assessed scalability and extrapolation by deploying the model over multiple 200 meter radius areas containing ALS data. These larger, fully independent areas tested the model's ability to synthesize realistic vegetation structure across broader, ALS only landscapes. This evaluation strategy enables comprehensive assessment of ForestGen3D's robustness and flexibility from fine-grained, tree-level detail to broad-scale generation in mixed conifer ecosystems.

%
\subsection{Tree-Scale ALS-Guided Generation} \label{Ssec:treescale}
\begin{figure}

    \centering
    \begin{subfigure}{0.13\linewidth}
	\centering
	\includegraphics[height=14.5em, keepaspectratio]{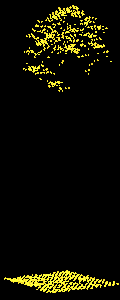}
	\caption{ALS \\ Input \\ \hspace{1em} \\ \hspace{1em}}
	\label{fig:ex1_1}
    \end{subfigure}%
    \begin{subfigure}{0.13\linewidth}
	\includegraphics[height=14.5em, keepaspectratio]{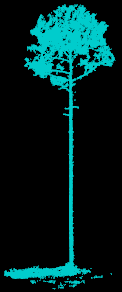}
	\caption{TLS \\ Ground Truth \\ \hspace{1em} \\ \hspace{1em}}
	\label{fig:ex1_2}
    \end{subfigure}
    \begin{subfigure}{0.13\linewidth}
	\includegraphics[height=14.5em, keepaspectratio]{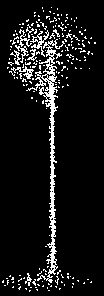}
	\caption{PCN \cite{yuan2018pcn} \\ CD: 7.91e-3 \\ EMD: 7.069e-2\\ EPC: 0.794 }
	\label{fig:ex1_3}
    \end{subfigure}
    \begin{subfigure}{0.13\linewidth}
	\includegraphics[height=14.5em, keepaspectratio]{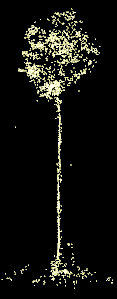}
	\caption{3D-GAN \cite{wu2016learning} \\ CD: 3.39e-3 \\ EMD: 4.549e-2\\ EPC: 0.939}
	\label{fig:ex1_4}
    \end{subfigure}
    \begin{subfigure}{0.13\linewidth}
	\includegraphics[height=14.5em, keepaspectratio]{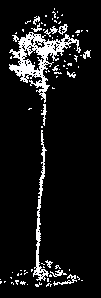}
	\caption{\textit{l}-GAN \cite{achlioptas2018learning} \\ CD: 3.12e-3 \\ EMD: 5.416e-2\\ EPC: 0.961}
	\label{fig:ex1_5}
    \end{subfigure}
    \begin{subfigure}{0.13\linewidth}
	\includegraphics[height=14.5em, keepaspectratio]{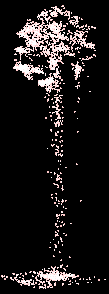}
	\caption{PointFlow \cite{yang2019pointflow} \\ CD: 4.41e-3 \\ EMD: 7.204e-2\\ EPC: 0.948}
	\label{fig:ex1_6}
    \end{subfigure}
    \begin{subfigure}{0.13\linewidth}
	\includegraphics[height=14.5em, keepaspectratio]{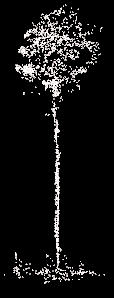}
	\caption{ForestGen3D {\tiny{(ours)}} \\ CD: 1.91e-3 \\ EMD: 4.88e-2\\ EPC: 0.956}
	\label{fig:ex1_7}
    \end{subfigure}

    \begin{subfigure}{0.13\linewidth}
	\centering
	\includegraphics[height=15em, keepaspectratio]{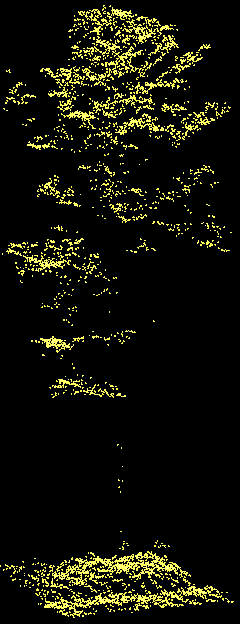}
	\caption{ALS \\ Input \\ \hspace{1em} \\ \hspace{1em}}
	\label{fig:ex2_1}
    \end{subfigure}%
    \begin{subfigure}{0.13\linewidth}
	\includegraphics[height=15em, keepaspectratio]{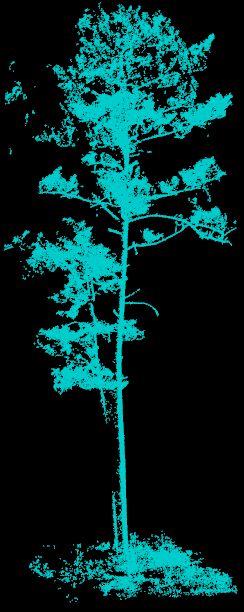}
	\caption{TLS \\ Ground Truth \\ \hspace{1em} \\ \hspace{1em}}
	\label{fig:ex2_2}
    \end{subfigure}
    \begin{subfigure}{0.13\linewidth}
	\includegraphics[height=15em, keepaspectratio]{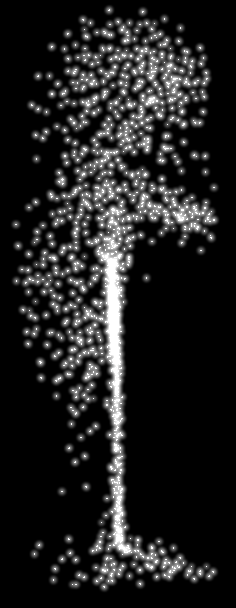}
	\caption{PCN \cite{yuan2018pcn} \\ CD: 4.024e-3 \\ EMD: 3.837e-2\\ EPC: 0.933 }
	\label{fig:ex2_3}
    \end{subfigure}
    \begin{subfigure}{0.13\linewidth}
	\includegraphics[height=15em, keepaspectratio]{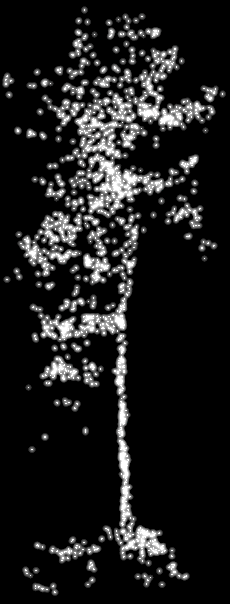}
	\caption{3D-GAN \cite{wu2016learning} \\ CD: 3.266e-3 \\ EMD: 4.657e-2\\ EPC: 0.943}
	\label{fig:ex2_4}
    \end{subfigure}
    \begin{subfigure}{0.13\linewidth}
	\includegraphics[height=15em, keepaspectratio]{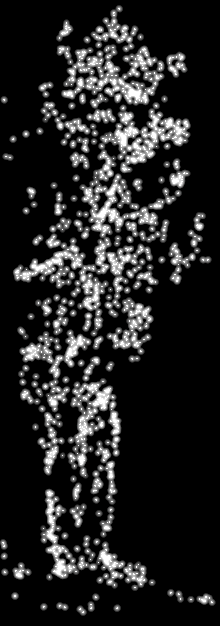}
	\caption{\textit{l}-GAN \cite{achlioptas2018learning} \\ CD: 4.126e-3 \\ EMD: 5.015e-2\\ EPC: 0.944}
	\label{fig:ex2_5}
    \end{subfigure}
    \begin{subfigure}{0.13\linewidth}
	\includegraphics[height=15em, keepaspectratio]{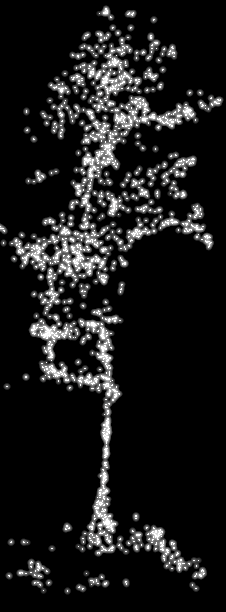}
	\caption{PointFlow \cite{yang2019pointflow} \\ CD: 3.451e-3 \\ EMD: 6.708e-2\\ EPC: 0.964}
	\label{fig:ex2_6}
    \end{subfigure}
    \begin{subfigure}{0.13\linewidth}
	\includegraphics[height=15em, keepaspectratio]{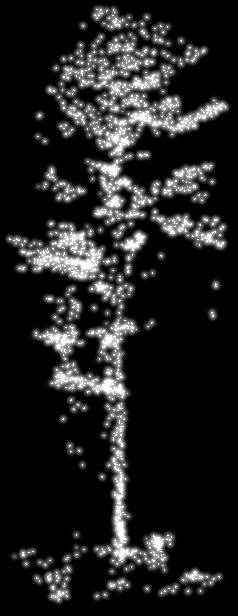}
	\caption{ForestGen3D {\tiny{(ours)}} \\ CD: 3.33e-3 \\ EMD: 4.407e-2\\ EPC: 0.962}
	\label{fig:ex2_7}
    \end{subfigure}

    \begin{subfigure}{0.13\linewidth}
	\includegraphics[height=11.em, keepaspectratio]{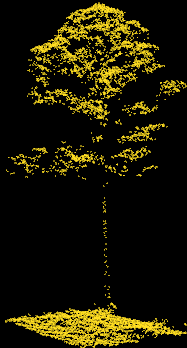}
	\caption{ALS \\ Input \\ \hspace{1em} \\ \hspace{1em}}
	\label{fig:ex3_1}
    \end{subfigure}%
    \begin{subfigure}{0.13\linewidth}
	\includegraphics[height=11.em, keepaspectratio]{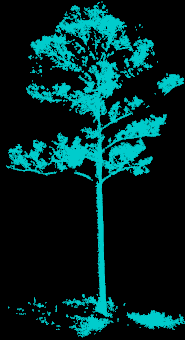}
	\caption{TLS \\ Ground Truth \\ \hspace{1em} \\ \hspace{1em}}
	\label{fig:ex3_2}
    \end{subfigure}
    \begin{subfigure}{0.13\linewidth}
	\includegraphics[height=11.em, keepaspectratio]{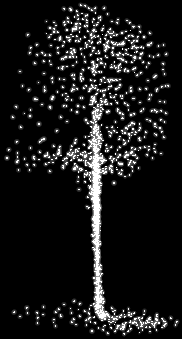}
	\caption{PCN \cite{yuan2018pcn} \\ CD: 11.11e-3 \\ EMD: 8.641e-2\\ EPC: 0.82 }
	\label{fig:ex3_3}
    \end{subfigure}
    \begin{subfigure}{0.13\linewidth}
	\includegraphics[height=11.em, keepaspectratio]{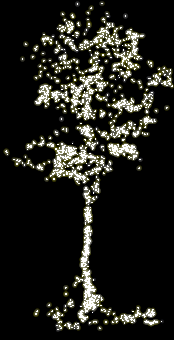}
	\caption{3D-GAN \cite{wu2016learning} \\ CD: 3.22e-3 \\ EMD: 4.493e-2\\ EPC: 0.936}
	\label{fig:ex3_4}
    \end{subfigure}
    \begin{subfigure}{0.13\linewidth}
	\includegraphics[height=11.em, keepaspectratio]{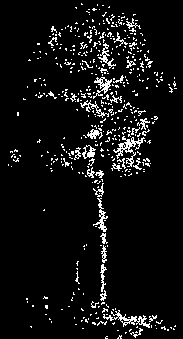}
	\caption{\textit{l}-GAN \cite{achlioptas2018learning} \\ CD: 3.554e-3 \\ EMD: 5.129e-2\\ EPC: 0.94}
	\label{fig:ex3_5}
    \end{subfigure}
    \begin{subfigure}{0.13\linewidth}
	\includegraphics[height=11.em, keepaspectratio]{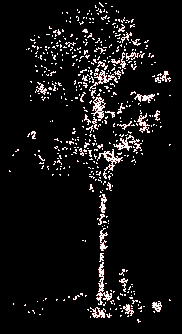}
	\caption{PointFlow \cite{yang2019pointflow} \\ CD: 3.476e-3 \\ EMD: 6.564e-2\\ EPC: 0.962}
	\label{fig:ex3_6}
    \end{subfigure}
    \begin{subfigure}{0.13\linewidth}
	\includegraphics[height=11.em, keepaspectratio]{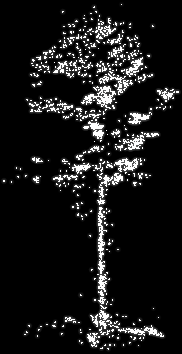}
	\caption{ForestGen3D {\tiny{(ours)}} \\ CD: 3.141e-3 \\ EMD: 4.56e-2\\ EPC: 0.957}
	\label{fig:ex3_7}
    \end{subfigure}
     
    \caption{Qualitative and Quantitative comparison of ALS-conditioned tree structure generation across three representative examples. (a,h,o) ALS input and (b,i,p) TLS reference; (c,j,q) PCN, (d,k,r) 3D-GAN, (e,l,s) latent-GAN, (f,m,t) PointFlow, and (g,n,u) ForestGen3D (ours). PCN oversmooths crowns, producing coarse silhouettes with little fine-scale detail. The other baselines generate shapes that more closely resemble TLS references but with inconsistencies such as distorted trunks (latent-GAN in (l)) or more discontinuous stems (PointFlow in (f)). 3D-GAN and ForestGen3D yield more coherent reconstructions, with ForestGen3D showing smaller metric errors with respect to ground truth. Pairwise point cloud distance metrics included are Chamfer Distance (CD), Earth's movers distance (EMD) and Expected Point containment (EPC).}
    \label{fig:qualitative_baselines}
\end{figure}

In Fig. \ref{fig:qualitative_baselines} we include three representative qualitative comparisons from the test set to illustrate differences between completion/generation approaches across a range of tree structures. First two columns across all three cases represent ALS input and TLS ground truth, while the remaining columns correspond to the completion/generation resulting from the application of different trained models. In all three examples, PCN \cite{yuan2018pcn} tends to oversmooth crowns, recovering coarse silhouettes while missing high-frequency branching detail. The other baseline generative models (3D-GAN \cite{wu2016learning}, latent-GAN \cite{achlioptas2018learning},  and PointFlow \cite{yang2019pointflow}) produce shapes that appear closer to the TLS shape, yet are less consistent across examples in general, in comparison to those from our ForestGen3D model. For instance, in the second example, latent-GAN (l) yields a form whose stem does not resemble the straight trunk structure of the TLS reference. Similarly, PointFlow in (f) generates a discontinuous tree stem of larger a diameter compared to ground truth TLS. 3D-GAN and ForestGen3D produce consistently closer results to TLS with canopy structure close to that of ALS and coherent stem-to-crown transitions, recognizable branching patterns, and overall shapes that better match human expectations of realistic tree architecture. These qualitative outcomes mirror the quantitative CD and EMD error, and EPC consistency results below each figure, highlighting the capabilities of generative models to produce realistic structural properties of forests.
Note also, that the generated points are generally tightly packed and bounded by the ALS canopy points. This behavior stems from the inherent nature of the ALS training data, where the distribution of the ALS convex-hull is typically broader and encompasses the finer details captured by TLS. Consequently, the generated points do not generally escape the span of the ALS canopy diameter. From a statistical standpoint, as the generative approach aims to produce samples from a distribution that is largely encompassed by the ALS input, it is inherently constrained, making it less likely for the generated points to statistically escape the overall ALS convex envelope.

\begin{table}
    \begin{minipage}{1.0\linewidth} 
        \centering 
        \captionof{table}{Quantitative generation and completion errors evaluated on the test set. Metrics include Chamfer Distance (CD), Earth Mover’s Distance (EMD), and Expected Point Containment (EPC). CD and EMD quantify geometric similarity between generated point clouds and TLS references, while EPC measures spatial containment with the ALS convex envelope when TLS ground truth is unavailable. Lower CD and EMD values and higher EPC indicate better reconstruction quality and structural fidelity.
        } 
        \label{tab:table_distances} 
        \begin{tabular}{llll} 
            \toprule 
            & \multicolumn{3}{c}{\textbf{Distance Metrics}}\\
            \textbf{Method} & CD & EMD & EPC (\%) \\
            \midrule 
                PCN \cite{yuan2018pcn} & 4.63e-3 & 9.58e-2 & 0.921\\
                3D-GAN \cite{wu2016learning}  & 4.37e-3 & 5.761e-2 & 0.926\\
                \textit{l}-GAN \cite{achlioptas2018learning}  & 6.10e-3 & 5.59e-2 & 0.862\\  
                PointFlow \cite{yang2019pointflow}  & 4.52e-3 & 5.429e-2 & 0.939\\   
                ForestGen3D (Ours) & \textbf{4.30e-3} & \textbf{5.06e-2} & \textbf{0.945}\\
            \bottomrule 
        \end{tabular} 
    \end{minipage} 
\end{table}

Table \ref{tab:table_distances} shows a quantitative evaluation summarizing the CD/EMD and EPC metrics over the test dataset of 1457 test examples. This table summarizes the resulting metrics for the completion PCN \cite{yuan2018pcn} method and the generative models including 3D-GAN \cite{wu2016learning}, latent \textit{l}-GAN \cite{achlioptas2018learning}, PointFlow \cite{yang2019pointflow} and our ForestGen3D denoising diffusion based model. For reference, example CD/EMD/EPC values are also reported in Fig. \ref{fig:qualitative_baselines} for their corresponding examples.
Beyond the numerical values, the metrics highlight important differences between the approaches. PCN, while competitive in CD, performs worst in EMD, reflecting its bias toward capturing coarse, low frequency geometry at the expense of fine structural detail. Latent-GAN shows the largest CD, due to plausible over-compressing structural features in the latent layers, while 3D-GAN improves CD, it still remains above our model. PointFlow achieves stronger balance across CD and EMD, though it is also higher than the performance of our model. In contrast, ForestGen3D consistently achieves the lowest CD and EMD, while also attaining the highest EPC, indicating superior alignment with TLS reference structure and better recovery of fine-scale details. These results quantitatively demonstrate that ForestGen3D more effectively learns the underlying distribution of forest structure from ALS/TLS data, achieving overall improved performance compared to both completion/generation-based baselines.

\subsection{Plot-scale ALS-Guided Generation} \label{Ssec:plotscale}

\begin{figure}
    \centering
            \begin{subfigure}{0.27\linewidth}
	       \centering
	       \includegraphics[width=1.0\linewidth]{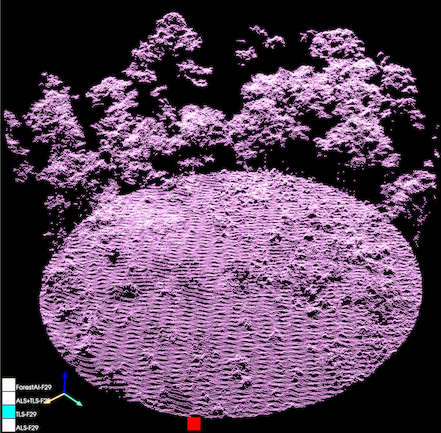}
	       \caption{Ex.1:ALS.}
	       \label{fig:ex1_alsf26}
            \end{subfigure} 
            \begin{subfigure}{0.27\linewidth}
	       \includegraphics[width=1.0\linewidth]{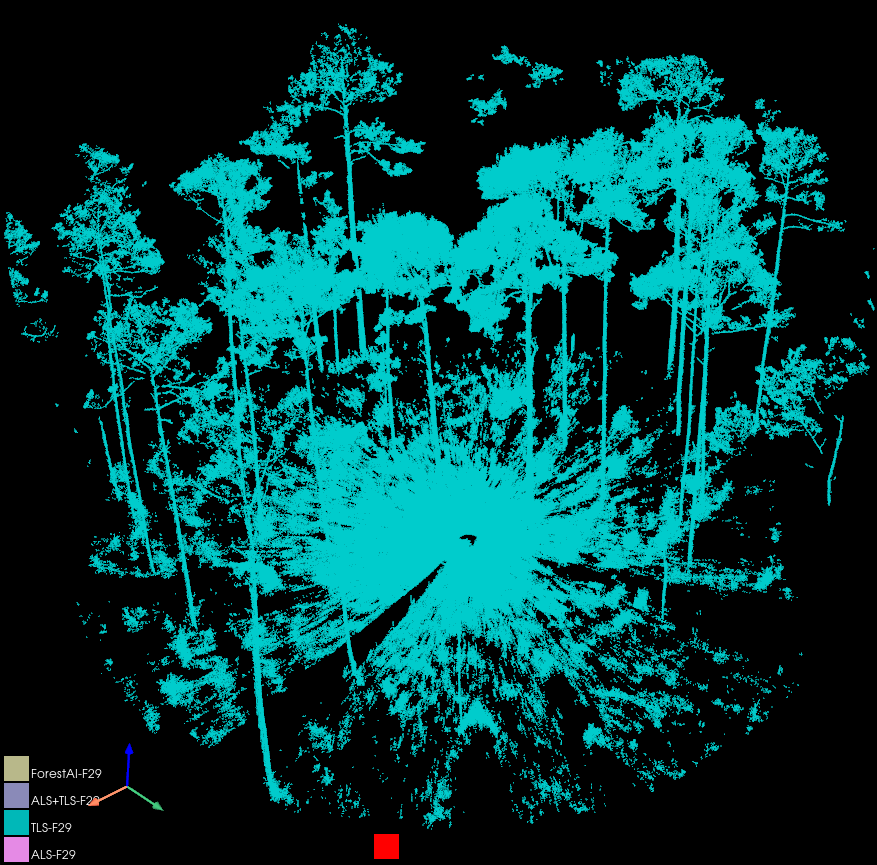}
	       \caption{Ex.1:TLS.}
	       \label{fig:ex1_als_tlsf26}
            \end{subfigure}
            \begin{subfigure}{0.27\linewidth}
	       \includegraphics[width=1.0\linewidth]{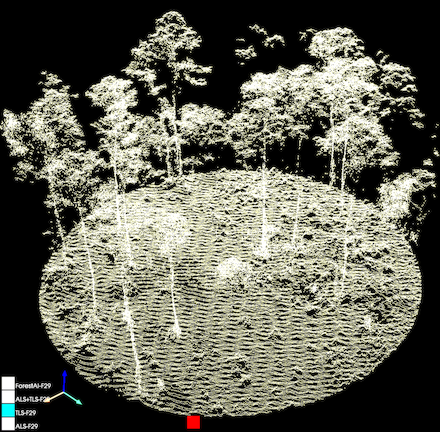}
	       \caption{Ex.1:ALS + ForestGen3D.}
	       \label{fig:ex1_als_genf26}
            \end{subfigure} 
            %

            \begin{subfigure}{0.27\linewidth}
	       \centering
	       \includegraphics[width=1.0\linewidth]{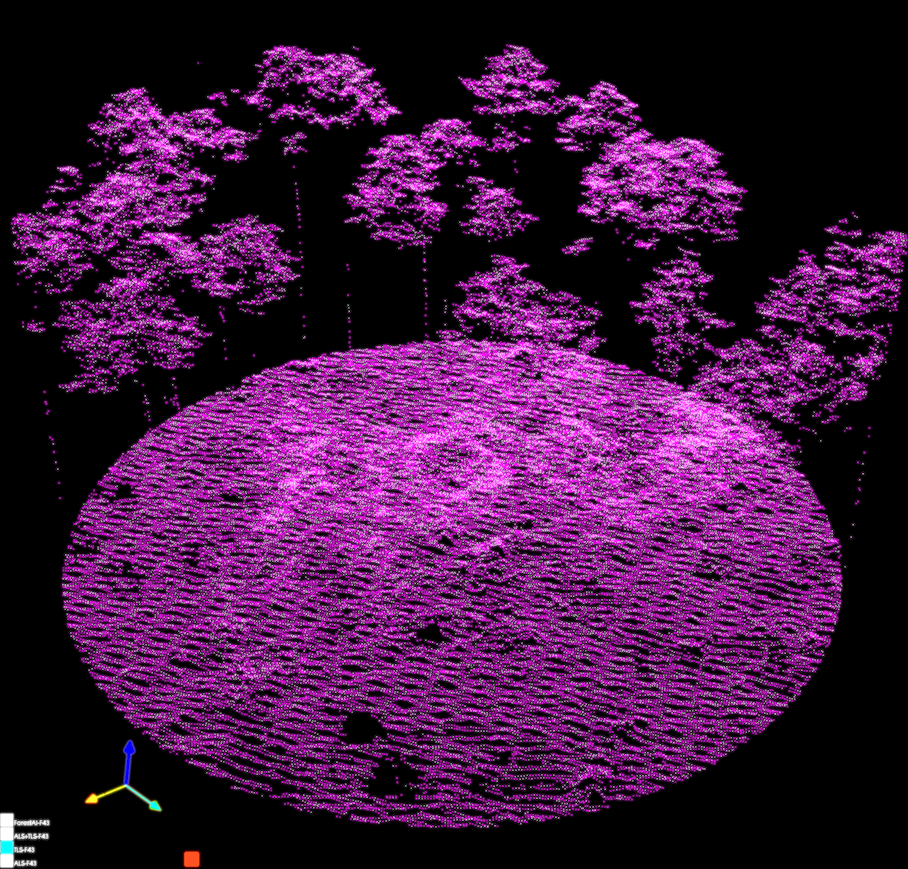}
	       \caption{Ex.4:ALS.}
	       \label{fig:ex4_alsf43}
            \end{subfigure} 
            \begin{subfigure}{0.27\linewidth}
	       \includegraphics[width=1.0\linewidth]{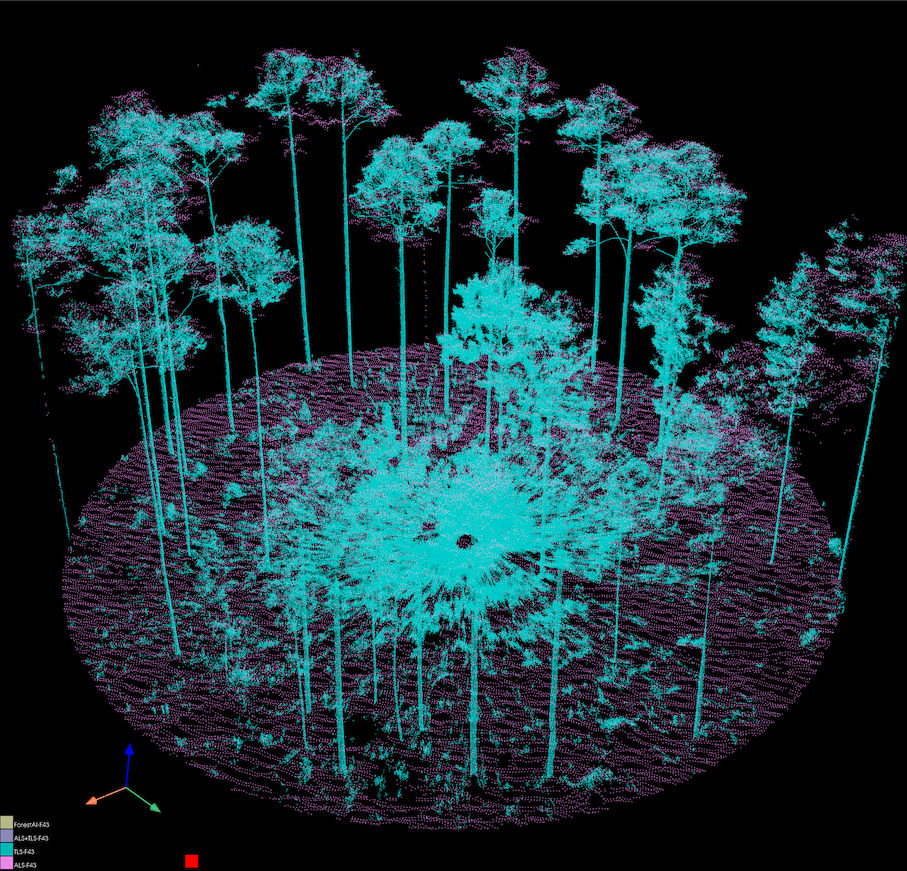}
	       \caption{Ex.4:TLS.}
	       \label{fig:ex4_tlsf43}
            \end{subfigure}
            \begin{subfigure}{0.27\linewidth}
e	       \includegraphics[width=1.0\linewidth]{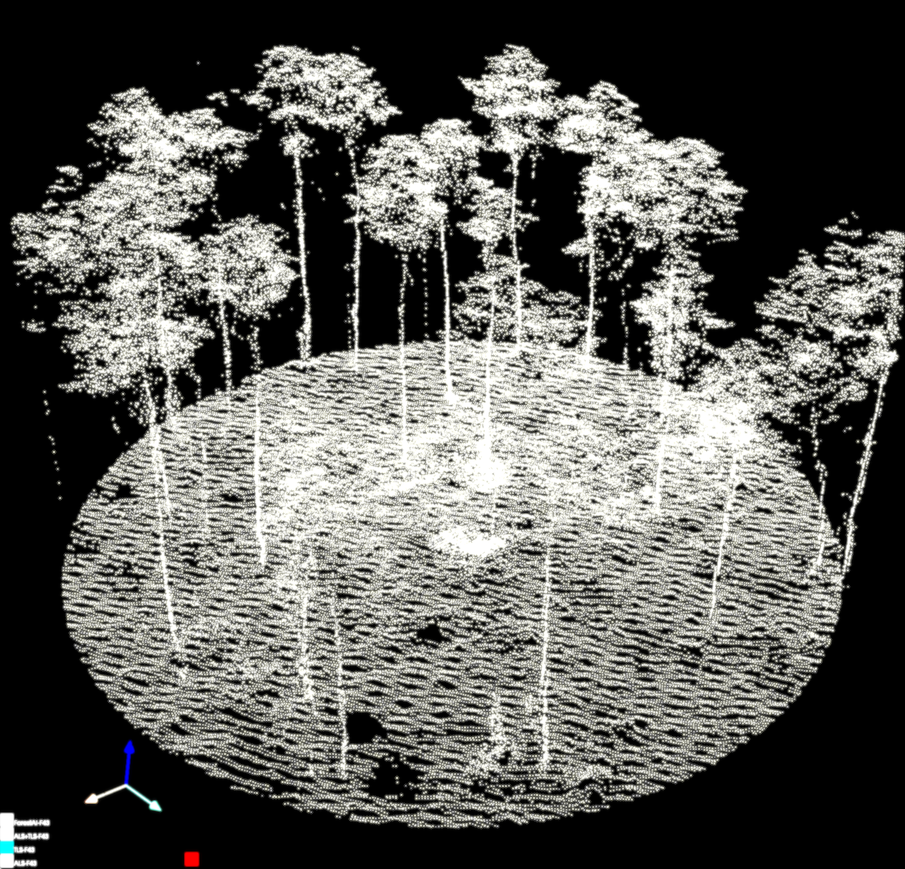}
	       \caption{Ex.4:ALS + ForestGen3D.}
	       \label{fig:ex4_als_genf43}
            \end{subfigure} 

            \begin{subfigure}{0.27\linewidth}
	       \centering
	       \includegraphics[width=1.0\linewidth]{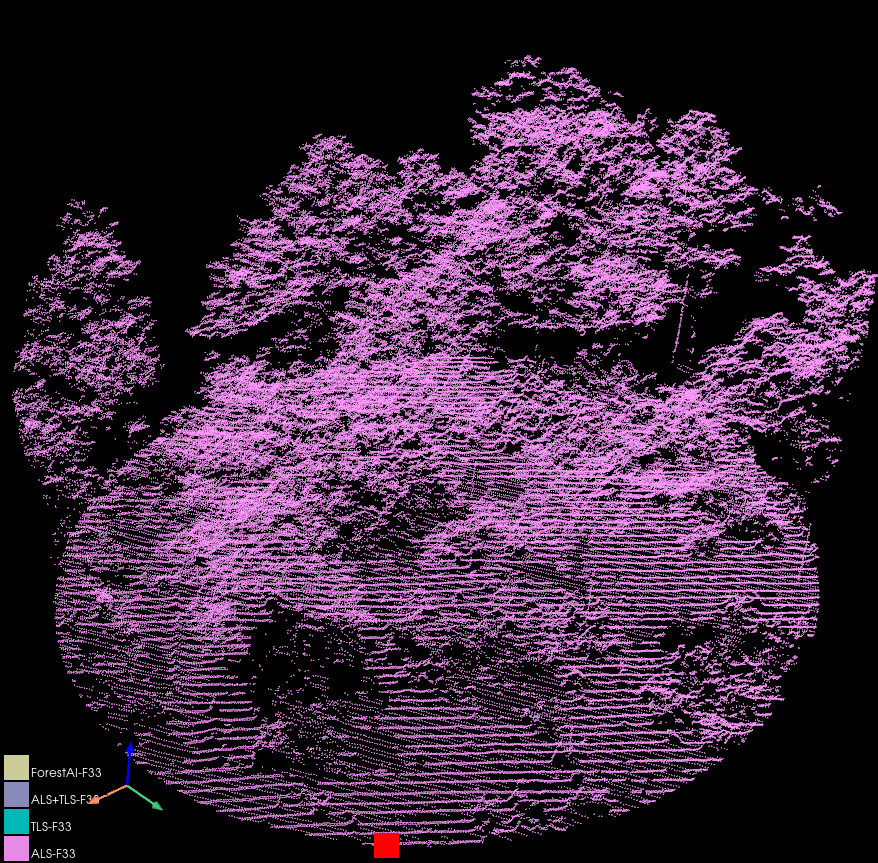}
	       \caption{Ex.2:ALS.}
	       \label{fig:ex2_alsf33}
            \end{subfigure} 
            \begin{subfigure}{0.27\linewidth}
	       \includegraphics[width=1.0\linewidth]{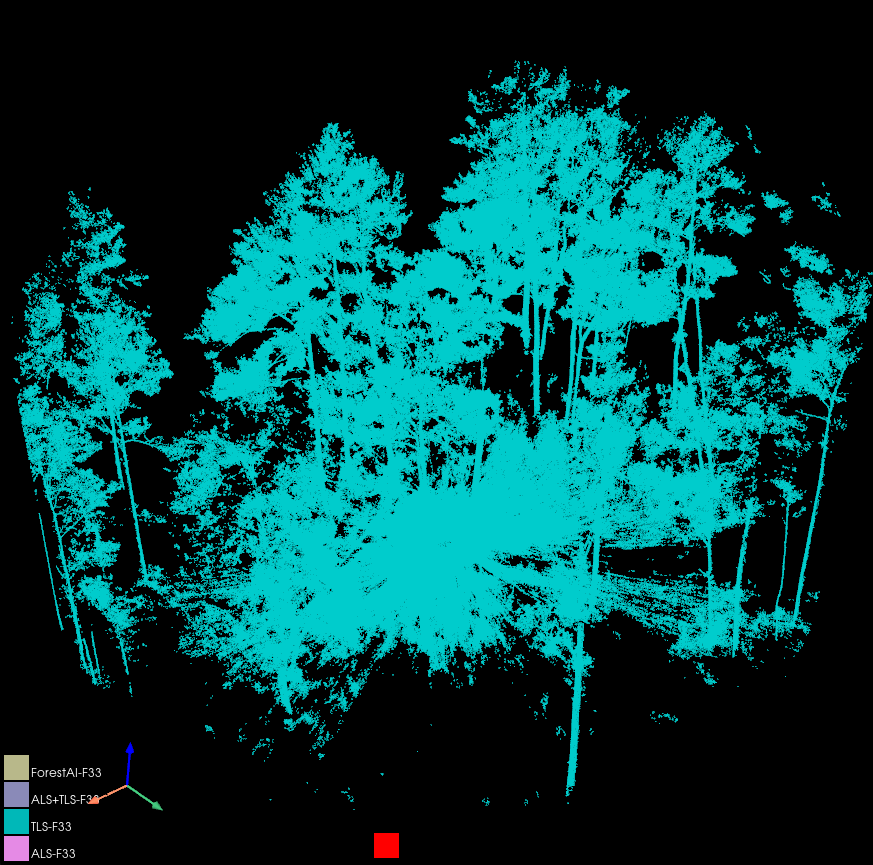}
	       \caption{Ex.2:TLS.}
	       \label{fig:ex2_als_tlsf33}
            \end{subfigure}
            \begin{subfigure}{0.27\linewidth}
	       \includegraphics[width=1.0\linewidth]{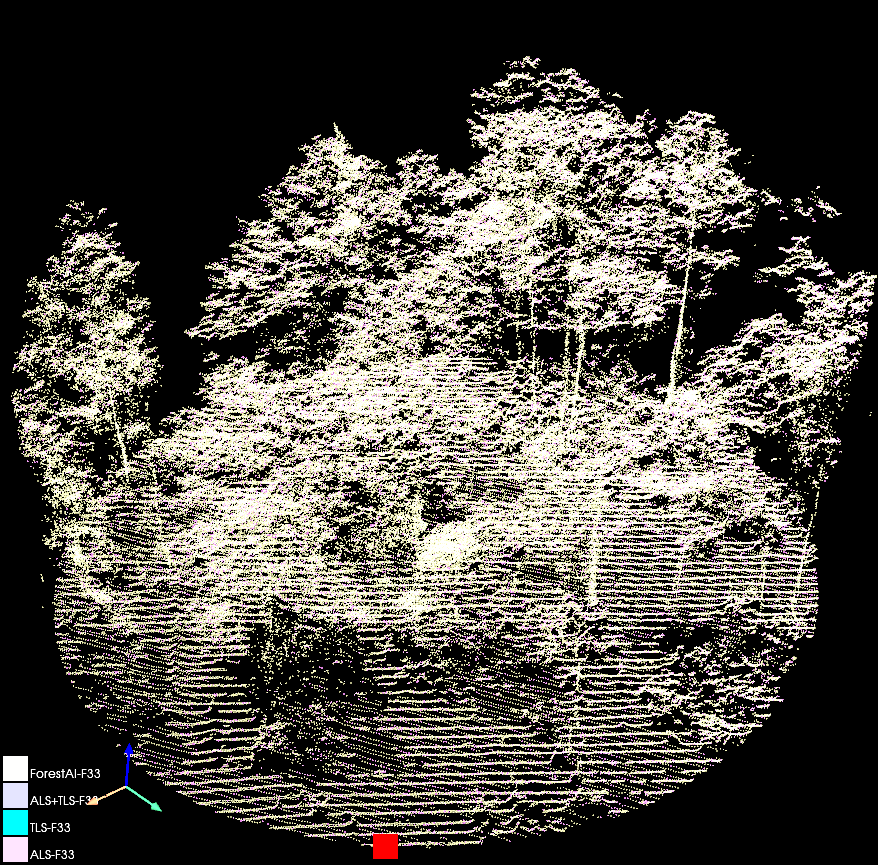}
	       \caption{Ex.2:ALS + ForestGen3D.}
	       \label{fig:ex2_als_genf33}
            \end{subfigure} 

            \begin{subfigure}{0.27\linewidth}
	       \centering
	       \includegraphics[width=1.0\linewidth]{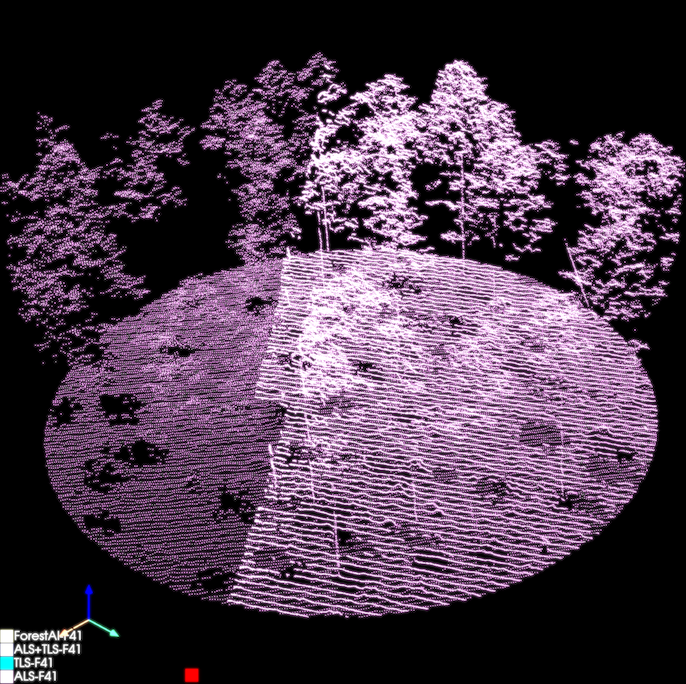}
	       \caption{Ex.3:ALS.}
	       \label{fig:ex3_alsf41}
            \end{subfigure} 
            \begin{subfigure}{0.27\linewidth}
	       \includegraphics[height=1.001\textwidth, width=1.0\linewidth]{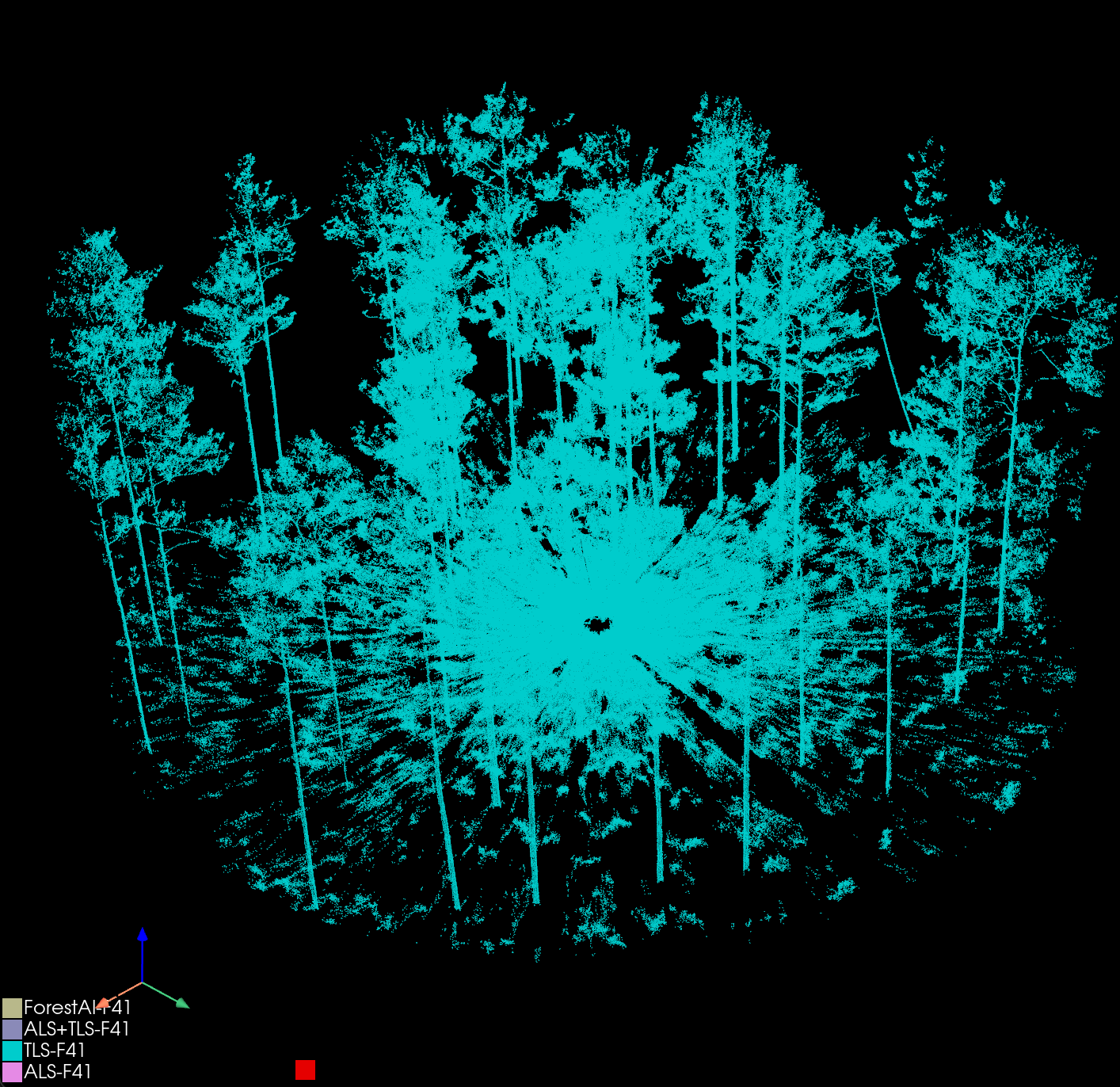}
	       \caption{Ex.3:TLS.}
	       \label{fig:ex3_als_tlsf41}
            \end{subfigure}
            \begin{subfigure}{0.27\linewidth}
	       \includegraphics[height=1.001\textwidth, width=1.0\linewidth]{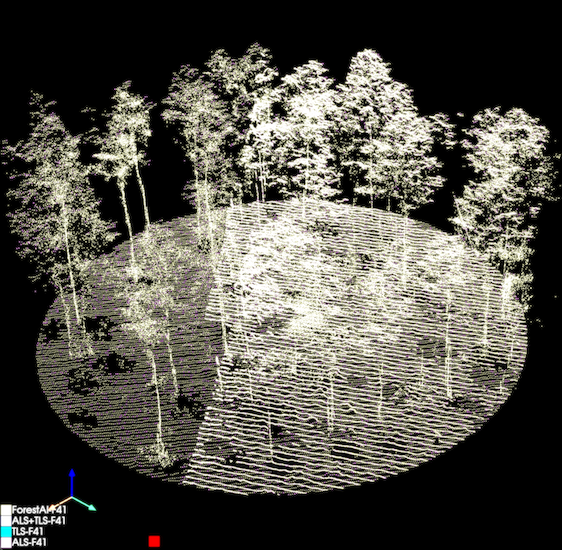}
	       \caption{Ex.3:ALS + ForestGen3D.}
	       \label{fig:ex3_als_genf41}
            \end{subfigure} 
        
    \caption{Representative 3D structural comparisons across sensing sources. Each row corresponds to a distinct 25~m radius plot with overlapping ALS and TLS coverage. Columns show ALS (left), TLS (middle), and ALS+ForestGen3D (right). ForestGen3D reconstructs missing vertical stem and mid-story structure absent in ALS, smoothly transitioning from ALS-observed canopy and ground returns into the generated canopy interior and sub-canopy regions.}
    \label{fig:plot_gen_biophysics}
\end{figure}

We also conduct both qualitative and quantitative analyses at the plot scale, centered at TLS scans and covering a 25 meter radius. For the qualitative evaluation, we visually compare structural reconstructions obtained from ALS, TLS, ALS+TLS and a hybrid approach combining ALS with ForestGen3D generation enhancement. For each plot, individual trees were first detected using 3D bounding boxes, and our ForestGen3D model was then applied to generate plausible understory structure beneath each detected tree. Regions lacking tree or shrub like vegetation are left unaltered under the assumption that they do not suffer from occlusion in ALS.
To support the quantitative assessment, we perform an evaluation in the downstream task of vegetation biometric estimates derived from each data source: ALS only, TLS only, ALS+TLS and ALS enhanced with ForestGen3D. The combined ALS+TLS data serves as ground truth, given its ability to capture detailed, high-resolution vegetation structure across the entire vertical extent of the tree, including the ground, overstory and understory components.
Figure \ref{fig:plot_gen_biophysics} illustrates four representative testset examples (one per row) at plot-scale. The first column illustrates the ALS point cloud, the second column shows the corresponding co-registered ground truth TLS scan, and the third column presents the results of aggregating the ALS scan with the generated under-story from our ForestGen3D model.
Note that in all examples, the ALS plots suffer from either missing or highly sparse understory information between the ground and tree foliage, as can be observed in the images in the first column where points are missing. TLS, on the other hand, provides high resolution in the elevation/altitude direction, as shown in the second column, where the main tree stem is finely resolved along the vertical tree-stand direction. The third column, showing the point cloud from ALS+ForestGen3D, illustrates that our approach is capable of enhancing or generating realistic understory structure at the plot scale, where the generated stems and ground features fill in gaps and transition smoothly from the available ALS data. 

\subsubsection{Do ForestGen3D-Derived Biometrics Approximate TLS-Derived Distributions?}
Biometric estimation from 3D point clouds is treated here as a downstream evaluation task rather than the primary objective of ForestGen3D. The goal of this analysis is to assess whether the additional sub-canopy structure generated by the model, while respecting ALS-provided geometry, produces biophysical metric distributions that are consistent with those derived from ALS+TLS. 
Accordingly, we first evaluate similarity at the level of biometric distributions, which offers a robust and sensor-agnostic assessment of structural realism, and then complement this analysis with per-tree RMSE-based experiments where reliable cross-source tree correspondences are available. This combined evaluation mitigates biases introduced by source-specific tree detection and segmentation differences while still enabling direct quantitative comparison when appropriate.

In the first set of experiments for this task, we compute biophysical tree metric distributions derived from ALS, TLS, ALS+TLS and ALS+ForestGen3D data sources. For this, we use a histogram per metric (one per column) depicting the distribution of key biophysical tree attributes, offering insight into the variability of tree structures within the sampled plot region. Such histograms were computed over co-registered sources at individual plots of 25 meter radius.
To produce these histograms, we evaluate tree height, diameter at breast height (DBH), crown diameter (CrD), and crown volume (CrV) using established deep learning–based tree segmentation methods~\cite{windrim2019forest}. These models segment each point cloud within detected vegetation bounding boxes into structural components: ground, low vegetation, stems, branches, and foliage, which are then used as inputs to geometric fitting procedures that yield the final biophysical metrics.
Tree height is estimated using the highest elevation point within each detected tree, subtracting a projection of that point onto the ground surface. DBH is computed following the method of~\cite{rijal2012development} applied to segmented stems, with robustness improved through a RANSAC based linear interpolation of fitted circles across multiple heights around breast height to mitigate the effect of outliers. Crown diameter is estimated by projecting segmented foliage points onto a horizontal plane and measuring the maximum pairwise distance. Crown volume is computed using a $k$-means–based convex hull approach~\cite{zhu2021assessing}, which partitions foliage points into spatial clusters, computes the convex hull of each cluster, and sums their volumes. This approach mitigates overestimation caused by empty space inclusion when applying a single convex hull to the full crown.

\begin{figure}
    \centering
	   \includegraphics[height=0.75\textwidth, width=0.75\linewidth]{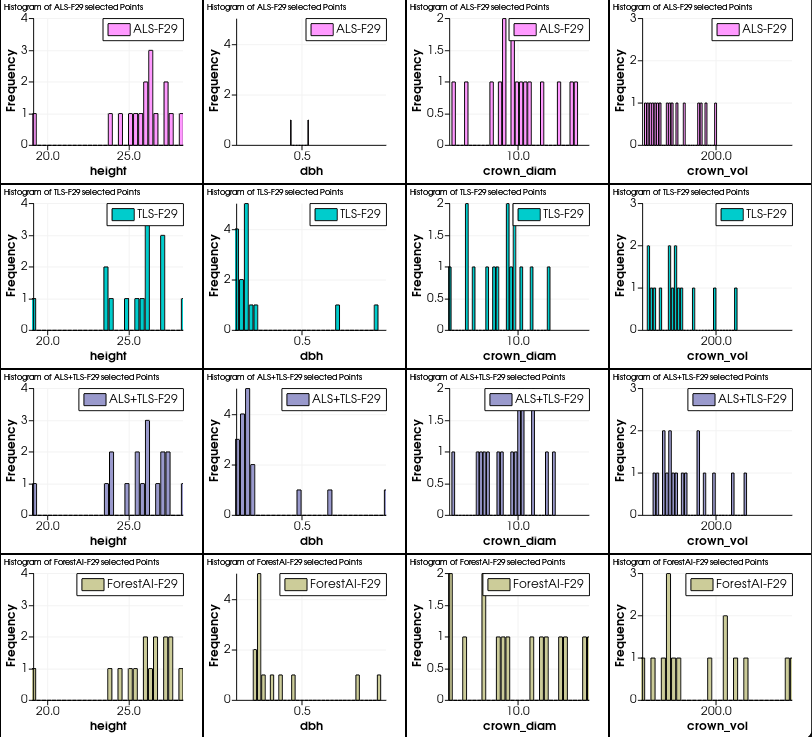}
	    \caption{Plot-scale distributions of biophysical tree metrics derived from different sensing sources. Rows correspond to ALS (row 1), TLS (row 2), ALS+TLS (row 3), and ALS+ForestGen3D (row 4). Columns show distributions for tree height (col. 1), diameter at breast height (DBH; col. 2), crown diameter (CrD; col. 3), and crown volume (CrV; col. 4). Each histogram represents metrics computed over a 25~m radius plot. ALS+TLS is treated as the structural reference, while ALS+ForestGen3D closely approximates its metric distributions, indicating recovery of sub-canopy structure while preserving ALS-derived spatial constraints.}
	     \label{fig:ex1_distf26}
\end{figure}

In Figure~\ref{fig:ex1_distf26}, integrated ALS+TLS data (third row across all examples) provides the most comprehensive representation of forest structure by combining ALS’s canopy coverage from the top with TLS’s detailed near-ground sampling. As such, biometrics derived from ALS+TLS are treated as a gold-standard reference for comparison.
ALS-only data (first row) efficiently captures canopy-top structure and yields reliable estimates of tree height and crown diameter, but lacks the resolution necessary to characterize lower-canopy and understory features due to occlusion. This limitation leads to sparse and biased DBH estimates across most examples, while height remains closer to the ALS+TLS reference because it depend less directly on stem-level information.
TLS-only data (second row) provides dense sampling of stems and lower-canopy elements, resulting in well-defined DBH distributions. TLS-derived crown diameter and height distributions are generally contained within ALS-derived ranges, consistent with the assumption that TLS crown points are spatially embedded within the ALS envelope. Smaller height values observed in TLS reflect understory trees occluded in ALS data or missed upper canopy. 
Differences in crown volume distributions reflect the complementary directional sampling strengths of ALS (horizontal canopy extent) and TLS (vertical crown structure).
The ALS+ForestGen3D results (last row) indicate that the proposed model compensates for ALS structural blind spots while preserving ALS derived landscape geometry. The resulting biometric distributions consistently approximate those of the ALS+TLS reference across all evaluated metrics. Improvements are most pronounced for DBH and crown volume, indicating recovery of sub-canopy and stem structure absent in ALS-only data.
Importantly, tree height and crown diameter distributions remain largely contained within the corresponding ALS-derived ranges, supporting the model’s geometric containment guarantees under the ELBO objective. Together, these results indicate that ForestGen3D generates structurally plausible 3D vegetation consistent with TLS-derived distributions while maintaining fidelity to ALS-provided canopy geometry. Other representative example illustrations can be found in Figures \ref{fig:ex2_distf33}, \ref{fig:ex3_distf26} and \ref{fig:ex4_distf26} in the Appendix.

Beyond qualitative assessment, we quantitatively evaluate model performance in this downstream task, by comparing the distributions of tree height, diameter at breast height, crown diameter, and crown volume derived from the different 3D sources. Because plot-scale biometric distributions are typically sparse (on the order of 10–30 trees per plot), we adopt the Wasserstein distance (WD) as our primary distributional comparison metric. WD is particularly well suited for this domain, as it remains stable under limited sample sizes and partial support overlap, unlike divergence-based measures such as the Kullback–Leibler divergence.
The Wasserstein distance (see \cite{panaretos2019statistical} for an overview) provides an intuitive measure of dissimilarity by quantifying the minimum cost required to transform one distribution into another and is therefore also referred to as the Earth Mover’s Distance. We explicitly distinguish this use of WD for comparing biometric distributions from the EMD metric which operates directly on 3D point sets.

\begin{table}[ht]
    \begin{minipage}{1.0\linewidth} 
        \centering 
        
        \captionof{table}{Average Wasserstein distances between plot-scale biophysical metric distributions derived from ALS, TLS, and ALS+ForestGen3D relative to ALS+TLS, computed over all 51 plots in the test set. Results confirm that ALS-ForestGen3D yields biometric distributions that most closely approximate those derived from ALS+TLS across all evaluated metrics, demonstrating consistent performance across diverse plots within the mixed-conifer ecosystem.} 
        \label{tab:table4_b} 
        \begin{tabular}{lllll} 
            \toprule 
            & \multicolumn{4}{c}{\textbf{Plot-scale Biophysical metrics}}\\
            \textbf{Sensing Source} & Height & DBH & CrD & CrV \\ 
            \midrule 
                ALS & 0.94 & 0.29 & 0.81 & $4.87^{3}$ \\ 
                TLS & 1.85 & 0.22 & 1.11 & $4.33^{3}$ \\
               ALS + ForestGen3D & 0.82 & 0.205 & 0.88 & $4.24^{3}$ \\
            \midrule
            \bottomrule 
        \end{tabular} 
    \end{minipage} 
\end{table}
Table~\ref{tab:table4_b} summarizes average Wasserstein distances computed across all 51 plots in the test set.
Across all examples and metrics, ALS+ForestGen3D consistently exhibits the smallest Wasserstein distances relative to ALS+TLS. Note in general, that in terms of distances the smallests' are observed relative to the sensing source that provides greater information content for the structural features underlying that metric. For example, Wasserstein distances for tree height are generally smaller between ALS+TLS and ALS than between ALS+TLS and TLS, reflecting the fact that ALS offers more uniform and less occlusion affected sampling of canopy tops and ground surfaces. Conversely, metrics such as DBH, which depend strongly on stem level sampling, show closer agreement between ALS+TLS and TLS. Overall, the results here indicates that the biometric distributions derived from ALS+ForestGen3D generation most closely approximate those obtained from the combined ALS+TLS data, which provides the most complete characterization of forest structure. These findings support that ForestGen3D effectively recovers below-canopy structural information missing from ALS-only observations while remaining consistent with ALS-provided canopy geometry. Note that WD for crown volume is shown in cubic meters, hence its high value difference compared to the other metrics.

The second portion of the biometric estimation task directly assesses the added value of ForestGen3D relative to conventional ALS-based allometric approaches. This consists in estimating DBH from ALS-derived crown diameter and tree height using established allometric relationships following Jucker et al.~\cite{jucker2017allometric}. Model parameters were trained using 453 matched trees from the Ft. Stewart 2022 dataset, drawn from geographically distinct plots relative to the test set but within the same landscape and vegetation regime.
Evaluation was performed on 246 test matched trees from the Ft. Stewart 2024 dataset, with ALS+TLS point clouds used as reference. To mitigate nonuniform sampling effects associated with fixed single scan TLS acquisitions, each point cloud was truncated to a 25~m radius around the TLS scan center. Trees were matched across co-registered ALS, TLS, ALS+TLS, and ALS+ForestGen3D sources using a 0.5~m spatial neighborhood.
Table \ref{tab:table5_allometry} summarizing RMSE's across all four evaluated biometrics: tree height, DBH, crown diameter, and crown volume show that ALS+ForestGen3D achieved the lowest RMSE. In particular, DBH error was reduced from 0.208~m (ALS allometry \cite{jucker2017allometric}) to 0.112~m (ALS+ForestGen3D), while crown volume error was reduced by more than 50\%. These results demonstrate that ForestGen3D not only reconstructs realistic 3D structure but also improves downstream biometric estimation relative to conventional ALS based allometric models.

\begin{table}[ht]
    \begin{minipage}{1.0\linewidth} 
        \centering 
        \captionof{table}{Root mean square error (RMSE) of biophysical metrics derived from different sensing sources relative to the ALS+TLS reference. Metrics include tree height, diameter at breast height (DBH), crown diameter (CrD), and crown volume (CrV). DBH estimates for ALS-only inputs are obtained using an allometric model based on crown diameter and height following \cite{jucker2017allometric}. Results are reported for a test of 265 matched trees across all sources within the Ft. Stewart 2024 dataset. Lower RMSE values indicate closer agreement with the ALS+TLS benchmark, with ALS+ForestGen3D achieving the lowest error across all metrics.}
        \label{tab:table5_allometry} 
        \begin{tabular}{lllll} 
            \toprule 
            & \multicolumn{4}{c}{\textbf{Plot-scale Biophysical metrics}}\\
            \textbf{Sensing Source} & Height & DBH & CrD & CrV \\
            \midrule 
                ALS & 0.531 & 0.208 (allometric \cite{jucker2017allometric}) & 0.566 & $5.112^3$ \\ 
                TLS & 1.508 & 0.143 & 0.6127 & $4.029^3$ \\
               ALS + ForestGen3D & 0.480 & 0.112 & 0.561 & $2.109^3$ \\
            \midrule
            \bottomrule 
        \end{tabular} 
    \end{minipage} 
\end{table}

\subsection{Regional-scale ALS-Guided Generation} \label{Ssec:regionalscale}
\begin{figure}
    \centering
    \begin{subfigure}{0.33\linewidth}
	\centering
	\includegraphics[width=1.0\linewidth]{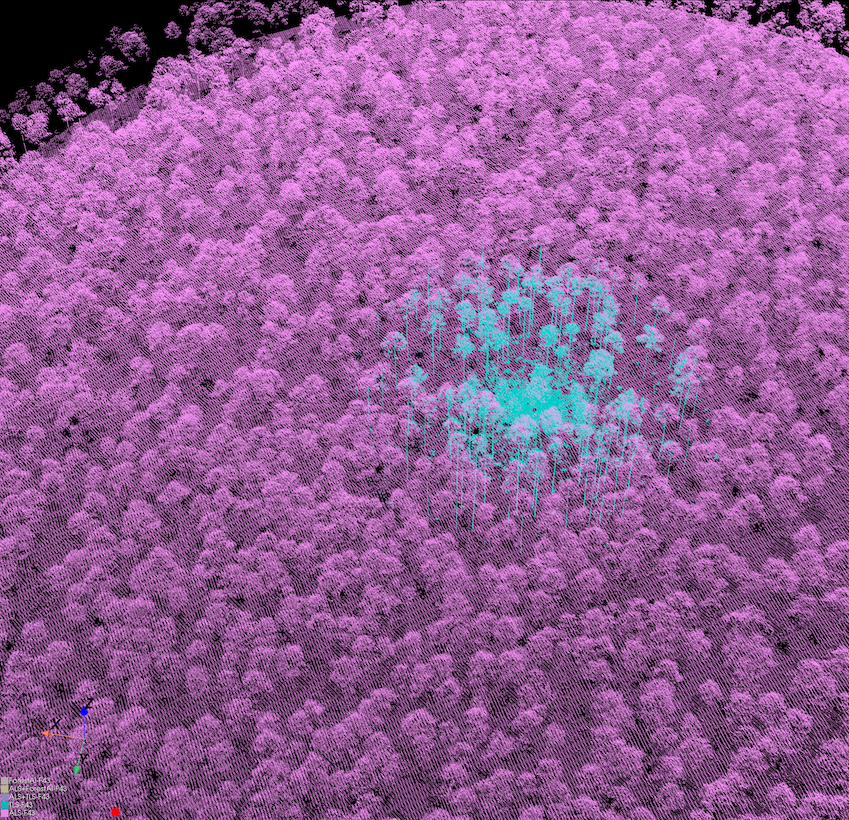}
	\caption{Ex.1: ALS (magenta)+TLS (Turquoise).}
	\label{fig:ex1_als_tlsf43_landscape}
    \end{subfigure}
    \begin{subfigure}{0.33\linewidth}
	\centering
	\includegraphics[ width=1.0\linewidth]{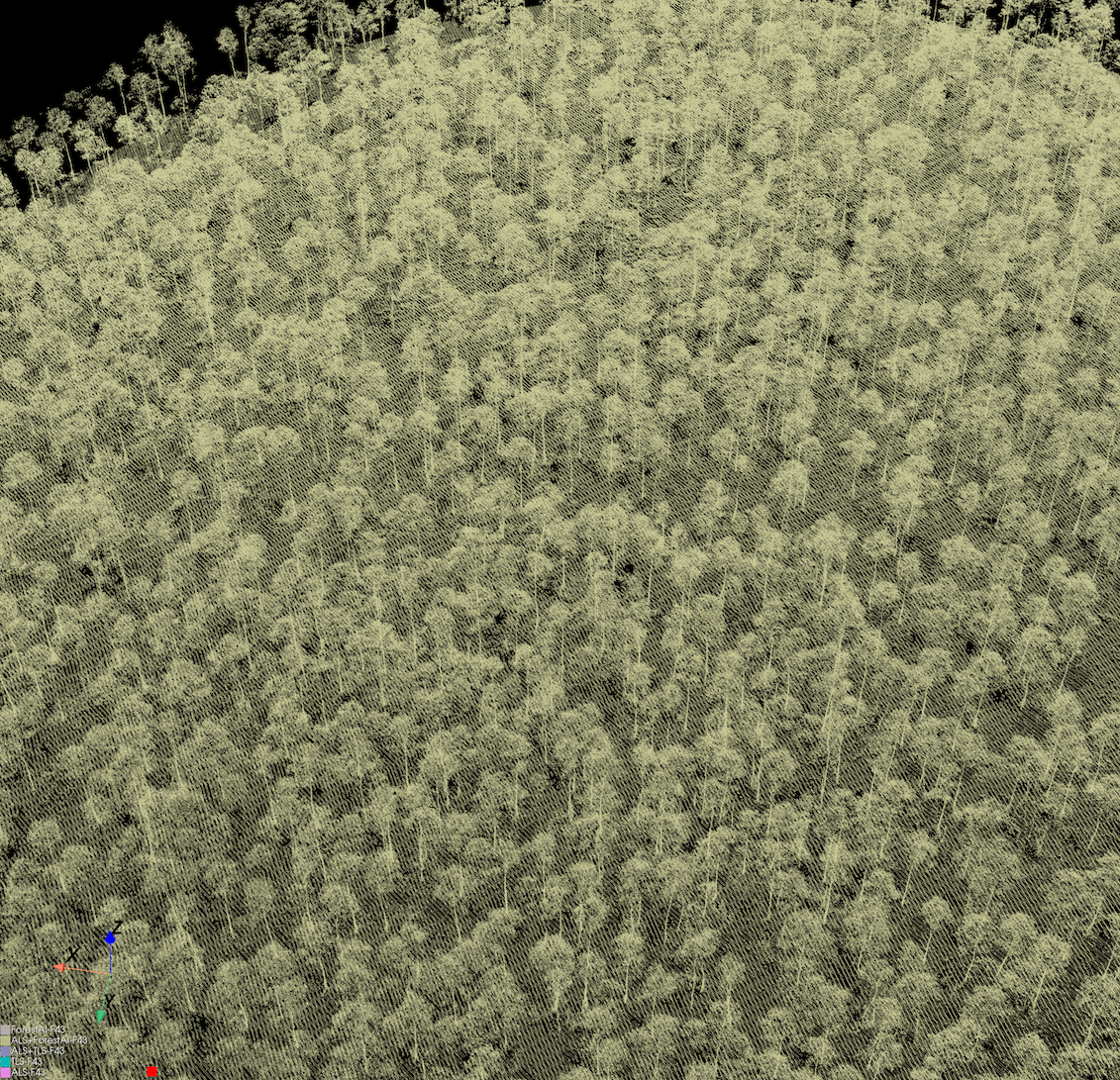}
	\caption{Ex.1: ALS + ForestGen3D (Peach)}
	\label{fig:ex1_als_genf43_landscape}
    \end{subfigure}
    \begin{subfigure}{0.33\linewidth}
	\includegraphics[width=1.0\linewidth]{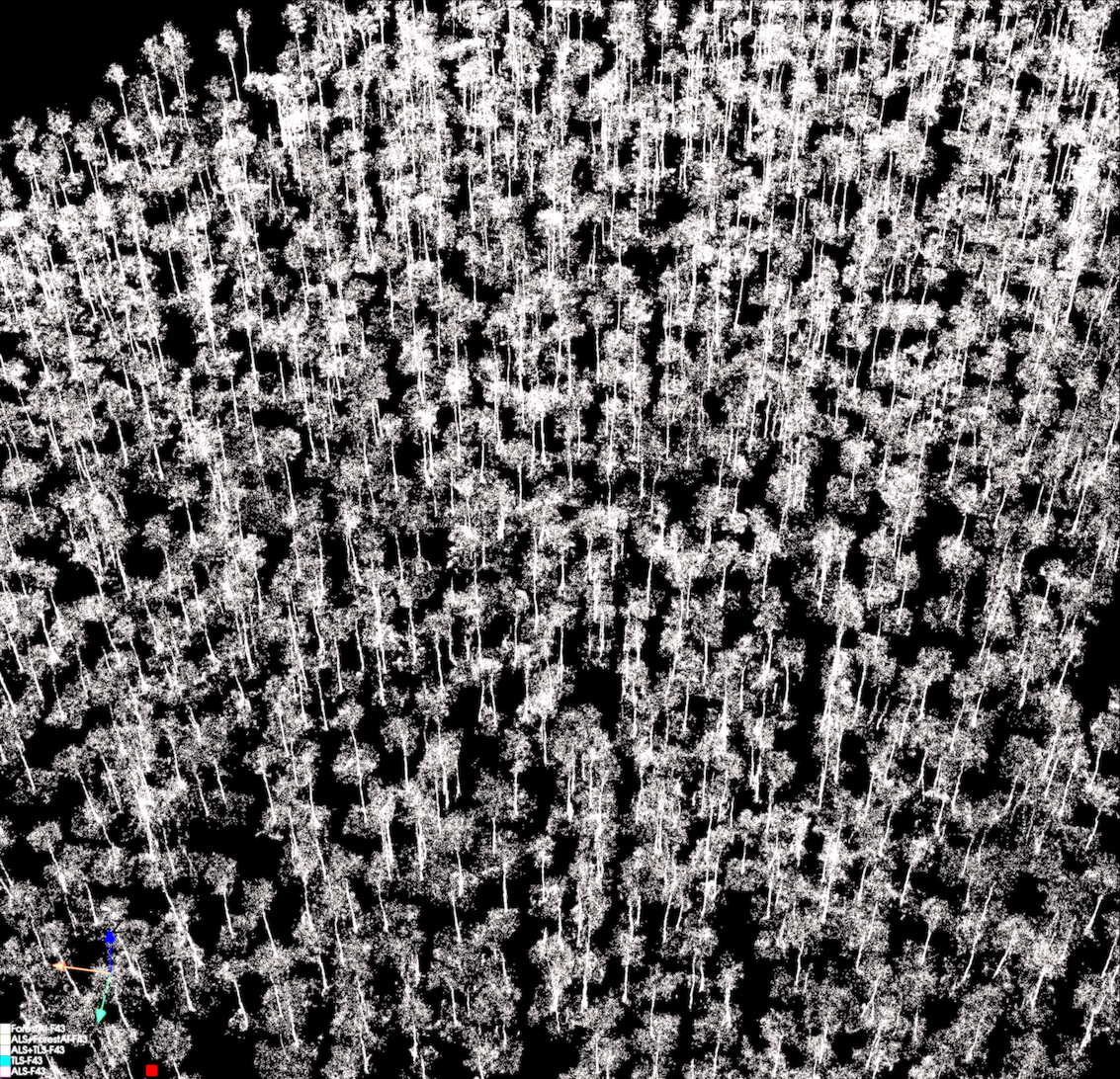}
	\caption{Ex.1: ForestGen3D (Light gray)}
	\label{fig:ex1_genf43_landscape}
    \end{subfigure}

    \begin{subfigure}{0.33\linewidth}
	\centering
	\includegraphics[width=1.0\linewidth]{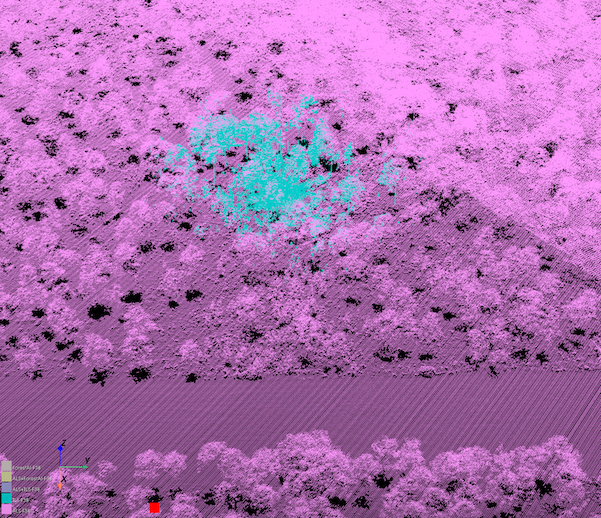}
	\caption{Ex.2: ALS (magenta)+TLS (Turquoise).}
	\label{fig:ex2_als_tlsf38_landscape}
    \end{subfigure}
    \begin{subfigure}{0.33\linewidth}
	\centering
	\includegraphics[ width=1.0\linewidth]{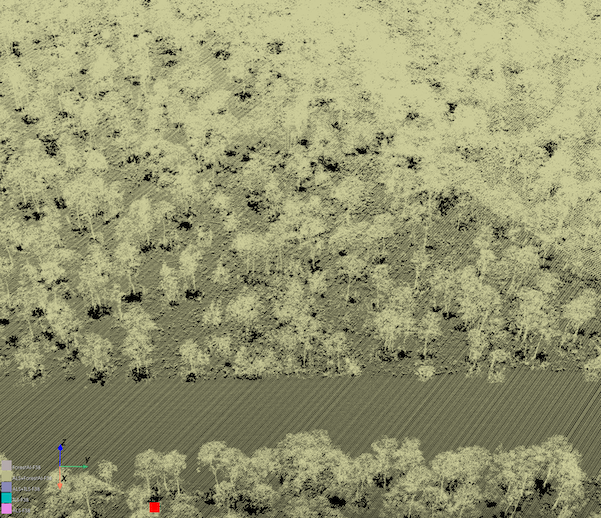}
	\caption{Ex.2: ALS + ForestGen3D (Peach)}
	\label{fig:ex2_als_genf38_landscape}
    \end{subfigure}
    \begin{subfigure}{0.33\linewidth}
	\includegraphics[width=1.0\linewidth]{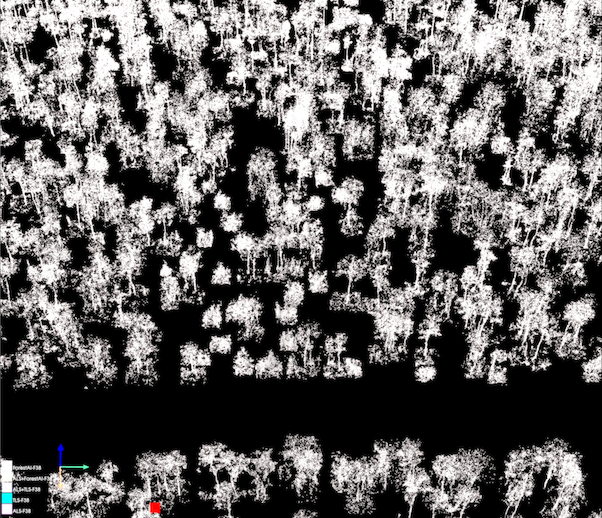}
	\caption{Ex.2: ForestGen3D (Light gray)}
	\label{fig:ex2_genf38_landscape}
    \end{subfigure}

    \begin{subfigure}{0.33\linewidth}
	\centering
	\includegraphics[width=1.0\linewidth]{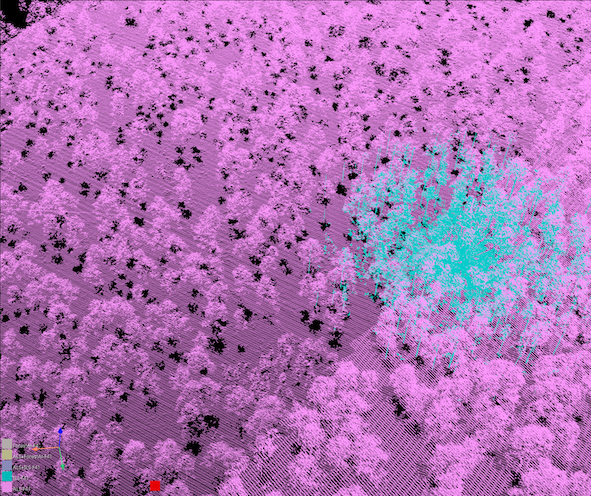}
	\caption{Ex.3: ALS (magenta)+TLS (Turquoise).}
	\label{fig:ex3_als_tlsf41_landscape}
    \end{subfigure}
    \begin{subfigure}{0.33\linewidth}
	\centering
	\includegraphics[ width=1.0\linewidth]{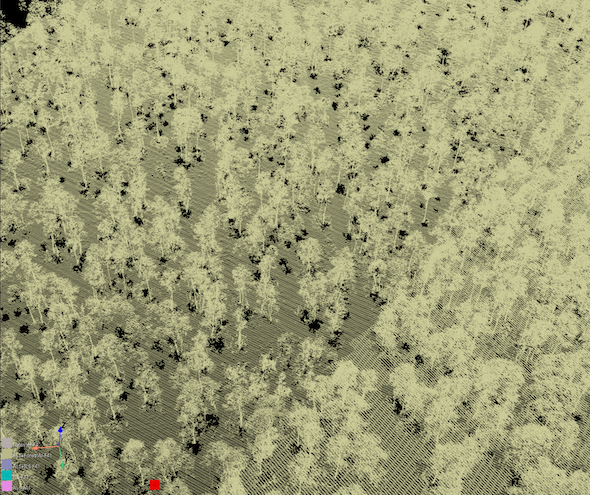}
	\caption{Ex.3: ALS + ForestGen3D (Peach)}
	\label{fig:ex3_als_genf41_landscape}
    \end{subfigure}
    \begin{subfigure}{0.33\linewidth}
	\includegraphics[width=1.0\linewidth]{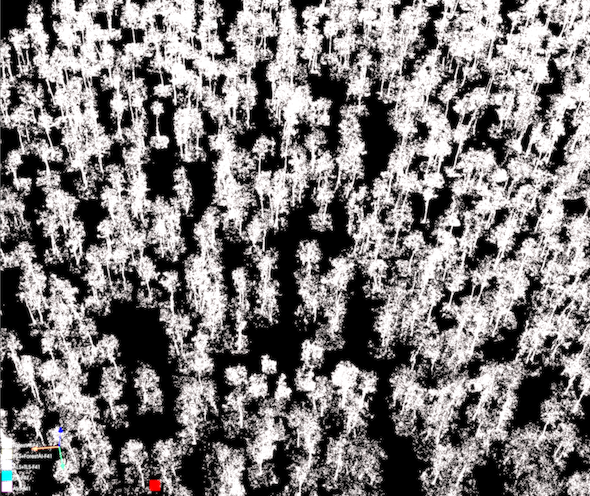}
	\caption{Ex.3: ForestGen3D (Light gray)}
	\label{fig:ex3_genf41_landscape}
    \end{subfigure}

    \caption{Landscape-scale generation examples from three different areas (one per row). Columns show (a,d,g) broad ALS (magenta) with \textit{in-situ} TLS reference (turquoise), (b,e,h) ALS combined with ForestGen3D output (peach), and (c,f,i) ForestGen3D alone (light gray). Each snapshot corresponds to a circular area of 200 m radius. ForestGen3D was applied selectively to regions with occlusion in ALS, filling in subcanopy structure while leaving unoccluded regions unchanged (black areas indicate no generation). Note that these examples are outside the training set.}
    \label{fig:landscape}
\end{figure}

To evaluate the scalability of the ForestGen3D model, we applied it to generate 3D forest structure over a larger spatial extent, a 200-meter radius region centered at TLS scan geo-referenced locations. While the TLS scan itself covers a range of only ~25 meters, this broader area enables us to test the model’s ability to extrapolate beyond small-scale. Notably, the ALS data used in this region was not part of the training set.
The generation procedure at this larger scale mirrors the approach used for plot-level synthesis. First, we apply a tree detection algorithm across the entire ALS region of interest to identify areas with occlusion when viewed from above, which are typically regions where airborne LiDAR fails to resolve understory structure. Second, we extract 3D bounding boxes around these detected trees and apply the ForestGen3D model independently within each bounding box using the local ALS point cloud as input. 
Finally, the resulting generated point clouds are aggregated over the full 200-meter radius area keeping track of their localization to produce a coherent forest 3D reconstruction.
Figure \ref{fig:landscape} illustrate representative snapshot examples of the resulting generation/reconstruction. In these visualizations, point clouds are color-coded as follows: magenta for ALS, turquoise for TLS, peach for ALS + ForestGen3D generation, and white for the generated ForestGen3D samples alone. 
In this test case where TLS data is not available, the model generates understory structure consistent with the characteristic statistical distribution observed in the training data.
Importantly, these large-scale results corroborate the findings from the plot-scale experiments. ForestGen3D generates foliage and branching structures that blend smoothly into those observed in the ALS data, maintaining spatial consistency and fidelity. In understory regions lacking ALS points, the model synthesizes realistic and plausible tree trunks in locations and with diameters that appear structurally consistent with the associated crown geometry provided by ALS. When sparse ALS points are present in the understory, the generated output enhances and complements this limited data by reinforcing plausible stem/branch geometry. At the ground level, the model enriches the representation of shrubs and low vegetation while also improving the resolution of ground returns and the near-ground structure. 

To explicitly evaluate whether the generated 3D structure remains spatially constrained by the ALS observations, particularly in the absence of TLS ground truth at landscape scales, we next introduce a complementary evaluation based on the expected point containment (EPC) in Eq. \eqref{epc}. This analysis quantifies the extent to which generated points respect the implicit spatial bounds defined by the ALS convex hull, providing a geometry aware validation of large scale structural plausibility.
\begin{figure*}
	   \centering
	   \includegraphics[width=1.0\linewidth]{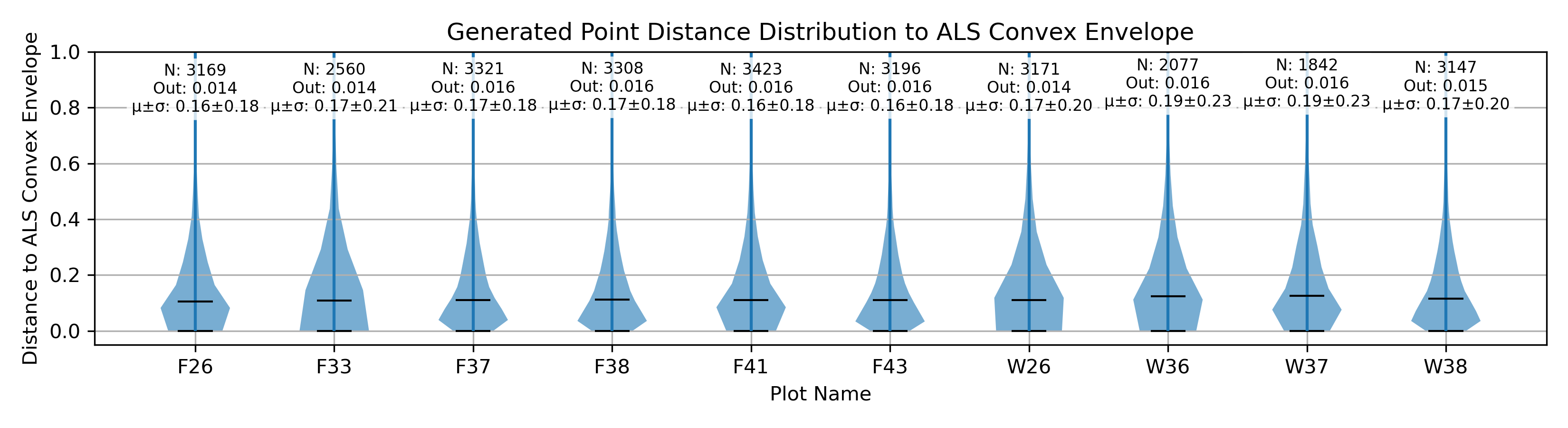}

     \caption{ 
        Distribution of point-to-envelope distances between ForestGen3D generated points and the ALS derived convex hull across ten landscape-scale (200 m radius) regions. The x-axis represents diferent geo-located areas, the y-axis indicates the minimum Euclidean distance from generated points lying outside the ALS convex hull to their nearest point on the hull surface. The violin shape represents the distribution of these distances; the internal black bar denotes the median. Annotations above each plot report: $N$, the number of trees generated in the region; \textit{Out}, the fraction of generated points falling outside the ALS envelope; and the mean ± standard deviation ($\mu \pm \sigma$) of out-of-envelope distances. Across all regions, the proportion of outside points remains below 1.6\%, and deviations remain small (typically <0.2 m). These results show that ForestGen3D generates structure that remains spatially consistent with the ALS envelope while enriching sub-canopy detail, confirming that the model respects the geometry encoded in ALS measurements.
    }
	\label{fig:tls_als_violin}
\end{figure*}
For this, we analyzed the EPC metric and the distribution of distances between generated points falling outside the ALS convex envelope and their nearest point on the envelope surface. Figure~\ref{fig:tls_als_violin} summarizes this analysis across 10 evaluation plots, each representing a 200 meter radius ALS-only region labeled by plot name on the x-axis. Each violin plot shows the distribution of out-of-envelope distances, computed only for generated points that lie outside the ALS convex hull. The line inside each violin represents the median, while the shape reflects the underlying distance histogram.
Text annotations above each violin indicate: (i) $N$, the number of trees generated per 200m radius region; (ii) \textit{Out}, the ratio of generated points outside the ALS convex hull (relative to the total number of points); and (iii) the mean ± standard deviation of the out-of-envelope distances. Across all plots, the fraction of points outside the ALS envelope remains consistently low, ranging from 1.4\% to 1.6\%, and mean distances remain below 0.2m, demonstrating good quality generations via the EPC. 
The results in Fig. \ref{fig:tls_als_violin} provide an important consistency check linking landscape scale ALS geometry with the fine scale structure generated by ForestGen3D. Because ALS often undersamples subcanopy regions, a generative model could in principle hallucinate structure that violates spatial constraints or extends far beyond the areas supported by ALS evidence. The violin plot analysis quantitatively shows that this does not occur in practice: the model generates new subcanopy detail while remaining tightly confined to the ALS spatial envelope. This demonstrates that ForestGen3D enriches incomplete ALS point clouds without distorting broader landscape geometry and, consistent with earlier findings linking high EPC to low Chamfer Distance (CD) and Earth Mover’s Distance (EMD) under TLS supervision, indicates that the model achieves low 3D reconstruction error and high-quality generation at the landscape scale.

Summarizing, across all evaluations, ForestGen3D demonstrates strong and consistent quantitative performance in reconstructing sub-canopy forest structure from ALS input alone. At the individual tree scale, the model achieves low reconstruction error relative to TLS references, as shown by Chamfer distance, Earth's Moving Distance, and expected point containment (Fig. \ref{fig:hyperparameter}, Fig. \ref{fig:epc_plots}, Table \ref{tab:table_distances}). Biophysical attributes derived from generated point clouds tree height, DBH, crown diameter, and crown volume closely track TLS measurements and outperform or complement ALS-based or ALS-allometry based estimates (Table \ref{tab:table4_b}, Table \ref{tab:table5_allometry}), indicating that the model recovers realistic structural detail not observable from ALS alone. At landscape scale, ForestGen3D preserves global spatial consistency with ALS geometry, with the EPC metric and out-of-envelope distance distributions (Fig. \ref{fig:tls_als_violin}) confirming that more than 98\% of generated points remain contained within the ALS convex hull and deviations remain minimal (<0.2 m). Together, these quantitative analyses demonstrate that ForestGen3D provides a reliable and ecologically coherent enhancement of ALS-only forest structure, producing high-fidelity 3D reconstructions across individual-tree, plot-scale and landscape-level settings.

\subsection{Discussion}
Our findings confirm that DDPM models can generate realistic 3D representations of forest tree structures. The model's ability to preserve fidelity to the true 3D structure depends on the informativeness of additional conditioning data sources. Furthermore, the computational complexity of the generative process proposed here scales linearly with area coverage, enabling efficient deployment across large landscapes without increasing model complexity. With current implementations taking $\approx 0.4$ seconds per tree-scale generation in a Tesla V100S-PCIE-32GB GPU and with possibilities to further parallelize generation. However, the approach has several limitations.
First, the generative model is limited by the distribution it has learned from the training data, which in this case primarily consists of mixed coniferous forest types. When applied to generate trees with structural characteristics not well represented in the training set such as species with drastically different crown architectures, stem densities, branching patterns or ground vegetation (e.g., small shrubs, debri), the model may produce structures that deviate from true vegetation distributions. As a result, generalization to out-of-distribution ecosystems, such as boreal forests, tropical canopies, or hardwood-dominated stands, remains open. Addressing this limitation will require either broadening the training dataset to cover more ecological diversity or incorporating domain adaptation techniques to enable transfer across forest types with different structural priors. 
While the ALS convex hull provides a useful geometric prior for bounding plausible TLS generation, it is important to note that this containment assumption is approximate. The convex envelope may under/overestimate volume in sparsely sampled regions, and containment does not guarantee structural realism on its own. However, we empirically observe that TLS scans lie within the ALS envelope with over 96\% consistency in the training set, and that generation fidelity to the true structure and containment are strongly correlated. These findings support its use as a practical diagnostic metric in large-scale, ground-truth limited deployments.
In addition, while single-scan TLS provides a practical and widely used means of capturing tree-level structure, we recognize that multi-scan, multi-view TLS acquisitions would offer a more complete 3D characterization of vegetation by substantially reducing occlusion, especially in lower and mid-story foliage. Access to such high-quality, fully co-registered TLS datasets would benefit generative modeling in two ways. First, it would provide a less biased training distribution, allowing the DDPM to learn under-story geometry from point clouds that better approximate the true forest structure. Second, improved completeness in the TLS training dataset would likely enhance the model’s ability to reproduce fine-scale geometric features and further improve downstream LiDAR-based biometric estimate tasks. Exploring multi-scan TLS datasets therefore represents a promising direction for future work, particularly for advancing the model’s capacity to learn and generate detailed understory and ground-fuel structure.

In dense forest stands, tree detection is often challenged by dense canopies that occlude crown boundaries in ALS data. As a result, the detection pipeline may assign a single bounding box to multiple adjacent trees, which in turn affects the input to the generative model. Depending on the structural cues available in the ALS input, such as crown asymmetry, partial separation, or variations in canopy height the generative model may synthesize either a single fused structure or multiple distinct stems. Because the training dataset includes several examples of closely packed multi-tree configurations, the capacity to distinguish between adjacent trees when sufficient information is present in the point cloud. Nevertheless, in cases where ALS input lacks clear separation cues, the resulting generations may overestimate stem diameter or merge tree forms. To improve robustness in these scenarios, future work will explore both during training and deployment, forest context–aware adaptive tree detection methods, including the use of segmentation masks for individual crown delineation as in \cite{yang2020individual}, as well as stem count aware conditioning strategies to better guide the generative process when multiple trees are present within a single detection.
Finally, the model does not explicitly account for temporal variations in forest structure caused by seasonal changes, prescribed burns, wildfires, thinning, or disease. Ignoring these dynamic factors can create a mismatch between generated and real world forest conditions, limiting the model’s applicability for long-term ecological monitoring and fuel characterization. The aim in future studies is to develop models with additional training/conditional data informative of the potential temporal dynamics either occurring naturally or from human intervention.

Despite these limitations, the ability to reconstruct fine-scale 3D vegetation structure from ALS at tree, plot, and regional scales offers immediate value for next-generation wildfire and ecosystem modeling frameworks. ForestGen3D generates spatially explicit 3D point clouds that preserve the heterogeneous landscape geometry encoded in ALS while reconstructing missing sub-canopy and stem level structure. This capability is particularly important for advancing next-generation fire behavior and fuel modeling systems based on FIRETEC \cite{linn2002studying}, QUIC-Fire \cite{linn2020quic}, and FASTFuels \cite{Marcozzi_2025}, which are beginning to use 3D representations of canopy gaps, ladder fuels, and vertical fuel continuity. By producing high-resolution 3D forest structure that remains spatially consistent with ALS observations, ForestGen3D enables high-resolution 3D models to be applied in ALS-only environments while supplying the sub-canopy detail required for physically realistic simulation.
While ALS-based allometric models have proven effective for estimating scalar forest attributes such as DBH, biomass, and crown dimensions from canopy-level measurements, they fundamentally operate on summary metrics and do not reconstruct explicit three-dimensional geometry. In this work, biometric estimation is therefore treated as a downstream evaluation task rather than the primary objective. Our results show that ForestGen3D not only reconstructs realistic 3D structure but also yields biophysical metric distributions that closely match those derived from TLS. Quantitative evaluations using distributional similarity metrics and RMSE demonstrate that ALS+ForestGen3D improves DBH estimation relative to conventional ALS-based allometry, indicating that the model captures ecologically meaningful sub-canopy and stem-level variability.

\section{Conclusion}
\label{sec:conclusion}

We present ForestGen3D, a scalable generative modeling framework that produces high-resolution 3D forest structure from aerial LiDAR alone. By leveraging denoising diffusion probabilistic models trained on co-registered aerial and terrestrial LiDAR data, ForestGen3D reconstructs sub-canopy detail that is typically occluded in ALS, enabling realistic and ecologically grounded 3D generations across landscapes. Our work introduces a geometric containment prior, referred to as the expected point consistency, preserved over model generations ensuring spatial plausibility, and we show that this prior can serve as a useful proxy for reconstruction quality in the absence of ground truth TLS.
Empirical results across multiple spatial scales confirm that ForestGen3D approximates TLS structure with high fidelity in both geometric and ecological terms. Biophysical metric distributions derived from generated data such as tree height, DBH, crown diameter, and volume closely match those from full ALS+TLS data, demonstrating the model’s utility in ecological monitoring and risk analysis.
Despite these strengths, the model's generalizability is constrained by the structural diversity present in the training dataset. Ecosystems with markedly different vegetation forms or disturbance regimes may require adaptation via domain transfer or broader training sets. In addition, the model currently does not explicitly account for temporal dynamics such as seasonal change, management treatments, or natural disturbance recovery. Future work will extend the framework to incorporate more ecosystem diversity, temporal dynamics, and time-aware conditioning to further improve fidelity and robustness.
Overall, ForestGen3D provides a practical and extensible solution to the challenge of generating detailed 3D forest structure in ALS only environments, with promising applications in ecological forecasting, remote sensing, and wildfire modeling.

\section*{Acknowledgement}
Research presented in this article was supported by the Laboratory Directed Research and Development program of Los Alamos National Laboratory under project number 20220024DR and 20260010DR. This research was also funded by the Department of Defense, Environmental Security Technology Certification Program, grant number RC24-8162. The findings and conclusions in this publication are those of the authors and should not be construed to represent any official USDA or U.S. Government determination or policy.

\bibliographystyle{elsarticle-num} 

\section*{Appendix A: Expected Point Consistency Guarantee via ELBO Minimization}

\textbf{Assumptions.} Let $(\X_0, \Y)$ denote a paired TLS/ALS point cloud sample drawn from the empirical training distribution $\mathcal{D}$. Let $\mathcal{B}_{\text{ALS}}(\Y) = \mathrm{Conv}(\Y)$ denote the convex hull of the ALS point cloud. Define the point-level containment indicator $C(\x, \Y) = 1(\x \in \mathcal{B}_{\text{ALS}}(\Y))$.

We assume:
\begin{enumerate}
    \item \textbf{Empirical containment:} TLS points are spatially bounded by the ALS convex hull in expectation:
    \[
        \mathbb{E}_{(\X_0, \Y) \sim \mathcal{D}} \left[ \mathbb{E}_{\x \sim \X_0} \left[ C(\x, \Y) \right] \right] \geq 1 - \epsilon
    \]
for a small constant $\epsilon \in [0, 1]$.

    \item \textbf{Model approximation:} The learned model $p_\theta(\X_0 \mid \Y)$ satisfies a KL divergence bound:
    \[
        \mathrm{KL}(p_{\text{data}}(\X_0 \mid \Y) \,\|\, p_\theta(\X_0 \mid \Y)) \leq \delta_{\mathrm{KL}}
    \]
    for all $\Y$.
\end{enumerate}

Scope Note. The containment assumption stated above is specific to the tree scale, where $(\X_0$ and $\Y)$ are co-registered ALS and TLS point clouds corresponding to individual trees or small tree clusters. We do not assume global containment over entire plots or landscape-scale point clouds, where occlusion and ALS sparsity can invalidate this condition. In our dataset, this per-tree containment assumption is empirically validated, with over 96\% of TLS points contained within the ALS convex hull on average.

\textbf{Goal.} Show that the expected containment of generated points satisfies:
\[
\mathbb{E}_{\Y} \left[ \mathbb{E}_{\X_0 \sim p_\theta(\cdot \mid \Y)} \left[ \mathbb{E}_{\x \sim \X_0} \left[ C(\x, \Y) \right] \right] \right] \geq 1 - \epsilon - \delta
\]
for some $\delta = \sqrt{2 \delta_{\mathrm{KL}}}$, using Pinsker's inequality.

\vspace{1em}
\textbf{Proof.}
Let
\[
f(\X_0, \Y) = \mathbb{E}_{x \sim \X_0} \left[ C(\x, \Y) \right]
\]
be the average containment score for a TLS sample $\X_0$ with respect to the ALS envelope. By assumption (1), we have:
\[
\mathbb{E}_{(\X_0, \Y) \sim \mathcal{D}} \left[ f(\X_0, \Y) \right] \geq 1 - \epsilon.
\]

Now fix a conditioning ALS point cloud $\Y$. Then:
\[
\left| \mathbb{E}_{\X_0 \sim p_{\text{data}}(\cdot \mid \Y)} f(\X_0, \Y) - \mathbb{E}_{\X_0 \sim p_\theta(\cdot \mid \Y)} f(\X_0, \Y) \right|
\leq \| p_{\text{data}}(\cdot \mid \Y) - p_\theta(\cdot \mid \Y) \|_{\mathrm{TV}}.
\]

Here, $\|\cdot \|_{\text{TV}}$ denotes the total variation norm, a standard metric between probability distributions. It bounds the discrepancy in expectations of bounded functions and is upper-bounded by KL divergence via Pinsker’s inequality \cite{cover2006elements}. Applying such inequality:
\[
\| p_{\text{data}}(\cdot \mid \Y) - p_\theta(\cdot \mid \Y) \|_{\mathrm{TV}} \leq \sqrt{ \frac{1}{2} \mathrm{KL}(p_{\text{data}} \,\|\, p_\theta) } \leq \sqrt{ \frac{1}{2} \delta_{\mathrm{KL}} }.
\]

Since $f$ is bounded in $[0, 1]$, we obtain:
\[
\mathbb{E}_{\X_0 \sim p_\theta(\cdot \mid \Y)} f(\X_0, \Y) \geq \mathbb{E}_{\X_0 \sim p_{\text{data}}(\cdot \mid \Y)} f(\X_0, \Y) - \sqrt{ \frac{1}{2} \delta_{\mathrm{KL}} }.
\]

Taking the expectation over $\Y$ and combining with the empirical containment assumption:
\[
\mathbb{E}_{\Y} \left[ \mathbb{E}_{\X_0 \sim p_\theta(\cdot \mid \Y)} f(\X_0, \Y) \right] \geq \mathbb{E}_{\Y} \left[ \mathbb{E}_{\X_0 \sim p_{\text{data}}(\cdot \mid \Y)} f(\X_0, \Y) \right] - \sqrt{2 \delta_{\mathrm{KL}}} \geq 1 - \epsilon - \sqrt{2 \delta_{\mathrm{KL}}}.
\]

Therefore:
\[
\mathbb{E}_{\Y} \left[ \mathbb{E}_{\X_0 \sim p_\theta(\cdot \mid \Y)} \left[ \mathbb{E}_{\x \sim \X_0} \left[ 1(x \in \mathcal{B}_{\text{ALS}}(\Y)) \right] \right] \right] \geq 1 - \epsilon - \sqrt{2 \delta_{\mathrm{KL}}}.
\]
\hfill $\square$

\appendix
\section*{Appendix B: Unconditional 3D Tree Generation with DDPM}
\begin{figure}[ht]
    \includegraphics[width=0.5\linewidth]{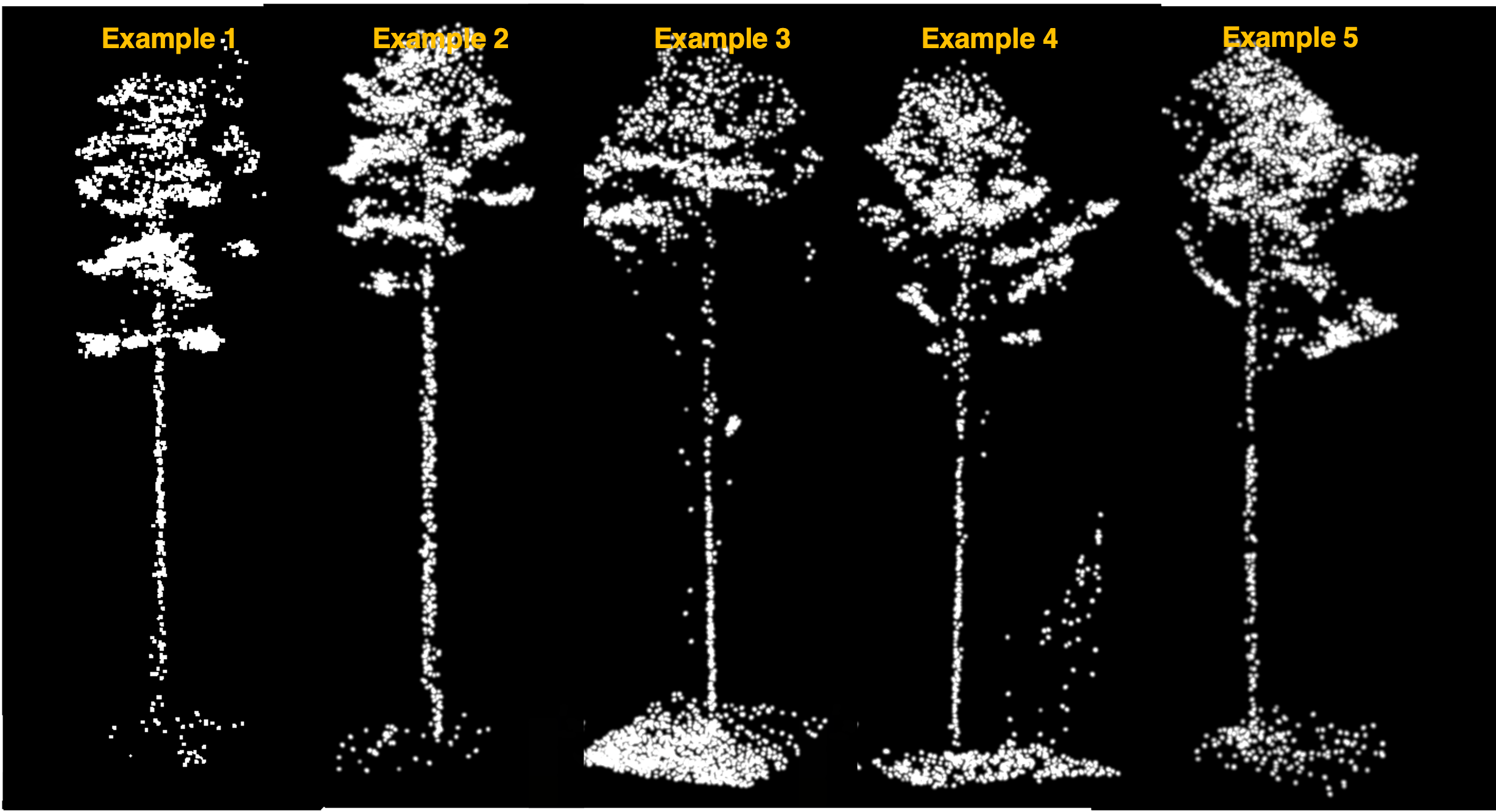}
    \includegraphics[width=0.488\linewidth]{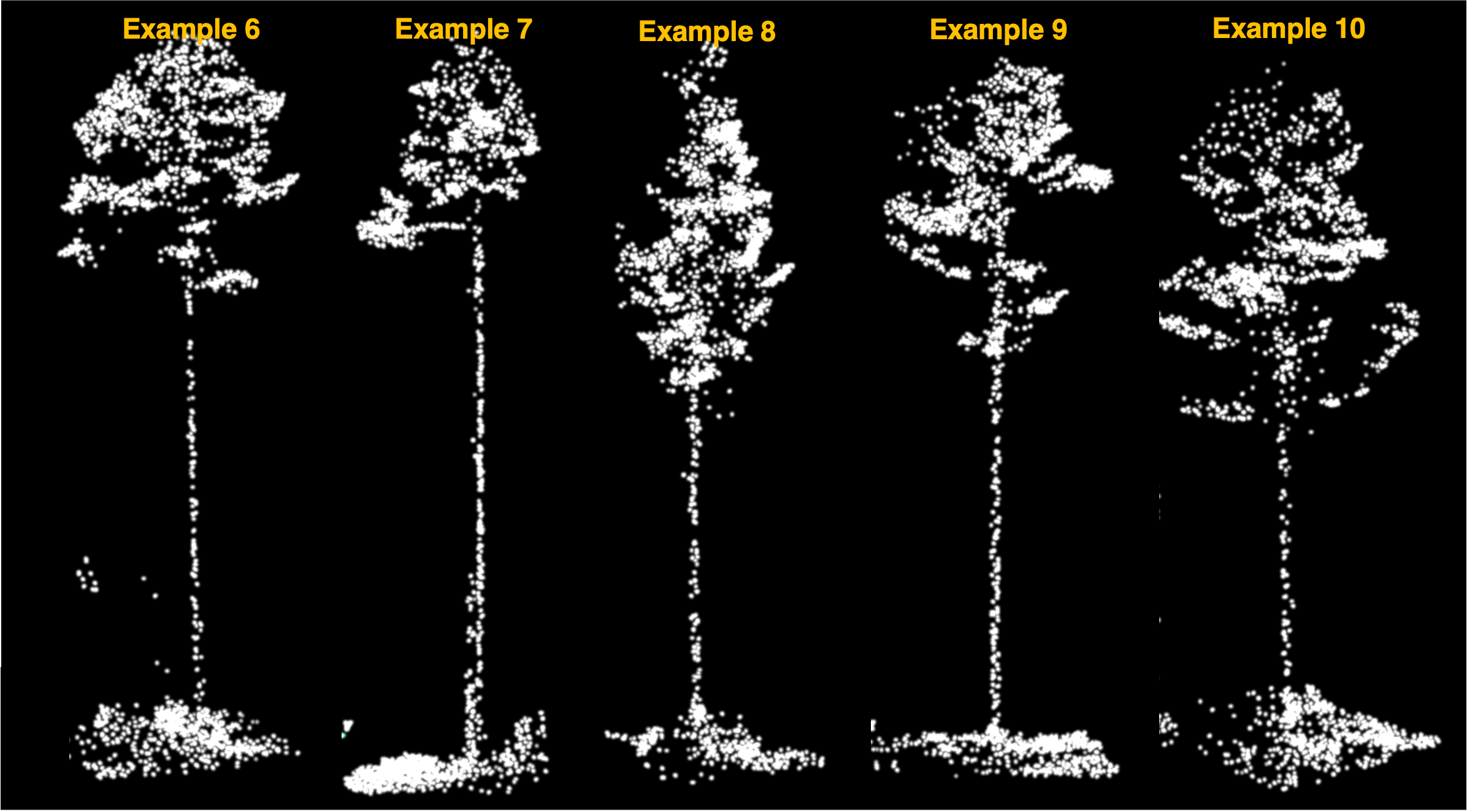}
    \caption{ ForestGen3D} model generation under the unconditional setting: Sample from Gaussian distribution and generate tree structures according to a training TLS scan dataset.
    \label{fig:unconditional_generation}
\end{figure}
In some regions, TLS or co-registered ALS–TLS data may be unavailable due to logistical constraints, cost, or historical limitations. In such cases, it becomes necessary to resort to statistical approximations such as species specific structural priors or to borrow training data from ecologically similar surrogate areas. One advantage of the proposed DDPM framework is that it supports unconditional generation, allowing plausible 3D vegetation structure to be synthesized even without ALS guidance. This is particularly useful in applications where only species composition, vegetation type, or biophysical statistical information is known to populate a geo-referenced region.
We include here several illustrative examples at the tree-scale that demonstrate that the model, even in the absence of ALS conditioning, can produce tree structures that are structurally realistic and ecologically consistent with those found in the training distribution.

Figure \ref{fig:unconditional_generation} presents ten illustrative samples from the model, trained on the 2900 extracted trees from TLS scans without using ALS conditioning. The generated tree structures exhibit fine details along the vertical tree-stand direction, mirroring the characteristics of the TLS training data. The model successfully learns the spatial distribution of key tree components: trunk, branches, leaves, and ground, along with their spatial relationships.
Additionally, the model captures variations in canopy structure despite being trained on a relatively small dataset. The generated trees predominantly resemble mixed conifer species, including longleaf, slash, and loblolly pines, reflecting their prevalence in the training dataset. To the best of our knowledge, this model achieves an unprecedented level of detail in tree generation directly from real LiDAR forestry data.


\section*{Appendix C: Additional comparisons of biophysical metric distributions}

\begin{figure}[ht]
    \centering
	   \includegraphics[height=0.8\textwidth, width=0.8\linewidth]{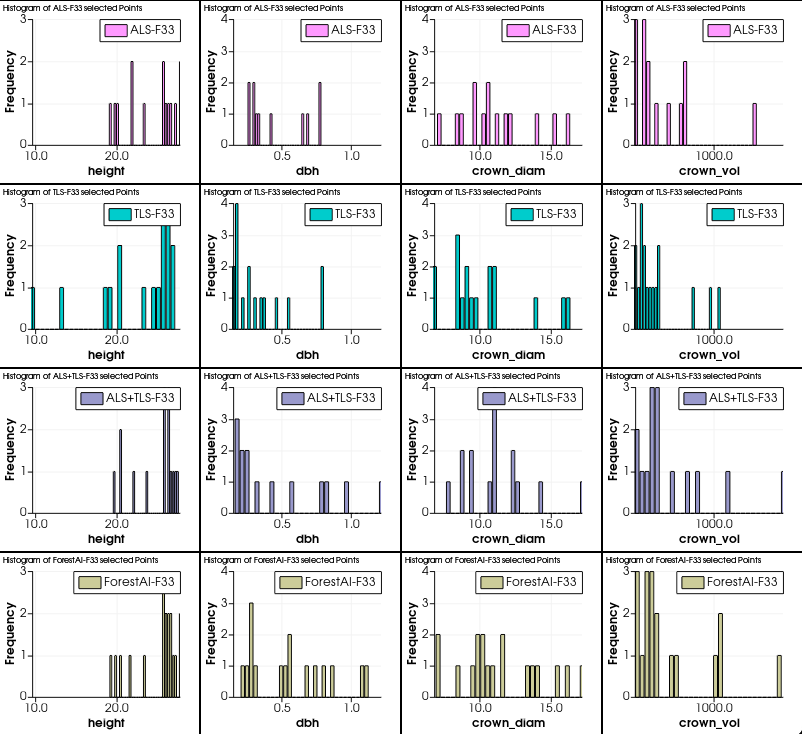}
	     \caption{Plot-scale distributions of biophysical tree metrics derived from different sensing sources. Rows correspond to ALS (row 1), TLS (row 2), ALS+TLS (row 3), and ALS+ForestGen3D (row 4). Columns show distributions for tree height (col. 1), diameter at breast height (DBH; col. 2), crown diameter (CrD; col. 3), and crown volume (CrV; col. 4). Each histogram represents metrics computed over a 25~m radius plot. ALS+TLS is treated as the structural reference, while ALS+ForestGen3D closely approximates its metric distributions, indicating successful recovery of sub-canopy structure while preserving ALS-derived spatial constraints.}
	     \label{fig:ex2_distf33}
        %
\end{figure}

\begin{figure}
    \centering        
        %
\includegraphics[height=0.8\textwidth,width=0.8\linewidth]{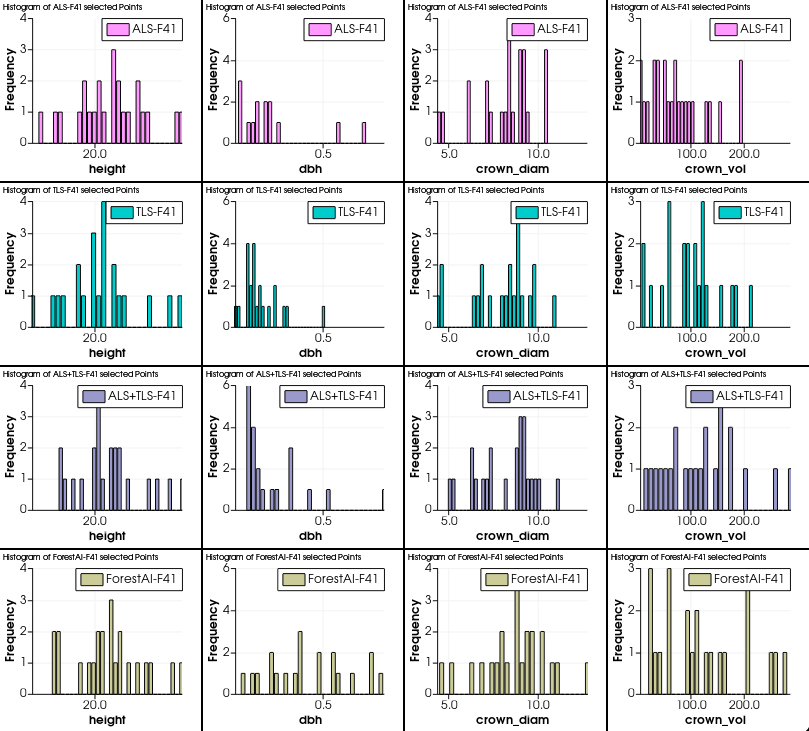}
	     \caption{Plot-scale distributions of biophysical tree metrics derived from different sensing sources. Rows correspond to ALS (row 1), TLS (row 2), ALS+TLS (row 3), and ALS+ForestGen3D (row 4). Columns show distributions for tree height (col. 1), diameter at breast height (DBH; col. 2), crown diameter (CrD; col. 3), and crown volume (CrV; col. 4). Each histogram represents metrics computed over a 25~m radius plot. ALS+TLS is treated as the structural reference, while ALS+ForestGen3D closely approximates its metric distributions, indicating successful recovery of sub-canopy structure while preserving ALS-derived spatial constraints.}
	     \label{fig:ex3_distf26}
        %
\end{figure}

\begin{figure}
    \centering
    %
	   \includegraphics[height=0.8\textwidth, width=0.8\linewidth]{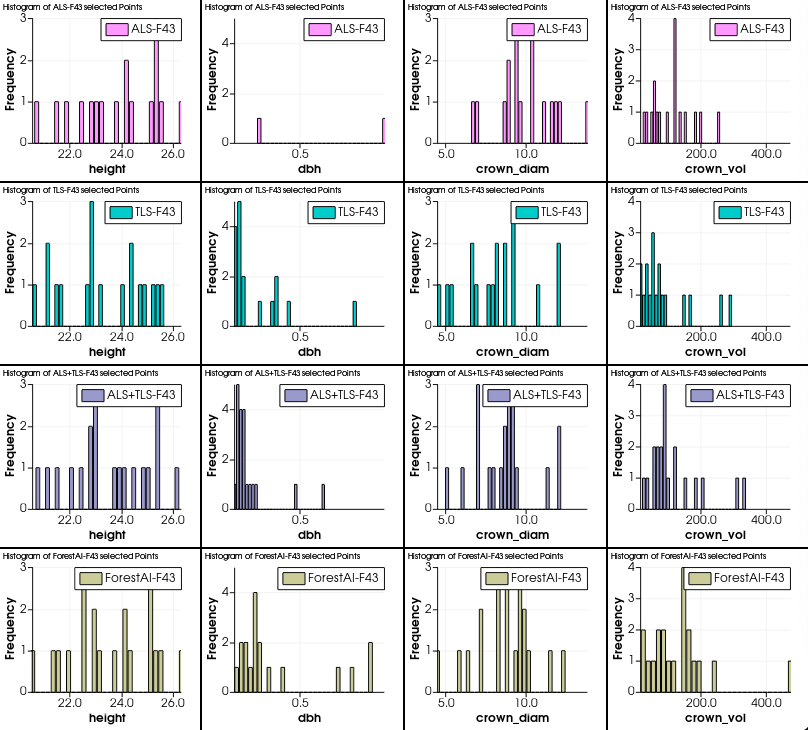}
	     \caption{Plot-scale distributions of biophysical tree metrics derived from different sensing sources. Rows correspond to ALS (row 1), TLS (row 2), ALS+TLS (row 3), and ALS+ForestGen3D (row 4). Columns show distributions for tree height (col. 1), diameter at breast height (DBH; col. 2), crown diameter (CrD; col. 3), and crown volume (CrV; col. 4). Each histogram represents metrics computed over a 25~m radius plot. ALS+TLS is treated as the structural reference, while ALS+ForestGen3D closely approximates its metric distributions, indicating successful recovery of sub-canopy structure while preserving ALS-derived spatial constraints.}
	     \label{fig:ex4_distf26}
\end{figure}

Figures~\ref{fig:ex2_distf33}, \ref{fig:ex3_distf26}, and \ref{fig:ex4_distf26} present additional plot-scale examples (25~m radius) of biophysical metric distributions for tree height, DBH, crown diameter, and crown volume, derived from ALS, TLS, ALS+TLS, and ALS+ForestGen3D point clouds. Across all examples, ALS+ForestGen3D consistently compensates for ALS structural blind spots while preserving ALS-derived spatial geometry. Tree height and crown diameter distributions remain largely contained within ALS-derived ranges, whereas the most pronounced improvements are observed for DBH and crown volume, reflecting recovery of sub-canopy and lower-canopy structure absent in ALS-only data. Collectively, these results further confirm that ForestGen3D produces structurally plausible 3D vegetation whose biometric distributions closely align with TLS-derived references while maintaining fidelity to ALS-provided canopy geometry.

\section*{Appendix D: Additional Regional-scale Generations}

\begin{figure}[ht]
    \centering
    \begin{subfigure}{0.3\linewidth}
	\centering
	\includegraphics[width=1.0\linewidth]{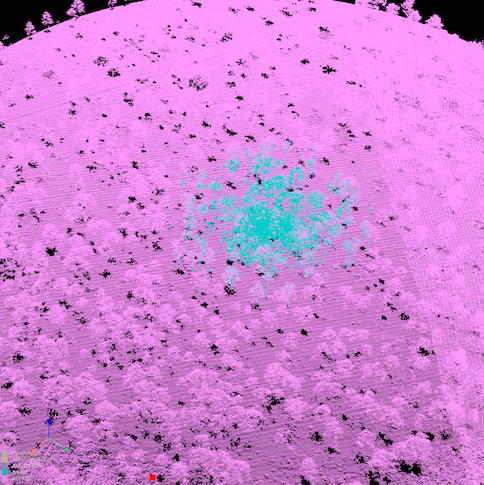}
	\caption{Ex.1: ALS (magenta)+TLS (Turquoise).}
	\label{fig:ex1_als_tls}
    \end{subfigure}
    \begin{subfigure}{0.3\linewidth}
	\centering
	\includegraphics[ width=1.0\linewidth]{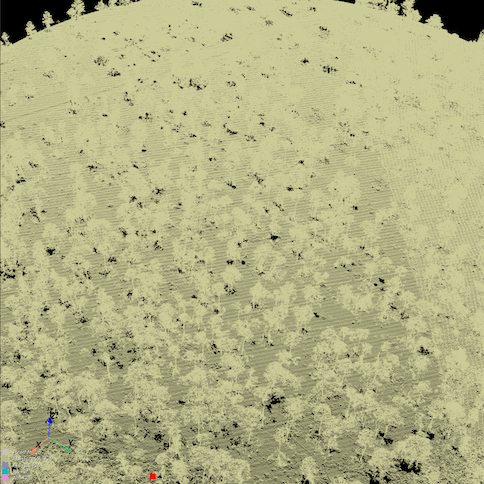}
	\caption{Ex.1: ALS + ForestGen3D (Peach)}
	\label{fig:ex1_als_gen}
    \end{subfigure}
    \begin{subfigure}{0.325\linewidth}
	\includegraphics[width=1.0\linewidth]{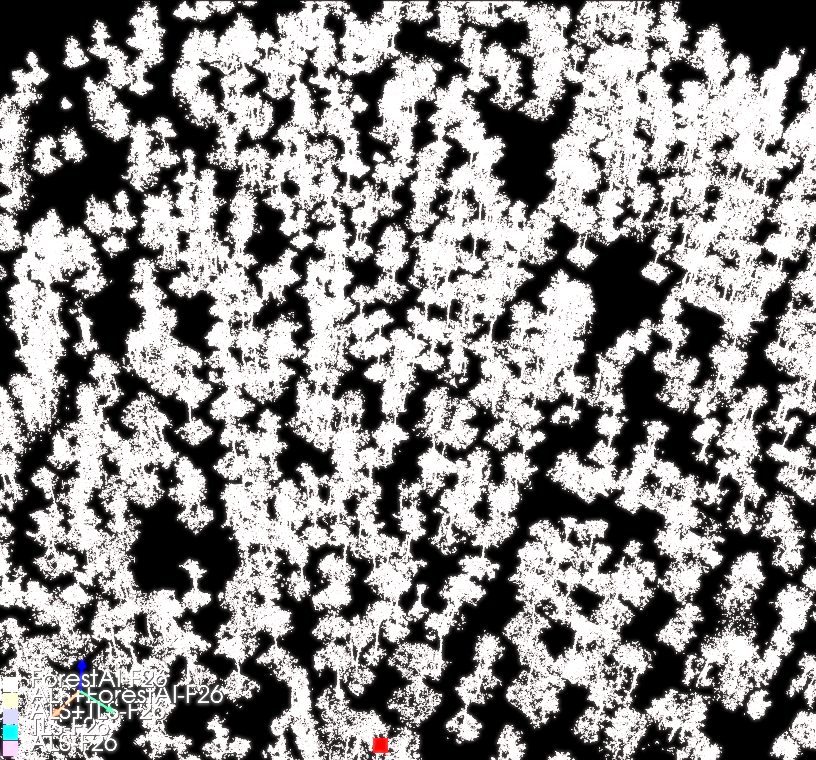}
	\caption{Ex.1: ForestGen3D (Light gray)}
	\label{fig:ex1_gen}
    \end{subfigure}

    \begin{subfigure}{0.31\linewidth}
	\centering
	\includegraphics[width=1.0\linewidth]{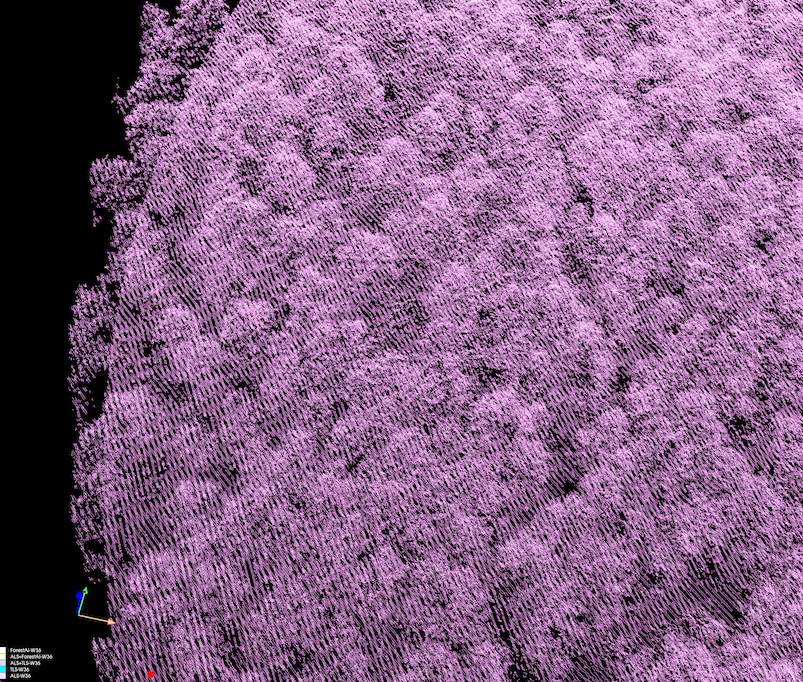}
	\caption{Ex.2: ALS (magenta)+TLS (Turquoise).}
	\label{fig:ex2_als_tls}
    \end{subfigure}
    \begin{subfigure}{0.31\linewidth}
	\centering
	\includegraphics[ width=1.0\linewidth]{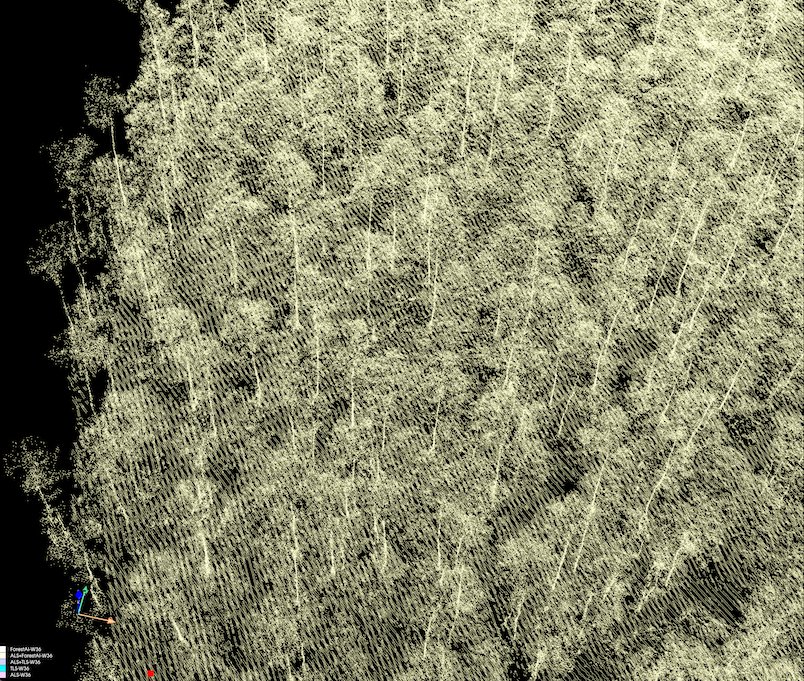}
	\caption{Ex.2: ALS + ForestGen3D (Peach)}
	\label{fig:ex2_als_gen}
    \end{subfigure}
    \begin{subfigure}{0.31\linewidth}
	\includegraphics[width=1.0\linewidth]{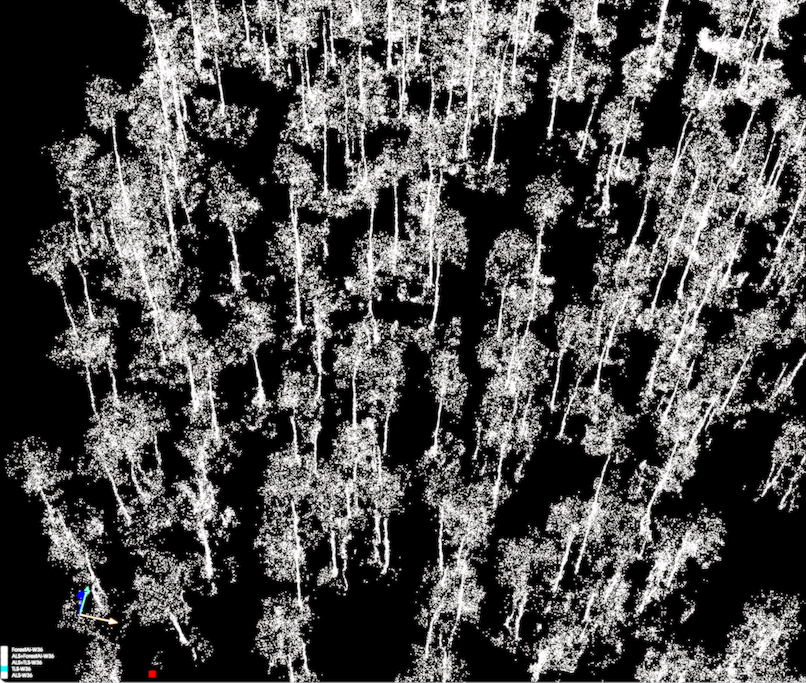}
	\caption{Ex.2: ForestGen3D (Light gray)}
	\label{fig:ex2_gen}
    \end{subfigure}
    \caption{Additional examples of landscape-scale ForestGen3D generation. Each row corresponds to a distinct 200~m radius area centered on a TLS scan location. Columns show (left) ALS combined with TLS reference data, (middle) ALS combined with ForestGen3D output, and (right) ForestGen3D-generated points alone. Point clouds are color-coded as follows: magenta for ALS, turquoise for TLS, peach for ALS + ForestGen3D, and white for ForestGen3D-generated points. ForestGen3D is applied selectively to regions exhibiting occlusion in the ALS data, generating sub-canopy structure while preserving ALS-observed geometry.}
    \label{fig:landscape_appendix1}
\end{figure}

\begin{figure}[ht]
    \centering
    \begin{subfigure}{0.31\linewidth}
	\centering
	\includegraphics[width=1.0\linewidth]{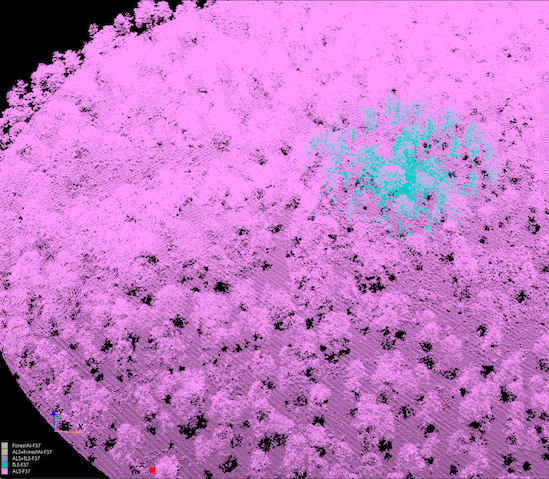}
	\caption{Ex.3: ALS (magenta)+TLS (Turquoise).}
	\label{fig:ex3_als_tls}
    \end{subfigure}
    \begin{subfigure}{0.31\linewidth}
	\centering
	\includegraphics[ width=1.0\linewidth]{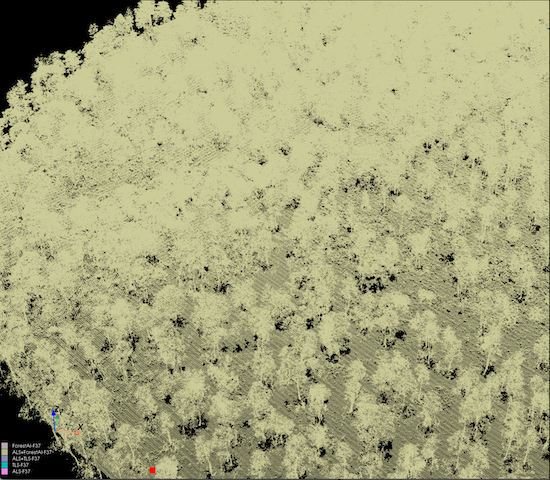}
	\caption{Ex.3: ALS + ForestGen3D (Peach)}
	\label{fig:ex3_als_gen}
    \end{subfigure}
    \begin{subfigure}{0.31\linewidth}
	\includegraphics[width=1.0\linewidth]{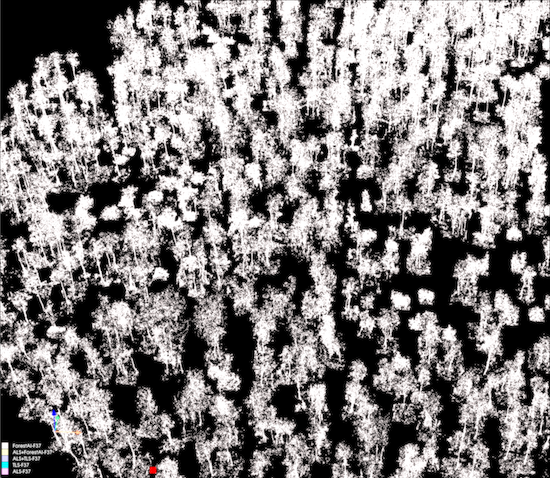}
	\caption{Ex.3: ForestGen3D (Light gray)}
	\label{fig:ex3_gen}
    \end{subfigure}

    \begin{subfigure}{0.31\linewidth}
	\centering
	\includegraphics[width=1.0\linewidth]{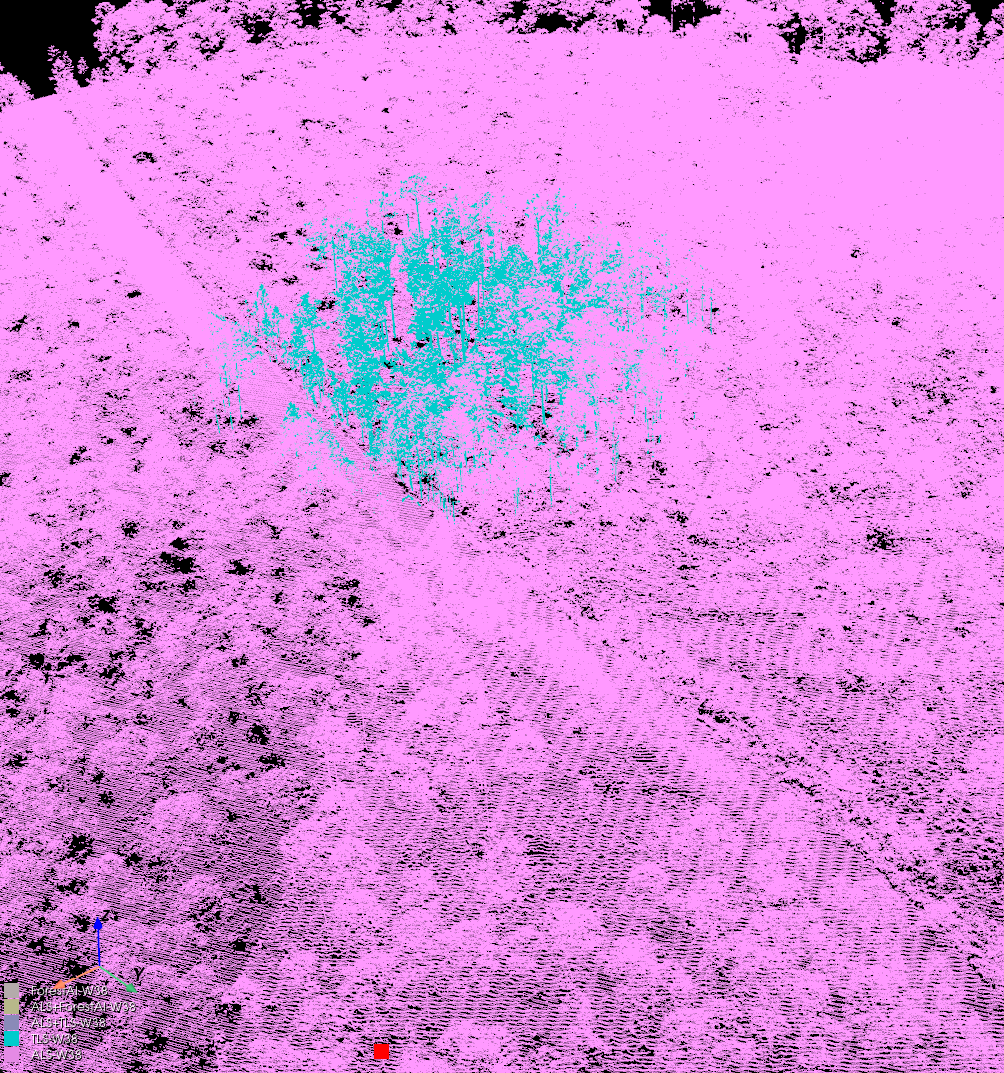}
	\caption{Ex.4: ALS (magenta)+TLS (Turquoise).}
	\label{fig:ex4_als_tls}
    \end{subfigure}
    \begin{subfigure}{0.31\linewidth}
	\centering
	\includegraphics[ width=1.0\linewidth]{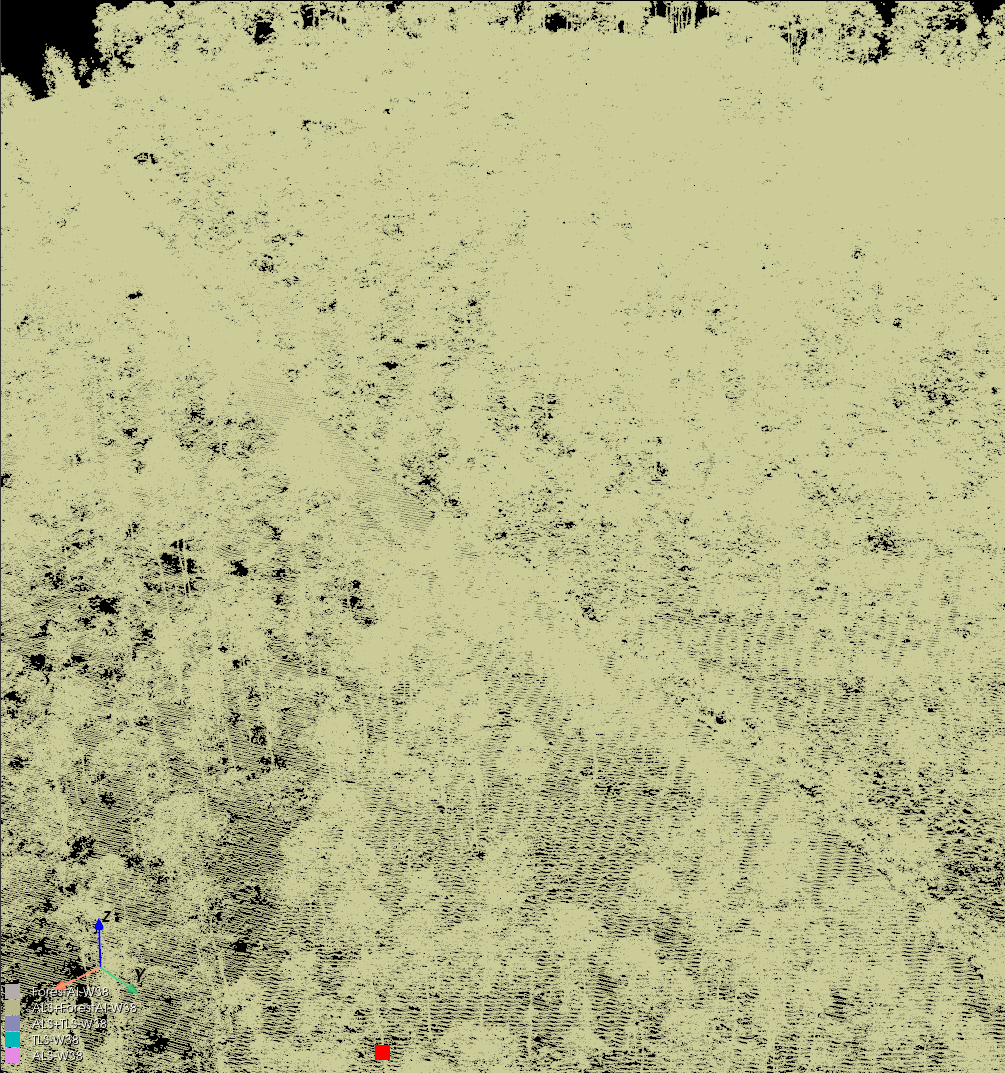}
	\caption{Ex.4: ALS + ForestGen3D (Peach)}
	\label{fig:ex4_als_gen}
    \end{subfigure}
    \begin{subfigure}{0.31\linewidth}
	\includegraphics[width=1.0\linewidth]{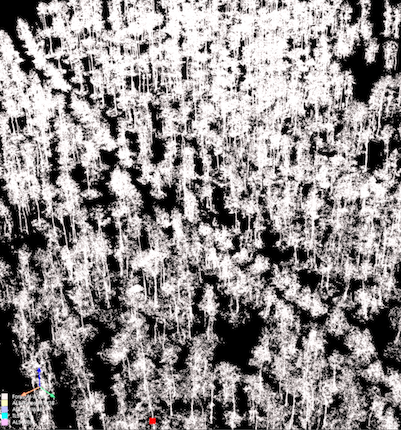}
	\caption{Ex.4: ForestGen3D (Light gray)}
	\label{fig:ex4_gen}
    \end{subfigure}

    \begin{subfigure}{0.31\linewidth}
	\centering
	\includegraphics[width=1.0\linewidth]{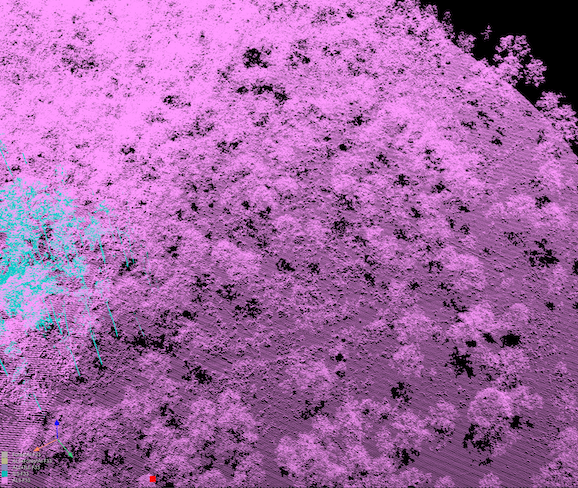}
	\caption{Ex.5: ALS (magenta)+TLS (Turquoise).}
	\label{fig:ex5_als_tls}
    \end{subfigure}
    \begin{subfigure}{0.31\linewidth}
	\centering
	\includegraphics[ width=1.0\linewidth]{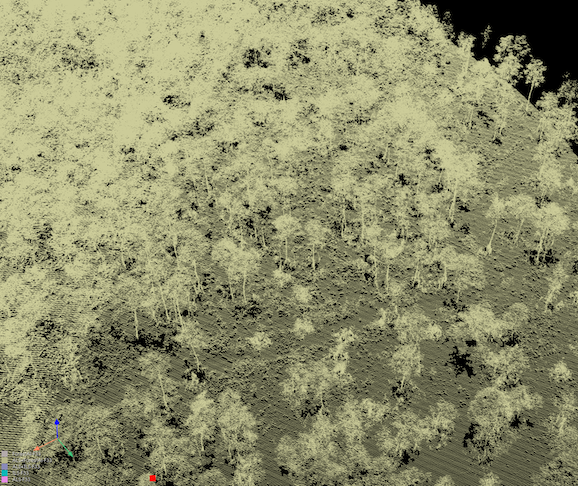}
	\caption{Ex.5: ALS + ForestGen3D (Peach)}
	\label{fig:ex5_als_gen}
    \end{subfigure}
    \begin{subfigure}{0.31\linewidth}
	\includegraphics[width=1.0\linewidth]{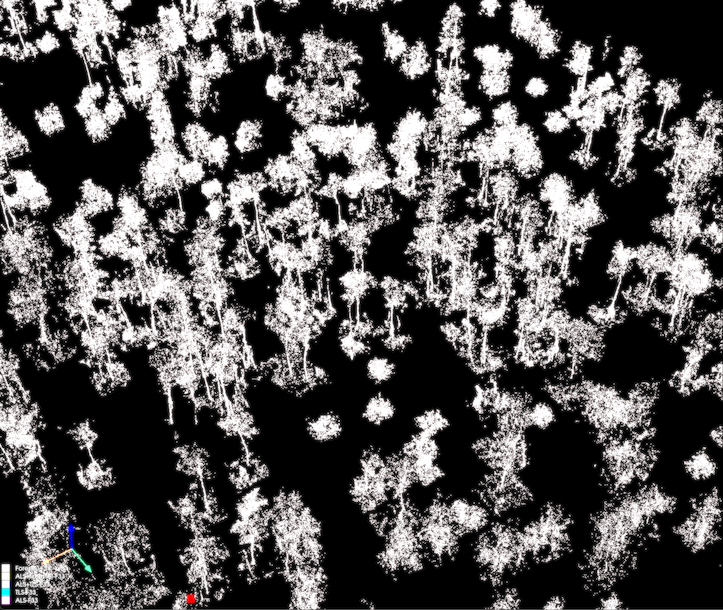}
	\caption{Ex.5: ForestGen3D (Light gray)}
	\label{fig:ex5_gen}
    \end{subfigure}

    \begin{subfigure}{0.31\linewidth}
	\centering
	\includegraphics[width=1.0\linewidth]{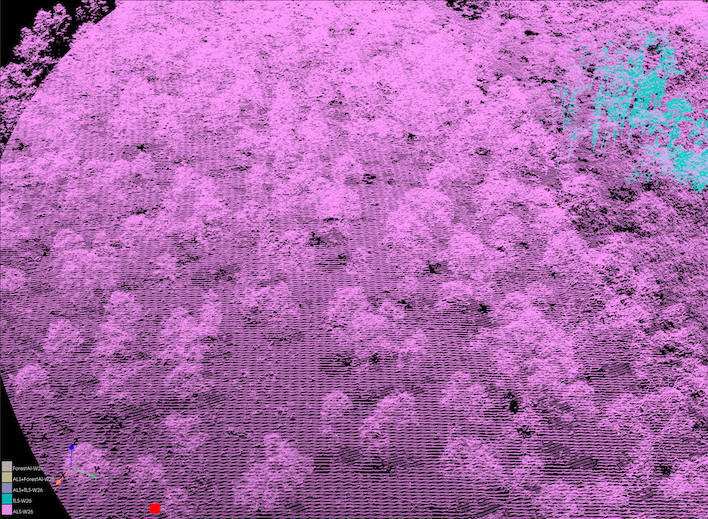}
	\caption{Ex.6: ALS (magenta)+TLS (Turquoise).}
	\label{fig:ex6_als_tls}
    \end{subfigure}
    \begin{subfigure}{0.31\linewidth}
	\centering
	\includegraphics[ width=1.0\linewidth]{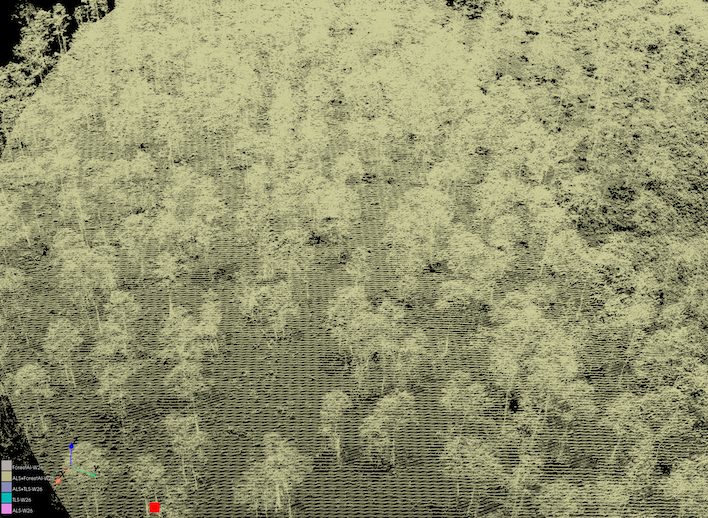}
	\caption{Ex.6: ALS + ForestGen3D (Peach)}
	\label{fig:ex6_als_gen}
    \end{subfigure}
    \begin{subfigure}{0.31\linewidth}
	\includegraphics[width=1.0\linewidth]{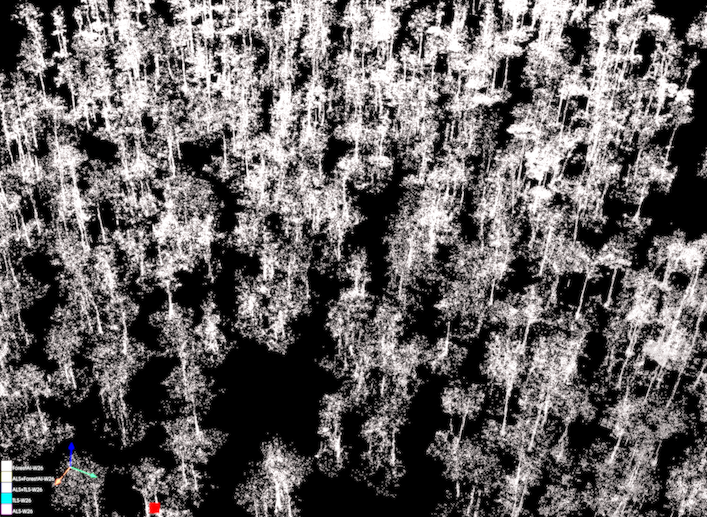}
	\caption{Ex.6: ForestGen3D (Light gray)}
	\label{fig:ex6_gen}
    \end{subfigure}

    \caption{Additional landscape-scale ForestGen3D examples. Rows show different 200~m radius areas; columns show ALS+TLS (left), ALS+ForestGen3D (middle), and ForestGen3D alone (right). Colors denote ALS (magenta), TLS (turquoise), ALS+ForestGen3D (peach), and generated points (white).}
    \label{fig:landscape_appendix2}
\end{figure}

This section provides additional visual examples of ForestGen3D outputs at the regional scale. Figures \ref{fig:landscape_appendix1} and \ref{fig:landscape_appendix2} show spatial snapshots of 200~m radius regions, with point clouds color-coded as follows: magenta for ALS, turquoise for TLS, peach for ALS combined with ForestGen3D output, and white for ForestGen3D-generated points alone. Note that ForestGen3D is applied only within regions identified as occluded in the ALS data. Thus, figures displaying only the generated points contain empty areas corresponding to locations where no synthesis was performed. When combined with the original ALS observations, these regions are filled by existing ALS ground and canopy returns, resulting in a complete point cloud representation. The generated points primarily populate the lower canopy and sub-canopy layers and extend toward the ground surface, following the spatial constraints imposed by the ALS geometry. The synthesized structures exhibit continuous transitions between ALS-observed and generated regions, yielding coherent tree-level and stand-level configurations. Each 200~m radius region contains approximately 3,000 trees. With an average inference time of about 0.4~seconds per tree, generation of a full region requires on the order of 20~minutes, illustrating the computational feasibility of applying ForestGen3D at regional scales.

\end{document}